\documentclass[twoside,11pt]{article}

\usepackage{jmlr2e}

\usepackage{graphicx}
\usepackage{amsmath,amssymb}
\usepackage{lscape}
\usepackage{rotating}
\usepackage{multirow}
\usepackage{booktabs}
\usepackage{bm}
\usepackage{url}
\usepackage{subcaption}  
\usepackage{tikz}
\usepackage{floatrow}

\usetikzlibrary{decorations.pathreplacing}
\usetikzlibrary{arrows,decorations.pathmorphing,fit,positioning}
\usetikzlibrary{calc}

\newfloatcommand{capbtabbox}{table}[][\FBwidth]

\def\|{\,|\,}

\def\obj{{\bf x}}
\def\eqref#1{Eq~\ref{#1}}

\def\P{{\rm P}}

\def\tinyspace{\mathchoice{\hspace*{1pt}}{\hspace*{0.75pt}}{\hspace*{0.5pt}}{\hspace*{0.33pt}}}
\def\mid{\tinyspace|\tinyspace}

\def\graph{{\mathcal{G}}}
\def\bn{{\mathcal{B}}}

\def\data{\ifmmode \mathcal D\else$\data$\fi}
\def\labels{\ifmmode \mathcal L\else$\labels$\fi}
\def\card{\ifmmode \mathcal X\else$\card$\fi}

\def\k{\ifmmode \|\!\mathcal{Y}\!\|\else$\k$\fi}

\def\train{\ifmmode \mathcal T\else$\train$\fi}
\def\test{\ifmmode \mathcal U\else$\train$\fi}
\def\model{\ifmmode \mathcal M\else$\model$\fi}

\def\P{{\rm P}}
\def\PLR{\ensuremath{\P_{\!\textit{LR}}}}
\def\PLRN{\ensuremath{\P_{\!\textit{LR}^n}\!}}
\def\PNB{\ensuremath{\P_{\!\textit{NB}}}}
\def\PW{\ensuremath{\P_{\!\textit{W}}}}

\def\ANDE^#1{\mathop{{\rm A}#1{\rm DE}}}

\def\weight{\ensuremath{w}}

\def\graph{{\mathcal{G}}}
\def\bn{{\mathcal{B}}}

\def\LR^#1{\ensuremath{\text{LR}^#1}}
\def\DBL^#1{\ensuremath{\text{DBL}^#1}}
\def\ALR^#1{\ensuremath{\text{ALR}^#1}}

\def\partition{\ensuremath{\mathcal P}}

\def\partitions{\ensuremath{\Upsilon^{\mathcal{A}}_N}}
\def\Pr{{\rm P}}
\def\attvals{{\bf x}}

\let\originalleft\left
\let\originalright\right
\def\left#1{\mathopen{}\originalleft#1}
\def\right#1{\originalright#1\mathclose{}}
\def\leftBig{\mathopen{}\Big}
\def\leftbig{\mathopen{}\big}

\jmlrheading{1}{2015}{1-24}{09/15; Revised 00/00}{00/00}{Nayyar~A.\ Zaidi and Francois Petitjean and Geoffrey~I.\ Webb}

\ShortHeadings{Deep Broad Learning}{Zaidi and Webb and Carman and Petitjean}

\firstpageno{1}

\begin{document}

%\title{Deep Broad Learning (DBL): Big Models for Big Data}
\title{Deep Broad Learning - Big Models for Big Data}

\author{\name Nayyar~A.~Zaidi \email nayyar.zaidi@monash.edu \\
	\name Geoffrey~I.~Webb \email geoff.webb@monash.edu \\
	\name Mark~J.~Carman \email mark.carman@monash.edu \\
	\name Francois Petitjean \email francois.petitjean@monash.edu \\
	\addr Faculty of Information Technology \\ Monash University \\ VIC 3800, Australia.
       }

\editor{xxxxxx xxxxxx}

\maketitle

%%%%%%%%%%%%%%%%%%%%%%%%%%%%%%%%%%%%%%%%%%%%%%%%%%%%%%%%%%%%%%%%%%%%%%%%%%%%%%%%%%%%%%%%%%%%%%%%%%%%%%%%
\begin{abstract}
Deep learning has demonstrated the power of detailed modeling of complex high-order (multivariate) interactions in data.
For some learning tasks there is power in learning models that are not only \emph{Deep} but also \emph{Broad}.
By \emph{Broad}, we mean models that incorporate evidence from large numbers of features. 
This is of especial value in applications where many different features and combinations of features all carry small amounts of information about the class.  
The most accurate models will integrate all that information.
In this paper, we propose an algorithm for \emph{Deep Broad Learning} called DBL. 
%\gw{Is the next sentence correct?} 
The proposed algorithm has a tunable parameter $n$, that specifies the depth of the model.
It provides straightforward paths towards out-of-core learning for large data.
We demonstrate that DBL learns models from large quantities of data with accuracy that is highly competitive with the state-of-the-art.
\end{abstract}
%%%%%%%%%%%%%%%%%%%%%%%%%%%%%%%%%%%%%%%%%%%%%%%%%%%%%%%%%%%%%%%%%%%%%%%%%%%%%%%%%%%%%%%%%%%%%%%%%%%%%%%%
%%%%%%%%%%%%%%%%%%%%%%%%%%%%%%%%%%%%%%%%%%%%%%%%%%%%%%%%%%%%%%%%%%%%%%%%%%%%%%%%%%%%%%%%%%%%%%%%%%%%%%%%

%%%%%%%%%%%%%%%%%%%%%%%%%%%%%%%%%%%%%%%%%%%%%%%%%%%%%%%%%%%%%%%%%%%%%%%%%%%%%%%%%%%%%%%%%%%%%%%%%%%%%%%%
\section{Introduction} \label{sec_intro}
%%%%%%%%%%%%%%%%%%%%%%%%%%%%%%%%%%%%%%%%%%%%%%%%%%%%%%%%%%%%%%%%%%%%%%%%%%%%%%%%%%%%%%%%%%%%%%%%%%%%%%%%

\begin{figure}[t]
  \centering
  \includegraphics[width=.8\linewidth]{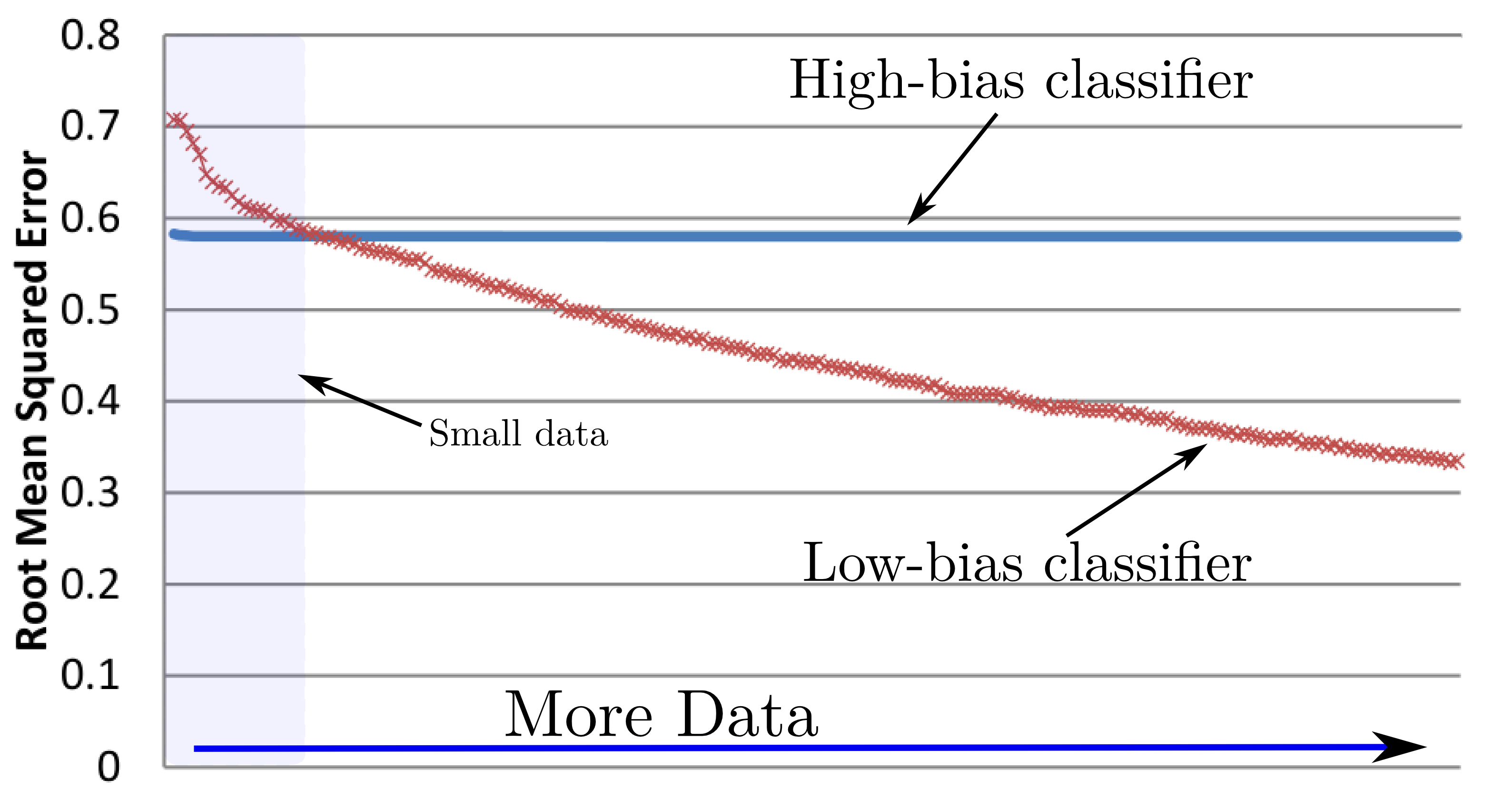}
  \caption{\label{fig:intro-learning-curves}Comparative study of the error committed by high- and low-bias classifiers on increasing quantities of data.}
\end{figure}

The rapid growth in data quantity~\citep{Ganz2012} makes it increasingly difficult for machine learning to extract maximum value from current data stores. 
Most state-of-the-art learning algorithms were developed in the context of small datasets. 
However, the amount of information present in big data is typically much greater than that present in small quantities of data.
As a result, big data can support the creation of very detailed models that encode complex higher-order multivariate distributions, whereas, for small data, very detailed models will tend to overfit and should be avoided~\citep{Brain2002,ana14a}. 
We highlight this phenomenon in Figure~\ref{fig:intro-learning-curves}. 
We know that the error of most classifiers decreases as they are provided with more data. 
This can be observed in Figure~\ref{fig:intro-learning-curves} where the variation in error-rate of two classifiers  is plotted with increasing quantities of training data on the poker-hand dataset~\citep{UCI:2005:MLR}. One is a low-bias high-variance learner (KDB $k=5$, taking into account quintic features, \citep{sahami:1996}) and the other is a low-variance high-bias learner (naive Bayes, a linear classifier). 
For small quantities of data, the low-variance learner achieves the lowest error.
However, as the data quantity increases, the low-bias learner comes to achieve the lower error as it can better model higher-order distributions from which the data might be sampled. 

The capacity to model different types of interactions among variables in the data is a  major determinant of a  learner's \emph{bias}.
The greater the capacity of a learner to model differing distributions, the lower its bias will tend to be. 
%However, one key type of interaction that many models have limited capacity to model is complex higher-order interactions between variables in the data.
However, many learners have limited capacity to model complex higher-order interactions.

Deep learning\footnote{Deep neural networks, convolutional deep neural networks, deep belief networks, etc.} has demonstrated some remarkable successes through its capacity to create detailed (deep) models of complex multivariate interactions in structured data (e.g., data in computer vision, speech recognition, bioinformatics, etc.).  
Deep learning can be characterized in several different ways. But the underlying theme is that of learning higher-order interactions among features using a cascade of many layers. This process is known as `feature extraction' and can be un-supervised as it leverages the structure within the data to create new features. Higher-order features are created from lower-order features creating a hierarchical structure.
We conjecture that the deeper the model the higher the order of interactions that are captured in the data and the lower the bias that the model exhibits. 

We argue that in many domains there is value in creating models that are \textit{broad} as well as deep. For example, when using web browsing history, or social network likes, or when analyzing text, it is often the case that each feature  provides only extremely small amounts of information about the target class. It is only by combining very large amounts of this \emph{micro-evidence} that reliable classification is possible.

We call a model \emph{broad}  if it utilizes large numbers of variables.  We call a model deep and broad if it captures many complex interactions each between numerous variables.  
For example, typical linear classifiers such as Logistic Regression (LR) and Naive Bayes (NB) are \emph{Broad Learners}, in that they utilize all variables.
However, these models are not deep, as they do not directly model interactions between variables. 
In contrast, Logistic Regression\footnote{Logistic Regression taking into account all $n$-level features is denoted by \LR^n, e.g., \LR^2, \LR^3, \LR^4, etc.\ takes into account all quadratic, cubic, quartic, etc. features.} with cubic features (\LR^3)~\citep{Langford2007} and Averaged 2-Dependence Estimators (A2DE)~\citep{webb:lbefmt,zaidi12}, both of which consider all combinations of  3 variables,  are both Deep and Broad. 
The parameters of the former are fit discriminatively through computationally intensive gradient descent-based search, while the parameters of the latter are fit generatively using computationally efficient maximum-likelihood estimation. This efficient estimation of A2DE parameters makes it computationally well-suitable for big data.
In contrast, we argue that \LR^3's discriminative parameterization can more closely fit the data than A2DE, making it lower bias and hence likely to have lower error when trained on large training sets.
However, the computation required to optimize the parameters for \LR^3 becomes computationally intensive even on moderate dimensional data.

Recently, it has been shown that it is possible to form a hybrid generative-discriminate learner that exploits the strengths of both naive Bayes (NB) and Logistic Regression (LR) by creating a weighted variant of NB in which the weights are optimized using discriminative minimization of conditional log-likelihood~\citep{zaidi13a,zaidi14}.
From one perspective, the resulting learner can be viewed as  using weights to alleviate the attribute independence assumption of NB. From another perspective it can be seen to use the maximum likelihood parameterization of NB to pre-condition the discriminative search of LR. 
The result is a learner that learns models that are exactly equivalent to LR, but does so much more efficiently.

In this work, we show how to achieve the same result with \LR^n, creating a hybrid generative-discriminative learner named \DBL^n for categorical data that learns equivalent deep broad models to those of \LR^n, but does so more efficiently.  
We further demonstrate that the resulting models have low bias and have very low error on large quantities of data.
However, to create this hybrid learner we must first create an efficient generative counterpart to \LR^n.

In short, the contributions of this work are:
\begin{itemize}
\item developing an efficient generative counter-part to \LR^n, named Averaged $n$-Join Estimators (AnJE), %AnJE will be discussed in Section~\ref{sec_anje},
\item developing \DBL^n, a hybrid of \LR^n and AnJE, %(Section~\ref{sec_wanje}),
\item demonstrating that \DBL^n has equivalent error to LR$^n$, but is more efficient, %(Section~\ref{subsec_DBLvsLR}),
\item demonstrating that \DBL^n has low error on large data. %(Section~\ref{sec_discussion}).
\end{itemize}

%%%%%%%%%%%%%%%%%%%%%%%%%%%%%%%%%%%%%%%%%%%%%%%%%%%%%%%%%%%%%%%%%%%%%%%%%%%%%%%%%%%%%%%%%%%%%%%%%%%%%%%%
\section{Notation}\label{sec_notation}
%%%%%%%%%%%%%%%%%%%%%%%%%%%%%%%%%%%%%%%%%%%%%%%%%%%%%%%%%%%%%%%%%%%%%%%%%%%%%%%%%%%%%%%%%%%%%%%%%%%%%%%%

We seek to assign  a value $y\in\Omega_Y=\{y_1, \ldots y_C\}$ of the class variable
$Y$, to a given example $\mathbf{x} = (x_1, \ldots, x_a)$, where the $x_i$ are value assignments for the $a$ attributes $\mathcal{A} = \{X_1,\ldots, X_a\}$. 
We define $\mathcal{A}\choose n$ as the set of all subsets of $\mathcal{A}$ of size $n$, where each subset in the set is denoted as $\alpha$:
\begin{equation}
{\mathcal{A} \choose n}=\{\alpha\subseteq\mathcal{A}: |\alpha|=n\}. \nonumber
\end{equation}
We use $x_\alpha$ to denote the set of values taken by attributes in the subset $\alpha$ for any data object $\mathbf{x}$.

LR for categorical data learns a weight for every attribute value per class. Therefore, for LR, we denote, $\beta_y$ to be the weight associated with class $y$, and $\beta_{y,i,x_i}$ to be the weight associated with attribute $i$ taking value $x_i$ with class label $y$. 
For \LR^n, $\beta_{y,\alpha,x_{\alpha}}$ specifies the weight associated with class $y$ and attribute subset $\alpha$ taking value $x_\alpha$.
The equivalent weights for \DBL^n are denoted by $\weight_y$, $\weight_{y,i,x_i}$ and $\weight_{y,\alpha,x_\alpha}$.

The probability of attribute $i$ taking  value $x_i$ given class $y$ is denoted by $\P(x_i\|y)$. Similarly, probability of attribute subset $\alpha$, taking value $x_\alpha$ is denoted by $\P(x_\alpha|y)$.

%%%%%%%%%%%%%%%%%%%%%%%%%%%%%%%%%%%%%%%%%%%%%%%%%%%%%%%%%%%%%%%%%%%%%%%%%%%%%%%%%%%%%%%%%%%%%%%%%%%%%%%%
\section{Using generative models to precondition discriminative learning}\label{sec_preconditioning}
%%%%%%%%%%%%%%%%%%%%%%%%%%%%%%%%%%%%%%%%%%%%%%%%%%%%%%%%%%%%%%%%%%%%%%%%%%%%%%%%%%%%%%%%%%%%%%%%%%%%%%%%

There is a direct equivalence between a weighted NB and LR~\citep{zaidi13a,zaidi14}.
We write LR for categorical features as:
\begin{equation} \label{eq_LR}
%\P(y\mid\obj) = \frac{\exp \left(\beta_y + \sum_{i=1}^{a} \beta_{y,x_i} \right)}{\sum_{c\in\Omega_Y} \exp \left( \beta_c + \sum_{j=1}^{a} \beta_{c,x_j} \right)},
\PLR(y\mid\obj) = \exp \leftBig( \beta_y + \sum_{i=1}^{a} \beta_{y,i,x_i} - \log\!\! \sum_{c \in \Omega_Y}\!\! \exp \leftBig( \beta_c + \sum_{j=1}^{a} \beta_{c,j,x_j} \Big) \Big)
\end{equation}
and NB as:
\begin{equation} \label{eq_NB}
\PNB(y\mid\obj) = \frac{\P(y)\prod_{i=1}^{a}\P(x_i\mid y)}{\sum_{c\in\Omega_Y} \P(c)\prod_{i=1}^{a}\P(x_i\mid c) }. \nonumber
\end{equation}
One can add the weights in NB to alleviate the attribute independence assumption, resulting in the WANBIA-C formulation, that can be written as: 
\begin{align} 
\PW(y\mid\obj) & =\frac{\P(y)^{\weight_y}\prod_{i=1}^{a}\P(x_i\mid y)^{\weight_{y,i,x_i}}}{\sum_{c\in\Omega_Y} \P(c)^{\weight_c}\prod_{j=1}^{a}\P(x_i\mid c)^{\weight_{c,j,x_j}} } \nonumber \\
               & = \exp  \leftBig( \weight_y \log\P(y) + \sum_{i=1}^{a} \weight_{y,i,x_i}\log\P(x_i\mid y)  - \nonumber \\
               &   \log \sum_{c\in\Omega_Y} \exp \leftBig(\weight_c \log\P(c) + \sum_{j=1}^{a}\! \weight_{c,j,x_j} \log\P(x_j\!\mid c) \Big) \Big). \label{eq_WNB}
\end{align}
%\gw{Don't we need a log before the sum in the line above?}
When conditional log likelihood (CLL) is maximized for LR and weighted NB using Equation~\ref{eq_LR} and~\ref{eq_WNB} respectively, we get an equivalence such that $\beta_c\propto\weight_c\log\P(c)$ and $\beta_{c,i,x_i}\propto\weight_{c,i,x_i}\log\P(x_i\mid c)$.  Thus, WANBIA-C and LR generate equivalent models.  
While it might seem less efficient to use WANBIA-C which has twice the number of parameters of LR, the probability estimates are learned very efficiently using maximum likelihood estimation, and provide useful information about the classification task that in practice serve to effectively precondition the search for the parameterization of weights to maximize conditional log likelihood.

%%%%%%%%%%%%%%%%%%%%%%%%%%%%%%%%%%%%%%%%%%%%%%%%%%%%%%%%%%%%%%%%%%%%%%%%%%%%%%%%%%%%%%%%%%%%%%%%%%%%%%%%
\section{Deep Broad Learner (DBL)}\label{sec_dbl}
%%%%%%%%%%%%%%%%%%%%%%%%%%%%%%%%%%%%%%%%%%%%%%%%%%%%%%%%%%%%%%%%%%%%%%%%%%%%%%%%%%%%%%%%%%%%%%%%%%%%%%%%

In order to create an efficient and effective low-bias learner, we want to perform the same trick that is used by WANBIA-C for LR with higher-order categorical features.  We define \LR^n as:
\begin{equation} \label{eq_LRk}
\PLRN(y\mid\obj) = \frac{\exp\leftbig(\beta_y + \sum_{\alpha\in{\mathcal{A}\choose n}} \beta_{y,\alpha,x_\alpha}\big)}
                     {\sum_{c\in\Omega_Y} \exp \leftbig( \beta_c + \sum_{\alpha^*\in{\mathcal{A}\choose n}} \beta_{c,\alpha^*,x_{\alpha^*}} \big)}.
\end{equation}
We do not include lower-order terms. For example, if $n=2$ we do not include terms for $\beta_{y,i,x_i}$ as well as for $\beta_{y,i,x_i,j,x_j}$, because doing so does not increase the space of distinct distributions that can be modeled but does increase the number of parameters that must be optimized.

To precondition this model using generative learning, we need a generative model of the form
\begin{align}
\P(y\hspace*{1.5pt}|\hspace*{1.5pt}\obj) & = \frac{\P(y) \prod_{\alpha\in{\mathcal{A}\choose n}} \P(x_\alpha \mid y)}{\prod_{c\in\Omega_Y} \left( \P(c) \prod_{\alpha^*\in{\mathcal{A}\choose n}}\P(x_{\alpha^*}\mid c)\right)} \label{eq:BNx} \\
              & = \exp\leftBig(\log\P(y) + \sum_{\alpha\in{\mathcal{A}\choose n}} \log\P(x_\alpha\mid y) - \nonumber \\
              & \quad\log\! \sum_{c\in\Omega_Y} \exp\leftBig( \log\P(c) + \sum_{\alpha^*\in{\mathcal{A}\choose n}} \log\P(x_{\alpha^*}\mid c)\Big)\Big). 
\label{eq:BNxlog}
\end{align}
The only existing generative model of this form is a log-linear model, which requires computationally expensive conditional log-likelihood optimization and consequently would not be efficient to employ.  It is not possible to create a Bayesian network of this form as it would require that $\P(x_i,x_j)$ be independent of $\P(x_i,x_k)$.
However, we can use a variant of the AnDE~\citep{webb:lbefmt,webb:nsnbao} approach of averaging many Bayesian networks. 
Unlike AnDE, we cannot use the arithmetic mean, as we require a product of terms in Equation~\ref{eq:BNx} rather than a sum, so we must instead use a geometric mean.

%%%%%%%%%%%%%%%%%%%%%%%%%%%%%%%%%%%%%%%%%%%
\subsection{Averaged n-Join Estimators (AnJE)} \label{sec_anje}
%%%%%%%%%%%%%%%%%%%%%%%%%%%%%%%%%%%%%%%%%%%

%We can create a base Bayesian network classifier of the form we need to average by partitioning the attribute values.  
Let $\partition$ be a partition of the attributes $\mathcal{A}$. By assuming independence only between the sets of attributes $A \in \partition$ one obtains an n-joint estimator:
\begin{equation}\label{eq:singleModel}
\Pr_{\textrm{AnJE}}(\attvals\mid y)=\prod_{\alpha\in\partition}\Pr(x_\alpha\mid y). \nonumber
\end{equation}
For example, if there are four attributes $X_1$, $X_2$ , $X_3$ and  $X_4$ that are partitioned into the sets 
$\{X_1,X_2\}$ and $\{X_3,X_4\}$ 
then by assuming conditional independence between the sets we obtain
\begin{equation}
\Pr_{\textrm{AnJE}}(x_1,x_2,x_3,x_4\mid y) = \Pr(x_1,x_2 \mid y)\Pr(x_3,x_4 \mid y). \nonumber
\end{equation}
Let $\Psi^{\mathcal{A}}_n$ be the set of all partitions of $\mathcal{A}$ such that $\forall_{\partition\in\Psi^{\mathcal{A}}_n}\forall_{\alpha\in\partition}|\alpha|=n$. For convenience we assume that $|\mathcal{A}|$ is a multiple of $n$.
Let $\partitions$ be a subset of $\Psi^{\mathcal{A}}_n$ that includes each set of $n$ attributes once,
\begin{equation}
\partitions\subseteq\Psi^{\mathcal{A}}_n:\forall_{\alpha\in{\mathcal{A}\choose n}}\left|\{\partition\in\partitions:\alpha\in\partition\}\right|=1. \nonumber
\end{equation}
The AnJE model is the geometric mean of the set of n-joint estimators for the partitions $Q \in \partitions$.

The AnJE estimate of conditional likelihood on a per-datum-basis can be written as:
\begin{eqnarray}\label{eq:ANJE}
\Pr_{\textrm{AnJE}}(y\mid\attvals) & \propto & \Pr(y) \P_{\textrm{AnJE}}(\obj|y)\nonumber\\
                   & \propto & \Pr(y)\prod_{\alpha\in{\mathcal{A}\choose n}}\Pr(x_\alpha\mid y)^{\frac{(n{-}1)!(a{-}n)}{(a{-}1)}}. 
\end{eqnarray}
This is derived as follows.
Each $\partition$ is of size $s=a/n$.  
There are $a\choose n$ attribute-value $n$-tuples.  Each must occur in exactly one partition, so the number of partitions must be 
\begin{align}
%m&={a\choose n}/s, \nonumber \\
%&=n{a\choose n}/a, \nonumber \\
%&=nN!/(n![a-n]!a), \nonumber \\
p ={a\choose n}/s = \frac{(a-1)!}{(n-1)!(a-n)!} \label{eq_anjeweights}.
\end{align}
The geometric mean of all the AnJE models is thus 
\begin{align}
\Pr_{\textrm{AnJE}}(\attvals\mid y)&=\sqrt[p]{\prod_{\alpha\in{\attvals\choose a}}\Pr(x_\alpha\mid y)}, \nonumber \\
%&=\prod_{\alpha\in{\attvals\choose a}}\Pr(\alpha\mid y)^{1/M}, \nonumber \\
&=\prod_{\alpha\in{\attvals\choose a}}\Pr(x_\alpha\mid y)^{(n{-}1)!(a{-}n)!/(a{-}1)!}.\label{eq:ANJEx|y}
\end{align}
%Substituting (\ref{eq:ANJEx|y}) into Bayes rule gives us (\ref{eq:ANJE}).  
Using Equation~\ref{eq:ANJE}, we can write the $\log$ of $\P(y\mid\attvals)$ as:
\begin{eqnarray}
\log \Pr_{\textrm{AnJE}}(y\mid\attvals) & \propto & \log \Pr(y) + \frac{(n{-}1)!(a{-}n)}{(a{-}1)!} \sum_{\alpha\in{\attvals\choose a}} \log \Pr(x_\alpha\mid y). 
\end{eqnarray}

%%%%%%%%%%%%%%%%%%%%%%%%%%%%%%%%%%%%%%%%%%%%%%%%%%%%%%%%%%%%%%%%%%%%%%%%%%%%%%%%%%%%%%%%%%%%%%%%%%%%%%%%
\subsection{\DBL^n} \label{sec_wanje}
%%%%%%%%%%%%%%%%%%%%%%%%%%%%%%%%%%%%%%%%%%%%%%%%%%%%%%%%%%%%%%%%%%%%%%%%%%%%%%%%%%%%%%%%%%%%%%%%%%%%%%%%

%Note that AnJE provides us with the generative counterpart that we need for \LR^n.  
%We will show that \DBL^n learns a model of the same form as AnJE but incorporates both generatively and discriminatively learned parameters. 
%In this section, we will introduce our proposed \DBL^n algorithm.

It can be seen that AnJE is a simple model that places the  weight  defined in Equation~\ref{eq_anjeweights} on all feature subsets in the ensemble. The main advantage of this weighting scheme is that it requires no optimization, making AnJE learning extremely efficient. All that is required for training is to calculate the counts from the data.
However, the disadvantage  AnJE  is its inability to perform any form of discriminative learning. 
Our proposed algorithm, \DBL^n uses AnJE to precondition $\LR^n$ by placing weights on all probabilities in Equation~\ref{eq:BNx} and learning these weights by optimizing the conditional-likelihood\footnote{One can initialize these weights with weights in Equation~\ref{eq_anjeweights} for faster convergence.}.  
One can re-write AnJE models with this parameterization as:
\begin{align} \label{eq_wanjeCLL}
\P_{\textrm{DBL}}(y\mid\obj)  = \exp \leftBig(&\weight_y \log\P(y) + \sum_{\alpha\in{\mathcal{A}\choose n}} \weight_{y,\alpha,x_\alpha} \log \P(x_\alpha\mid y) -\nonumber\\
             &  \log\! \sum_{c\in\Omega_Y}\!\! \exp \leftBig( \weight_c\log\P(c) +\!\!\! \sum_{\alpha^*\in{\mathcal{A}\choose n}}\!\! \weight_{c,\alpha^*,x_{\alpha^*}}\log \P(x_{\alpha^*}\mid c) \Big) \Big).
\end{align}
Note that we can compute the likelihood and class-prior probabilities using either MLE or MAP. Therefore, we can write Equation~\ref{eq_wanjeCLL} as:
\begin{align} \label{eq_wanjeCLL2}
\log \P_{\textrm{DBL}}(y\mid\obj)  =& \weight_y \log \pi_y +\! \sum_{\alpha\in{\mathcal{A}\choose n}}\! w_{y,\alpha,x_\alpha} \log \theta_{x_\alpha\mid y} - \nonumber \\
                   & \log \sum_{c\in\Omega_Y}\! \exp \leftBig( w_c \log \pi_c +\! \sum_{\alpha^*\in{\mathcal{A}\choose n}}\! w_{c,\alpha^*,x_{\alpha^*}} \log \theta_{x_{\alpha^*}\mid c} \Big).
\end{align} 
Assuming a Dirichlet prior, a MAP estimate of $\P(y)$ is $\pi_y$ which  equals: 
\begin{eqnarray}
\frac{\#_y + m/|\mathcal{Y}|}{t + m}, \nonumber
\end{eqnarray} 
where $\#_y$ is the number of instances in the dataset with class $y$ and $t$ is the total number of instances, and $m$ is the smoothing parameter. 
We will set $m = 1$ in this work.
Similarly, a MAP estimate of $\P(x_\alpha\mid y)$ is $\theta_{x_\alpha|c}$ which equals:
\begin{eqnarray} 
\frac{\#_{x_\alpha,y} + m/| x_\alpha |}{\#_y + m}, \nonumber
\end{eqnarray}
 where $\#_{x_\alpha,y}$ is the number of instances in the dataset with class $y$ and attribute values $x_\alpha$. 

\DBL^n computes weights by optimizing  CLL. Therefore, one can compute the gradient of Equation~\ref{eq_wanjeCLL2} with-respect-to weights and rely on gradient descent based methods to find the optimal value of these weights. 
Since we do not want to be stuck in local minimums, a natural question to ask is whether the resulting objective function is convex~\cite{Boyd:2008:CO}. 
It turns out that the objective function of \DBL^n is indeed convex.
% .......................................
%\begin{theorem} \label{th:globalmin}
\cite{Roos2005} proved that an objective function of the form $\sum_{\obj \in \data} \log \P_\bn(y|\obj)$, optimized by any conditional Bayesian network model is convex if and only if the structure $\graph$ of the Bayesian network $\bn$ is perfect, that is, all the nodes in $\graph$ are moral nodes.
%\end{theorem}
% .......................................
\DBL^n is a geometric mean of several sub-models where each sub-model models $\lfloor \frac{a}{n} \rfloor$ interactions each conditioned on the class attribute. 
Each sub-model has a structure that is perfect.
Since, the product of two convex objective function leads to a convex function, one can see that \DBL^n's optimization function will also lead to a convex objective function.

Let us first calculate the gradient of Equation~\ref{eq_wanjeCLL2} with-respect-to weights associated with $\pi_y$. We can write:
\begin{eqnarray} \label{eq_wanjeGrad1}
\frac{\partial \log \P(y\|\obj)}{\partial w_{y}}  & = &  \mathbf{1}_{y} \log \pi_y 
                   -\frac{\pi_y^{w_y}  \log \pi_y \prod_{\alpha\in{\mathcal{A}\choose n}} \theta_{x_\alpha\mid y}^{w_{y,\alpha,x_\alpha}}}
                         {\sum_{c\in\Omega_Y}\! \pi_c^{w_c}  \prod_{\alpha^*\in{\mathcal{A}\choose n}}\! \theta_{x_\alpha^*\mid c}^{w_{c,\alpha^*,x_{\alpha^*}}} } \nonumber \\
& = & \left(\mathbf{1}_{y} - \P(y|\obj) \right) \log \pi_y,
\end{eqnarray}
where $\mathbf{1}_{y}$ denotes an indicator function that is $1$ if derivative is taken with-respect-to class $y$ and $0$ otherwise.
Computing the gradient with-respect-to weights associated with $\theta_{x_\alpha|y}$ gives:
\begin{eqnarray} \label{eq_wanjeGrad2}
\frac{\partial \log \P(y\|\obj)}{\partial w_{y,\alpha,x_{\alpha}}} & = & \mathbf{1}_{y} \mathbf{1}_{\alpha} \log \theta_{x_\alpha|y} - 
    \frac{\pi_y^{w_y} \prod_{\alpha\in{\mathcal{A}\choose n}} \theta_{x_\alpha\mid y}^{w_{y,\alpha,x_{\alpha}}} \mathbf{1}_{\alpha} \log \theta_{x_\alpha|y}}
         {\sum_{c\in\Omega_Y}\! \pi_c^{w_c}  \prod_{\alpha^*\in{\mathcal{A}\choose n}}\! \theta_{x_{\alpha^*}\mid c}^{w_{c,\alpha^*,x_{\alpha^*}}} }  \nonumber \\
& = & \left(\mathbf{1}_{y} - \P(y|\obj) \right) \mathbf{1}_{\alpha} \log \theta_{x_\alpha|y},
\end{eqnarray}
where $\mathbf{1}_{\alpha}$ and $\mathbf{1}_{y}$ denotes an indicator function that is $1$ if the derivative is taken with-respect-to attribute set $\alpha$ (respectively, class $y$) and $0$ otherwise.

% -----------------------------------------
\subsection{Alternative Parameterization}
% -----------------------------------------

Let us reparameterize \DBL^n such that: 
\begin{equation} \label{eq_reparam}
\beta_y = w_y \log \pi_y, \;\;\;\; \textrm{and} \;\;\;\; \beta_{y,\alpha,x_{\alpha}} = w_{y,\alpha,x_{\alpha}} \log \theta_{x_\alpha \mid y}.
\end{equation}
Now, we can re-write Equation~\ref{eq_wanjeCLL2} as:
\begin{align} \label{eq_danjeCLL2}
\log \P_{LR}(y\mid\obj) & = \beta_y + \sum_{\alpha\in{\mathcal{A}\choose n}} \beta_{y,\alpha,x_{\alpha}} - \log \sum_{c\in\Omega_Y} \exp \leftBig( \beta_c + \sum_{\alpha^*\in{\mathcal{A}\choose n}} \beta_{c,\alpha^*,x_{\alpha^*}} \Big).
\end{align}
%\gw{I removed `$-$' from the first line of (\ref{eq_danjeCLL2}). Please check.}
It can be seen that this leads to Equation~\ref{eq_LRk}. We call this parameterization \LR^n.

Like \DBL^n, \LR^n also leads to a convex optimization problem, and, therefore, its parameters can also be optimized by simple gradient decent based algorithms. 
Let us compute the gradient of objective function in Equation~\ref{eq_danjeCLL2} with-respect-to $\beta_y$. In this case, we can write:
\begin{eqnarray} \label{eq_danjeGrad1}
\frac{\partial \log \P(y\|\obj)}{\partial \beta_{y}} & = & \left(\mathbf{1} - \P(y|\obj) \right).
\end{eqnarray}
Similarly, computing gradient with-respect-to $\beta_{\alpha|c}$, we can write:
\begin{eqnarray} \label{eq_danjeGrad2}
\frac{\partial \log \P(y\|\obj)}{\partial \beta_{y,\alpha,x_{\alpha}}} & = & \left(\mathbf{1} - \P(y|\obj) \right) \mathbf{1}_{\alpha}.
\end{eqnarray}

% ---------------------------------------------------------------------------------
\subsection{Comparative analysis of \DBL^n and \LR^n} \label{subsec_wanjevsdanjee}
% ---------------------------------------------------------------------------------

It can be seen that the two models are actually equivalent and each is a re-parameterization of the other. However, there are subtle distinctions between the two..
The most important distinction is the utilization of MAP or MLE probabilities in \DBL^n. Therefore, \DBL^n is a two step learning algorithm:
\begin{itemize}
\item Step 1 is the optimization of log-likelihood of the data ($\log \P(y,\obj)$) to obtain the estimates of the prior and likelihood probabilities. One can view this step as of \emph{generative learning}.
\item Step 2 is the introduction of weights on these probabilities and learning of these weights by maximizing CLL ($\P(y\|\obj)$) objective function. This step can be interpreted as \emph{discriminative learning}.
\end{itemize}
\DBL^n employs generative-discriminative learning as opposed to only discriminative learning by \LR^n.

One can expect a similar bias-variance profile and a very similar classification performance as both models will converge to a similar point in the optimization space, the only difference in the final parameterization being due to recursive descent being terminated before absolute optimization. However, the rate of convergence of the two models can be very different. 
\cite{zaidi14} show that for NB, such \DBL^n style parameterization with generative-discriminative learning can greatly speed-up convergence relative to only discriminative training. Note, discriminative training with NB as the graphical model is vanilla LR. 
We expect to see the same trend in the convergence performance of \DBL^n and \LR^n.

Another distinction between the two models becomes explicit if a regularization penalty is added to the objective function. One can see that in case of \DBL^n, optimizing parameters towards $1$ will effectively pull parameters back towards the generative training estimates. For smaller datasets, one can expect to obtain better performance by using a large regularization parameter and pulling estimates back towards $1$. However, one cannot do this for \LR^n. Therefore, \DBL^n models can very elegantly combine generative discriminative parameters.

An analysis of the gradient of \DBL^n in Equation~\ref{eq_wanjeGrad1} and~\ref{eq_wanjeGrad2} and that of \LR^n in Equation~\ref{eq_danjeGrad1} and~\ref{eq_danjeGrad2} also reveals an interesting comparison. We can write \DBL^n's gradients in terms of \LR^n's gradient as follows:
\begin{eqnarray}
\frac{\partial \log \P(y\|\obj)}{\partial w_{y}} & = & \frac{\partial \log \P(y\|\obj)}{\partial \beta_{y}} \log \pi_y, \nonumber \\
\frac{\partial \log \P(y\|\obj)}{\partial w_{y,\alpha,x_{\alpha}}} & = & \frac{\partial \log \P(y\|\obj)}{\partial \beta_{y,\alpha,x_{\alpha}}} \log \theta_{x_\alpha|y}. \nonumber
\end{eqnarray}
It can be seen that \DBL^n has the effect of re-scaling \LR^n's gradient by the log of the conditional probabilities. 
We conjecture that such re-scaling has the effect of pre-conditioning the parameter space and, therefore, will lead to faster convergence.

%%%%%%%%%%%%%%%%%%%%%%%%%%%%%%%%%%%%%%%%%%%%%%%%%%%%%%%%%%%%%%%%%%%%%%%%%%%%%%%%%%%%%%%%%%%%%%%%%%%%%%%%
\section{Related Work} \label{sec_rw}
%%%%%%%%%%%%%%%%%%%%%%%%%%%%%%%%%%%%%%%%%%%%%%%%%%%%%%%%%%%%%%%%%%%%%%%%%%%%%%%%%%%%%%%%%%%%%%%%%%%%%%%%

\begin{figure} \center
    	\begin{subfigure}[b]{0.3\textwidth}
        \centering
        \resizebox{\linewidth}{!}{
            \begin{tikzpicture}
    		[
      		observed/.style={minimum size=15pt,circle,draw=blue!50,fill=blue!20},
      		unobserved/.style={minimum size=15pt,circle,draw},
      		hyper/.style={minimum size=1pt,circle,fill=black},
      		post/.style={->,>=stealth',semithick},
    		]

		% Level 1
		\node (x1) [observed] at (1,-1) {$x_1$};    
    		\node (x2) [observed] at (3,-1) {$x_2$};
    		\node (x3) [observed] at (5,-1) {$x_3$};
    		\node (x4) [observed] at (7,-1) {$x_4$}; 
		\node (y) [unobserved] at (4,2) {$y$};
    		
    		\draw[thick,->] (y) -- (x1);
    		\draw[thick,->] (y) -- (x2);
    		\draw[thick,->] (y) -- (x3);
    		\draw[thick,->] (y) -- (x4);
		
		\path[->] (x1) edge (x2);
		\path[->] (x1) edge [bend right] (x3);		
		\path[->] (x1) edge [bend right] (x4);

            \end{tikzpicture}
        }
        %\caption{Sub-model 1}
        %\label{fig_a1de1}
    \end{subfigure}
    	\begin{subfigure}[b]{0.3\textwidth}
        \centering
        \resizebox{\linewidth}{!}{
            \begin{tikzpicture}
    		[
      		observed/.style={minimum size=15pt,circle,draw=blue!50,fill=blue!20},
      		unobserved/.style={minimum size=15pt,circle,draw},
      		hyper/.style={minimum size=1pt,circle,fill=black},
      		post/.style={->,>=stealth',semithick},
    		]

		% Level 1
     		\node (x1) [observed] at (1,-1) {$x_1$};    
    		\node (x2) [observed] at (3,-1) {$x_2$};
    		\node (x3) [observed] at (5,-1) {$x_3$};
    		\node (x4) [observed] at (7,-1) {$x_4$}; 
		\node (y) [unobserved] at (4,2) {$y$};
    		
    		\draw[thick,->] (y) -- (x1);
    		\draw[thick,->] (y) -- (x2);
    		\draw[thick,->] (y) -- (x3);
    		\draw[thick,->] (y) -- (x4);
		
		\path[->] (x2) edge (x1);
		\path[->] (x2) edge [bend right] (x3);		
		\path[->] (x2) edge [bend right] (x4);

            \end{tikzpicture}
        }
        %\caption{Sub-model 2}
        %\label{fig_a1de2}
    \end{subfigure}

    	\begin{subfigure}[b]{0.3\textwidth}
        \centering
        \resizebox{\linewidth}{!}{
            \begin{tikzpicture}
    		[
      		observed/.style={minimum size=15pt,circle,draw=blue!50,fill=blue!20},
      		unobserved/.style={minimum size=15pt,circle,draw},
      		hyper/.style={minimum size=1pt,circle,fill=black},
      		post/.style={->,>=stealth',semithick},
    		]

		% Level 1
     		\node (x1) [observed] at (1,-1) {$x_1$};    
    		\node (x2) [observed] at (3,-1) {$x_2$};
    		\node (x3) [observed] at (5,-1) {$x_3$};
    		\node (x4) [observed] at (7,-1) {$x_4$}; 
		\node (y) [unobserved] at (4,2) {$y$};
    		
    		\draw[thick,->] (y) -- (x1);
    		\draw[thick,->] (y) -- (x2);
    		\draw[thick,->] (y) -- (x3);
    		\draw[thick,->] (y) -- (x4);
		
		\path[->] (x3) edge (x2);
		\path[->] (x3) edge [bend right] (x1);		
		\path[->] (x3) edge [bend right] (x4);

            \end{tikzpicture}
        }
        %\caption{Sub-model 3}
        %\label{fig_a1de3}
    \end{subfigure}
    	\begin{subfigure}[b]{0.3\textwidth}
        \centering
        \resizebox{\linewidth}{!}{
            \begin{tikzpicture}
    		[
      		observed/.style={minimum size=15pt,circle,draw=blue!50,fill=blue!20},
      		unobserved/.style={minimum size=15pt,circle,draw},
      		hyper/.style={minimum size=1pt,circle,fill=black},
      		post/.style={->,>=stealth',semithick},
    		]

		% Level 1
     		\node (x1) [observed] at (1,-1) {$x_1$};    
    		\node (x2) [observed] at (3,-1) {$x_2$};
    		\node (x3) [observed] at (5,-1) {$x_3$};
    		\node (x4) [observed] at (7,-1) {$x_4$}; 
		\node (y) [unobserved] at (4,2) {$y$};
    		
    		\draw[thick,->] (y) -- (x1);
    		\draw[thick,->] (y) -- (x2);
    		\draw[thick,->] (y) -- (x3);
    		\draw[thick,->] (y) -- (x4);
		
		\path[->] (x4) edge (x3);
		\path[->] (x4) edge [bend right] (x1);		
		\path[->] (x4) edge [bend right] (x2);

            \end{tikzpicture}
        }
        %\caption{Sub-model 4}
        %\label{fig_a1de4}
    \end{subfigure}

    \caption{Sub-models in an AnDE model with $n=2$ and with four attributes.} 
    \label{fig_A1DE}
\end{figure}
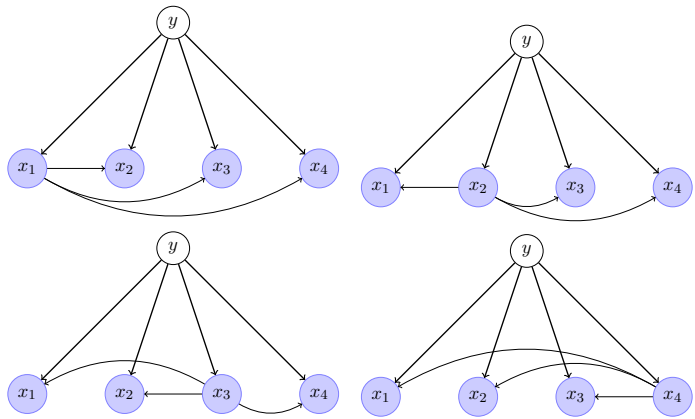

\emph{Averaged $n$-Dependent Estimators} (AnDE) is the inspiration for AnJE.
An AnDE model is the arithmetic mean of all Bayesian Network Classifiers in each of which all attributes depend on the class and the some $n$ attributes. A simple depiction of A1DE in graphical form in shown in Figure~\ref{fig_A1DE}. 
% ----------------------------------------------------------
% ----------------------------------------------------------
There are ${a \choose n}$ possible combination of attributes that can be used as parents, producing ${a\choose n}$ sub-models which are combined by averaging.

AnDE and AnJE both use simple generative learning, merely the counting the relevant \emph{sufficient statistics} from the data.
Second, both have only one tweaking parameter: $n$ -- that controls the bias-variance trade-off. Higher values of $n$ leads to low bias and high variance and vice-versa.
%And, third, both are an ensemble of $\dbinom{a}{n}$ sub-models.
%Now, more importantly, let us consider the distinctions between the two models.
%First, AnDE is based on arithmetic averaging of its sub-models whereas, AnJE is based on geometric averaging.
%Second, probabilities in each sub-model of AnDE are of the form: $\P(\obj_j) \P(x_i\|\obj_j,y)$, where $\obj_j$ is a CLIQUE-SET of size $n$.
%In AnJE, of course, each sub-model probability is of the form: $\P(\obj_j\|y)$.

It is important not to confuse the equivalence (in terms of the level of interactions they model) of AnJE and AnDE models. That is, the following holds:
\begin{eqnarray}
f(\textrm{A2JE}) & = & f(\textrm{A1DE}), \nonumber \\
f(\textrm{A3JE}) & = & f(\textrm{A2DE}), \nonumber \\
\vdots \;\;\;\; & = & \;\;\;\; \vdots \nonumber \\
f(\textrm{AnJE}) & = & f(\textrm{A(n-1)DE}), \nonumber
\end{eqnarray}
where $f(.)$ is a function that returns the number of interactions that the algorithm models.
Thus, an AnJE model uses the same core statistics as an A(n-1)DE model. 
At training time, AnJE and A(n-1)DE must learn the same information from the data. However, at classification time, each of these statistics is accessed once by AnJE and $n$ times by A(n-1)DE, making AnJE more efficient.
However, as we will show, it turns out that AnJE's use of the geometric mean results in a more biased estimator than than the arithmetic mean used by AnDE. As a result, in practice, an AnJE model is less accurate than the equivalent AnDE model.

%Given, geometric mean is a poor estimate of the mean, one can expect equivalent AnDE model performing better than AnJE model.

However, due to the use of arithmetic mean by AnDE, its weighted version would be much more difficult to optimize than AnJE, as transformed to log space it does not admit to a simple linear model.

A work relevant to \DBL^n is that of~\cite{GreinerExtLR2004,Greiner:2002:SELR}. The proposed technique in these papers named \emph{ELR} has a number of similar traits with \DBL^n. For example, the parameters associated with a Bayesian network classifier (naive Bayes and TAN) are learned by optimizing the CLL. Both \emph{ELR} and \DBL^n can be viewed as feature engineering frameworks. An ELR (let us say with TAN structure) model is a subset of \DBL^2 models. The comparison of \DBL^n with ELR is not the goal of this work. But in our preliminary results, \DBL^n produce models of much lower bias that ELR (TAN). Modelling higher-order interactions is also an issue with \emph{ELR}. One could learn a Bayesian network structure and create features based on that and then use \emph{ELR}. But several restrictions needs to be imposed on the structure, that is, it has to fulfill the property of perfectness, to make sure that it leads to a convex optimization problem.
With \DBL^n, as we discussed in Section~\ref{sec_wanje}, there are no restrictions. Need less to say, \emph{ELR} is neither broad nor deep.
Some related ideas to \emph{ELR} are also explored in~\cite{Pernkopf:2005:DGSPL,Pernkopf:2009:DPLBNC,Jiang:2008:DPLBN}.

Several 

%%%%%%%%%%%%%%%%%%%%%%%%%%%%%%%%%%%%%%%%%%%%%%%%%%%%%%%%%%%%%%%%%%%%%%%%%%%%%%%%%%%%%%%%%%%%%%%%%%%%%%%%
\section{Experiments}\label{sec_exp}
%%%%%%%%%%%%%%%%%%%%%%%%%%%%%%%%%%%%%%%%%%%%%%%%%%%%%%%%%%%%%%%%%%%%%%%%%%%%%%%%%%%%%%%%%%%%%%%%%%%%%%%%

In this section, we compare and analyze the performance of our proposed algorithms and related methods on $77$ natural domains from the UCI repository of machine learning~\citep{UCI:2005:MLR}.
% -----------------------------------------------------------------------
\begin{table}[bt] \center \tiny
\setlength{\topmargin}{1pt} \setlength{\tabcolsep}{0.3pt}\renewcommand{\arraystretch}{1}
\begin{center}
%\begin{tabular}{llrrcclrllrrr}
\tabcolsep=1.0pt
\begin{tabular}{lrrrrlrrrrrlrrrrrr}
\hline
\bf Domain & \bf Case & \bf Att & \bf Class && \bf Domain & \bf Case & \bf Att & \bf Class  \\
\hline
\bf Kddcup		&5209000	&41	&40	&&Vowel&990&14&11 \\
\bf Poker-hand		&1175067	&10	&10	&&Tic-Tac-ToeEndgame&958&10&2 \\
\bf MITFaceSetC		&839000		&361	&2	&&Annealing&898&39&6 \\
\bf Covertype		&581012		&55	&7	&&Vehicle&846&19&4 \\
\bf MITFaceSetB		&489400		&361	&2	&&PimaIndiansDiabetes&768&9&2 \\
\bf MITFaceSetA		&474000		&361	&2	&&BreastCancer(Wisconsin)&699&10&2 \\
\bf Census-Income(KDD)	&299285&40&2&&CreditScreening&690&16&2 \\
\bf Localization&164860&7&3&&BalanceScale&625&5&3 \\
Connect-4Opening&67557&43&3&&Syncon&600&61&6 \\
Statlog(Shuttle)&58000&10&7&&Chess&551&40&2 \\
Adult&48842&15&2&&Cylinder&540&40&2 \\
LetterRecognition&20000&17&26&&Musk1&476&167&2 \\
MAGICGammaTelescope&19020&11&2&&HouseVotes84&435&17&2 \\
Nursery&12960&9&5&&HorseColic&368&22&2 \\
Sign&12546&9&3&&Dermatology&366&35&6 \\
PenDigits&10992&17&10&&Ionosphere&351&35&2 \\
Thyroid&9169&30&20&&LiverDisorders(Bupa)&345&7&2 \\
Pioneer&9150&37&57&&PrimaryTumor&339&18&22 \\
Mushrooms&8124&23&2&&Haberman'sSurvival&306&4&2 \\
Musk2&6598&167&2&&HeartDisease(Cleveland)&303&14&2 \\
Satellite&6435&37&6&&Hungarian&294&14&2 \\
OpticalDigits&5620&49&10&&Audiology&226&70&24 \\
PageBlocksClassification&5473&11&5&&New-Thyroid&215&6&3 \\
Wall-following&5456&25&4&&GlassIdentification&214&10&3 \\
Nettalk(Phoneme)&5438&8&52&&SonarClassification&208&61&2 \\
Waveform-5000&5000&41&3&&AutoImports&205&26&7 \\
Spambase&4601&58&2&&WineRecognition&178&14&3 \\
Abalone&4177&9&3&&Hepatitis&155&20&2 \\
Hypothyroid(Garavan)&3772&30&4&&TeachingAssistantEvaluation&151&6&3 \\
Sick-euthyroid&3772&30&2&&IrisClassification&150&5&3 \\
King-rook-vs-king-pawn&3196&37&2&&Lymphography&148&19&4 \\
Splice-junctionGeneSequences&3190&62&3&&Echocardiogram&131&7&2 \\
Segment&2310&20&7&&PromoterGeneSequences&106&58&2 \\
CarEvaluation&1728&8&4&&Zoo&101&17&7 \\
Volcanoes&1520&4&4&&PostoperativePatient&90&9&3 \\
Yeast&1484&9&10&&LaborNegotiations&57&17&2 \\
ContraceptiveMethodChoice&1473&10&3&&LungCancer&32&57&3 \\
German&1000&21&2&&Contact-lenses&24&5&3 \\
LED&1000&8&10&&&&& \\
\hline
\end{tabular}
\end{center}
\caption{\small Details of Datasets}\label{UCIDatasets}
\end{table}
% -----------------------------------------------------------------------
The experiments are conducted on the datasets described in Table~\ref{UCIDatasets}. There are a total of $77$ datasets, $40$ datasets with less than $1000$ instances, $21$ datasets with instances between $1000$ and $10000$, and $16$ datasets with more than $10000$ instances. 
There are $8$ datasets with over $100000$ instances. These datasets are shown in bold font in Table~\ref{UCIDatasets}.
%It is the performance on these datasets that we will be interested in. 

Each algorithm is tested on each dataset using $5$ rounds of $2$-fold cross validation\footnote{Exception is \texttt{MITFaceSetA}, \texttt{MITFaceSetB} and \texttt{Kddcup} where results are reported with $2$ rounds of $2$-fold cross validation.}. 

We compare four different metrics, i.e., 0-1 Loss, RMSE, Bias and Variance\footnote{As discussed in Section~\ref{sec_intro}, the reason for performing bias/variance estimation is that it provides insights into how the learning algorithm will perform with varying amount of data. We expect low variance algorithms to have relatively low error for small data and low bias algorithms to have relatively low error for large data~\citep{Brain2002}.}.

We report Win-Draw-Loss (W-D-L) results when comparing the 0-1 Loss, RMSE, bias and variance of two models. 
A two-tail binomial sign test is used to determine the significance of the results. Results are considered significant if $p \leq 0.05$.  
%We give W-D-L for bias and variance, as we expect relative bias and variance to be relatively constant across different data quantities, whereas we expect relative error to be affected by data quantity and hence do not expect any algorithm to dominate another in error across a wide range of different datasets.

The datasets in Table~\ref{UCIDatasets} are divided into two categories. We call the following datasets \emph{Big} -- \texttt{KDDCup, Poker-hand, USCensus1990, Covertype, MITFaceSetB, MITFaceSetA, Census-income, Localization}.
All remaining datasets are denoted as \emph{Little} in the results.
Due to their size, experiments for most of the \emph{Big} datasets had to be performed in a heterogeneous environment (grid computing) for which CPU wall-clock times are not commensurable.  In consequence, when comparing classification and training time, the following $9$ datasets constitutes \emph{Big} category -- \texttt{Localization, Connect-4, Shuttle, Adult, Letter-recog, Magic, Nursery, Sign, Pendigits}. 

When comparing average results across \emph{Little} and \emph{Big} datasets, we normalize the results with respect to \DBL^2 and present a geometric mean. 

Numeric attributes are discretized by using the Minimum Description Length (MDL) discretization method~\citep{Fayyad1992}. A missing value is treated as a separate attribute value and taken into account exactly like other values.

We employed L-BFGS quasi-Newton methods~\citep{Zhu97} for solving the optimization\footnote{The original L-BFGS implementation of~\citep{Byrd:1995:LMA} from \url{http://users.eecs.northwestern.edu/~nocedal/lbfgsb.html} is used.}.

We used a Random Forest that is an ensemble of $100$ decision trees~\cite{Breiman:RF}. 

Both \DBL^n and \LR^n are $\textrm{L}_2$ regularized. The regularization constant $C$ is not tuned and is set to $10^{-2}$ for all experiments.

The detailed 0-1 Loss and RMSE results on \emph{Big} datasets are also given in Appendix~\ref{sec_discussion}. 

%%%%%%%%%%%%%%%%%%%%%%%%%%%%%%%%%%%%%%%%%%%%%%%%%%%%%%%%%%%%%%%%%
\subsection{\DBL^n vs. AnJE}
%%%%%%%%%%%%%%%%%%%%%%%%%%%%%%%%%%%%%%%%%%%%%%%%%%%%%%%%%%%%%%%%%
A W-D-L comparison of the 0-1 Loss, RMSE, bias and variance of \DBL^n and AnJE on \emph{Little} datasets is shown in Table~\ref{tab_wAnJEvsAnJE}.
We compare \DBL^2 with A2JE and \DBL^3 with A3JE only.
It can be seen that \DBL^n has significantly lower bias but significantly higher variance. 
The 0-1 Loss and RMSE results are not in favour of any algorithm. 
However, on \emph{Big} datasets, \DBL^n wins on 7 out of 8 datasets in terms of both RMSE and 0-1 Loss. 
%\gw{Why have you set a threshold of 0.005? Above you said we were using an alpha of 0.05.} 
The results are not significant since $p$ value of $0.070$ is greater than our set threshold of $0.05$.
One can infer that \DBL^n successfully reduces the bias of AnJE, at the expense of increasing its variance. 
 % --------------------------
\begin{table} \scriptsize
\begin{tabular}{lcccc}
\cline{1-5}
&\multicolumn{2}{c}{\bf \DBL^2 vs. A2JE} & \multicolumn{2}{c}{\bf \DBL^3 vs. A3JE} \\
\cmidrule {2-3}\cmidrule (l){4-5} 
&W-D-L& $p$& W-D-L& $p$ \\
\cline{1-5}
&\multicolumn{4}{c}{\emph{Little} Datasets} \\
\cline{1-5}
Bias	&66/4/5		&\bf $<$0.001	& 58/2/15& \bf $<$0.001\\
Variance&16/3/56	&\bf $<$0.001	& 19/2/54& \bf $<$0.001 \\
0-1 Loss&42/5/28	& 0.119		& 37/3/35& 0.906 \\
RMSE	&37/1/37	& 1.092	& 30/1/44& 0.130 \\
\cline{1-5}
&\multicolumn{4}{c}{\emph{Big} Datasets} \\
\cline{1-5}
0-1 Loss&7/0/1	& 0.070	& 7/0/1& 0.070 \\
RMSE	&7/0/1	& 0.070	& 7/0/1& 0.070 \\
\cline{1-5}
\end{tabular}
\caption{\small Win-Draw-Loss: \DBL^2 vs. A2JE and \DBL^3 vs A3JE. $p$ is two-tail binomial sign test. Results are significant if $p \leq 0.05$.}
\label{tab_wAnJEvsAnJE}
\end{table}
% ---------------------------------

\begin{figure}[t] 
\centering
\includegraphics[width=55mm,height=40mm]{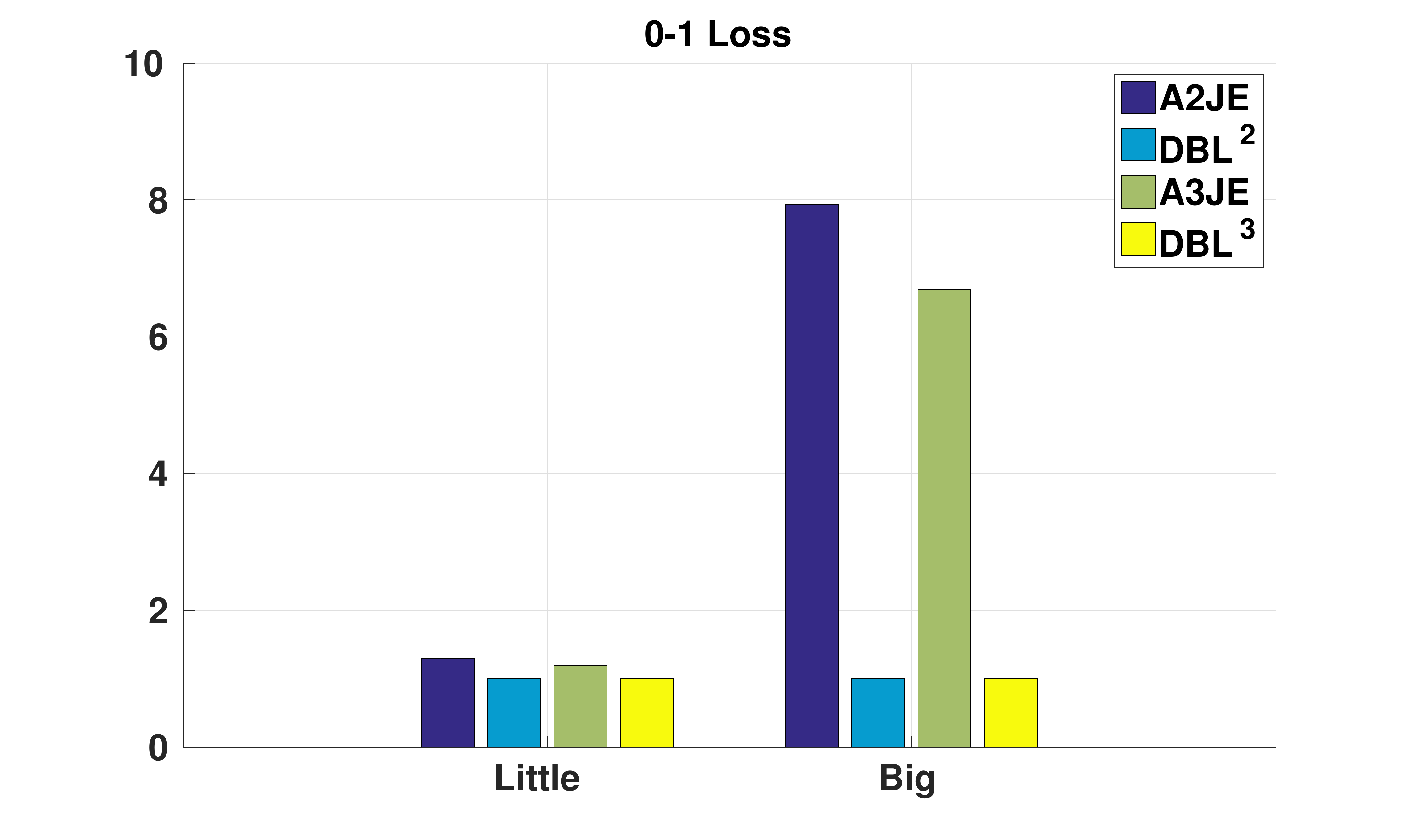}
\includegraphics[width=55mm,height=40mm]{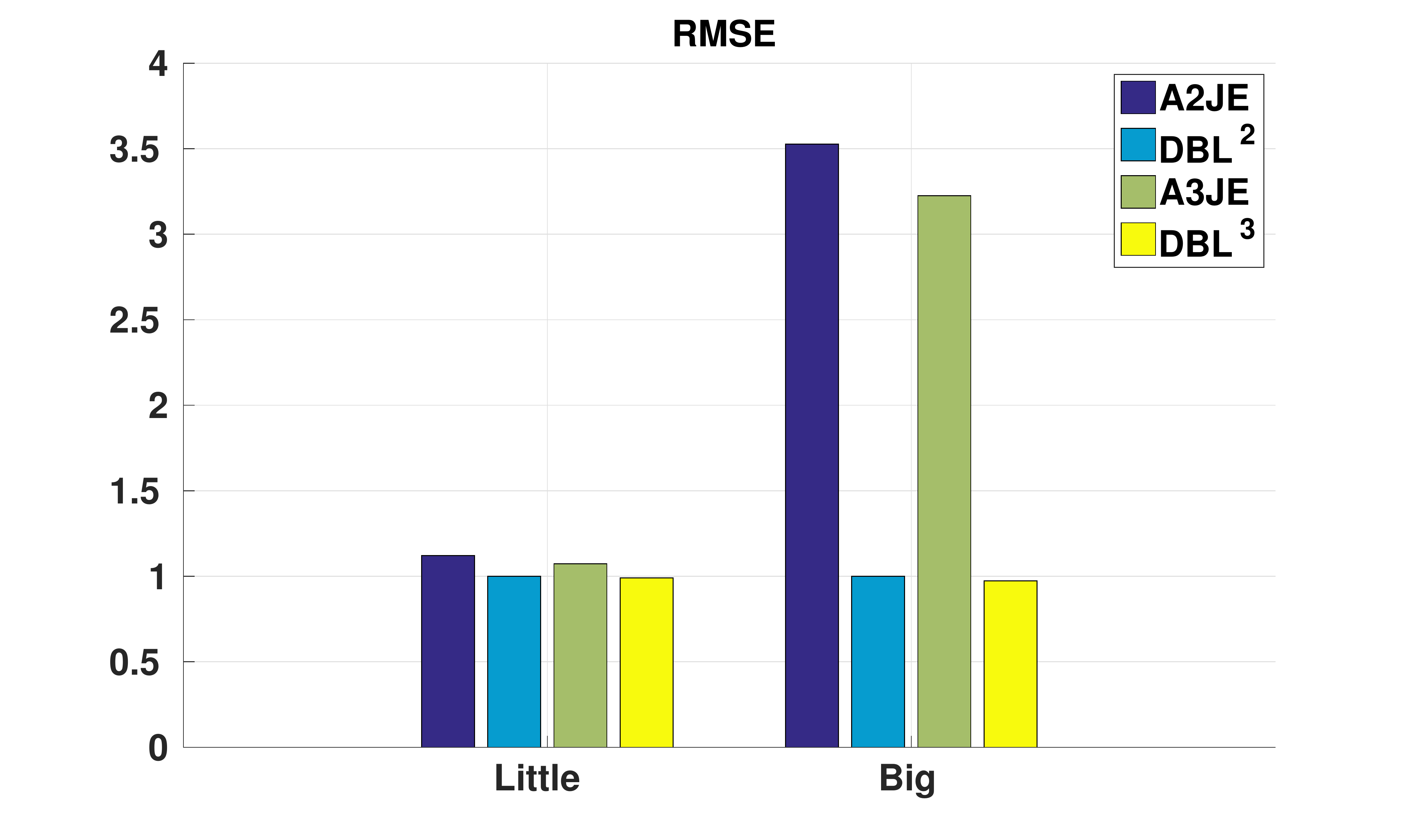}

%\includegraphics[width=55mm,height=40mm]{images2/AnJEvsALR_Bias}
%\includegraphics[width=55mm,height=40mm]{images2/AnJEvsALR_Var}
%\caption{\small Geometric mean of 0-1 Loss (Top Left), RMSE (Top Right), Bias (Bottom Left) and Variance (Bottom Right) performance of \DBL^2, A2JE, \DBL^3 and A3JE for \emph{All} and \emph{Big} datasets.}
\caption{\small Geometric mean of 0-1 Loss (Left), RMSE (Right) performance of \DBL^2, A2JE, \DBL^3 and A3JE for \emph{Little} and \emph{Big} datasets.}
\label{fig_wAnJEvsAnJE}
\end{figure}

\begin{figure}[t] 
\centering
\includegraphics[width=55mm,height=40mm]{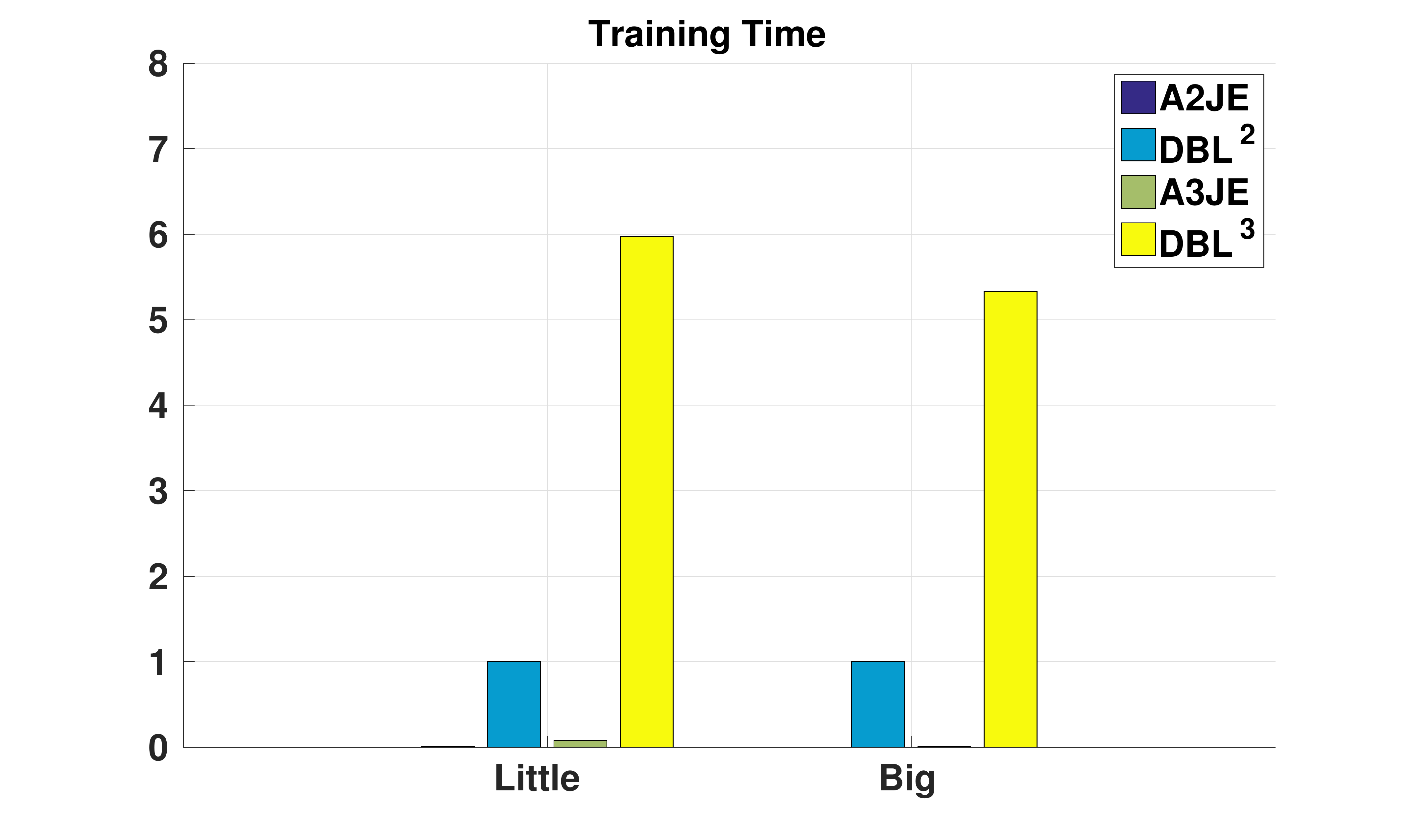}
\includegraphics[width=55mm,height=40mm]{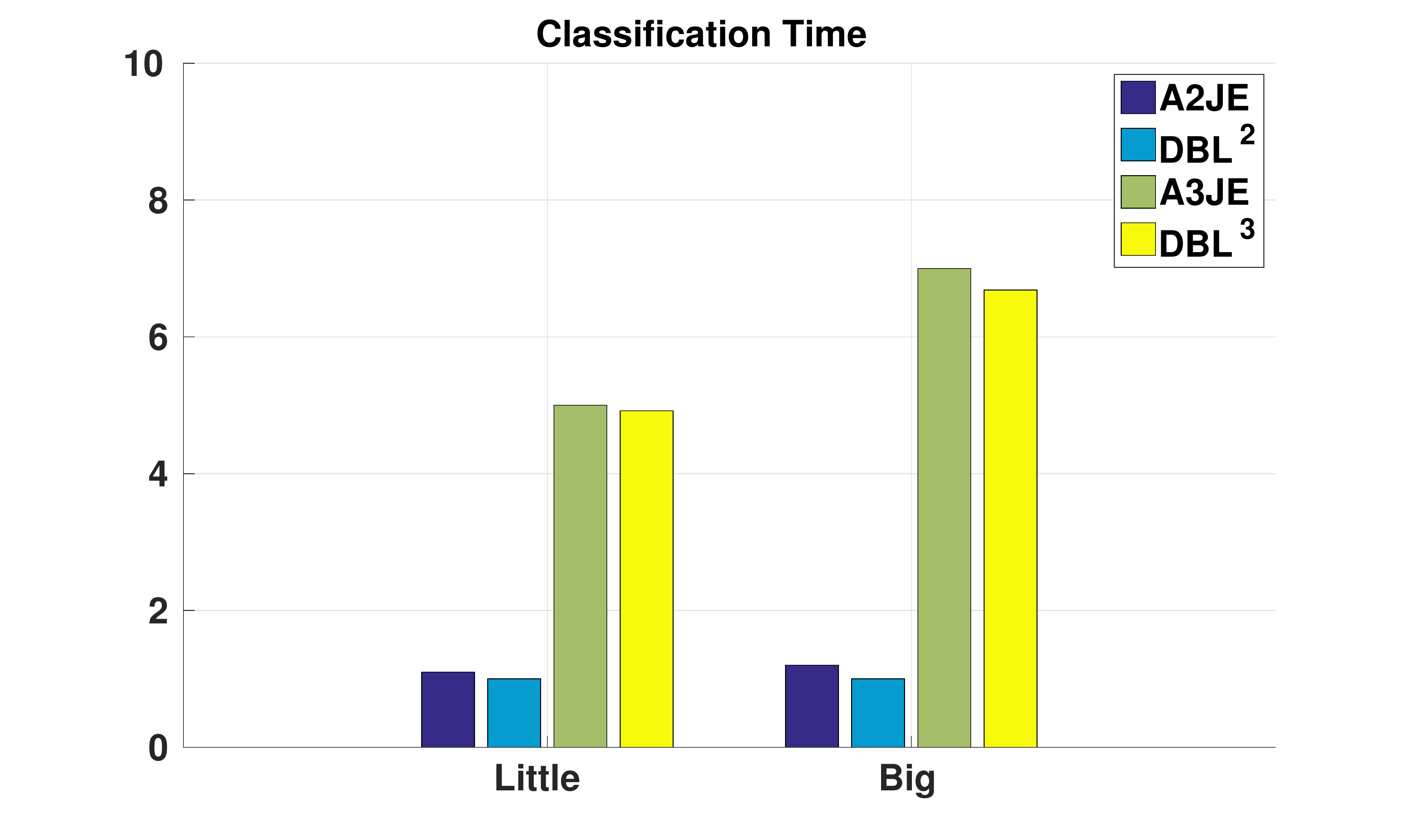}
\caption{\small Geometric mean of Training Time (Left), Classification Time (Right) of \DBL^2, A2JE, \DBL^3 and A3JE for \emph{All} and \emph{Big} datasets.}
\label{fig_wAnJEvsAnJE_Time}
\end{figure}

Normalized 0-1 Loss and RMSE results for both models are shown in Figure~\ref{fig_wAnJEvsAnJE}.
% --------------------------
% --------------------------
It can be seen that \DBL^n has a lower averaged 0-1 Loss and RMSE than AnJE. This difference is substantial when comparing on \emph{Big} datasets. 
The training and classification time of AnJE is, however, substantially lower than \DBL^n as can be seen from Figure~\ref{fig_wAnJEvsAnJE_Time}. This is to be expected as \DBL^n adds discriminative training to AnJE and uses twice the number of parameters at classification time. 
%\gw{Why does Figure 3 show DBL with faster classification time than ANJE?}
% --------------------------
% --------------------------

%%%%%%%%%%%%%%%%%%%%%%%%%%%%%%%%%%%%%%%%%%%%%%%%%%%%%%%%%%%%%%%%%
\subsection{\DBL^n vs. AnDE}
%%%%%%%%%%%%%%%%%%%%%%%%%%%%%%%%%%%%%%%%%%%%%%%%%%%%%%%%%%%%%%%%%

A W-D-L comparison for 0-1 Loss, RMSE, bias and variance results of the two \DBL^n models relative to the corresponding AnDE models are presented in Table~\ref{tab_wAnJEvsAnDE}. 
We compare \DBL^2 with A1DE and \DBL^3 with A2DE only.
It can be seen that \DBL^n has significantly lower bias and significantly higher variance variance than AnDE models.
Recently, AnDE models have been proposed as a fast and effective Bayesian classifiers when learning from large quantities of data~\citep{zaidi12}.
These bias-variance results make $\DBL^n$ a suitable alternative to AnDE when dealing with big data. 
The 0-1 Loss and RMSE results (with exception of RMSE comparison of \DBL^3 vs.\ A2DE) are similar.
% --------------------------
\begin{table} \scriptsize
\begin{tabular}{lcccc}
\cline{1-5}
 &\multicolumn{2}{c}{\bf \DBL^2 vs. A1DE} & \multicolumn{2}{c}{\bf \DBL^3 vs. A2DE} \\
\cmidrule {2-3}\cmidrule (l){4-5} 
&W-D-L& $p$& W-D-L& $p$ \\
\cline{1-5}
&\multicolumn{4}{c}{\emph{Little} Datasets} \\
\cline{1-5}
Bias	&65/3/7		&\bf $<$0.001	& 53/5/17& \bf $<$0.001\\
Variance&21/5/49	&\bf 0.001	& 26/5/44& 0.041 \\
0-1 Loss&42/4/29	& 0.1539	& 39/3/33& 0.556 \\
RMSE	&30/1/44	& 0.130		& 22/1/52& \bf $<$0.001\\
\cline{1-5}
&\multicolumn{4}{c}{\emph{Big} Datasets} \\
\cline{1-5}
0-1 Loss&8/0/0	& 0.007	& 7/0/1& 0.073 \\
RMSE	&7/0/1	& 0.073	& 6/0/2& 0.289 \\
\cline{1-5}
\end{tabular}
\caption{\small Win-Draw-Loss: \DBL^2 vs. A1DE and \DBL^3 vs A2DE. $p$ is two-tail binomial sign test. Results are significant if $p \leq 0.05$.}
\label{tab_wAnJEvsAnDE}
\end{table}
% ---------------------------------

\begin{figure}
\centering
\includegraphics[width=55mm,height=40mm]{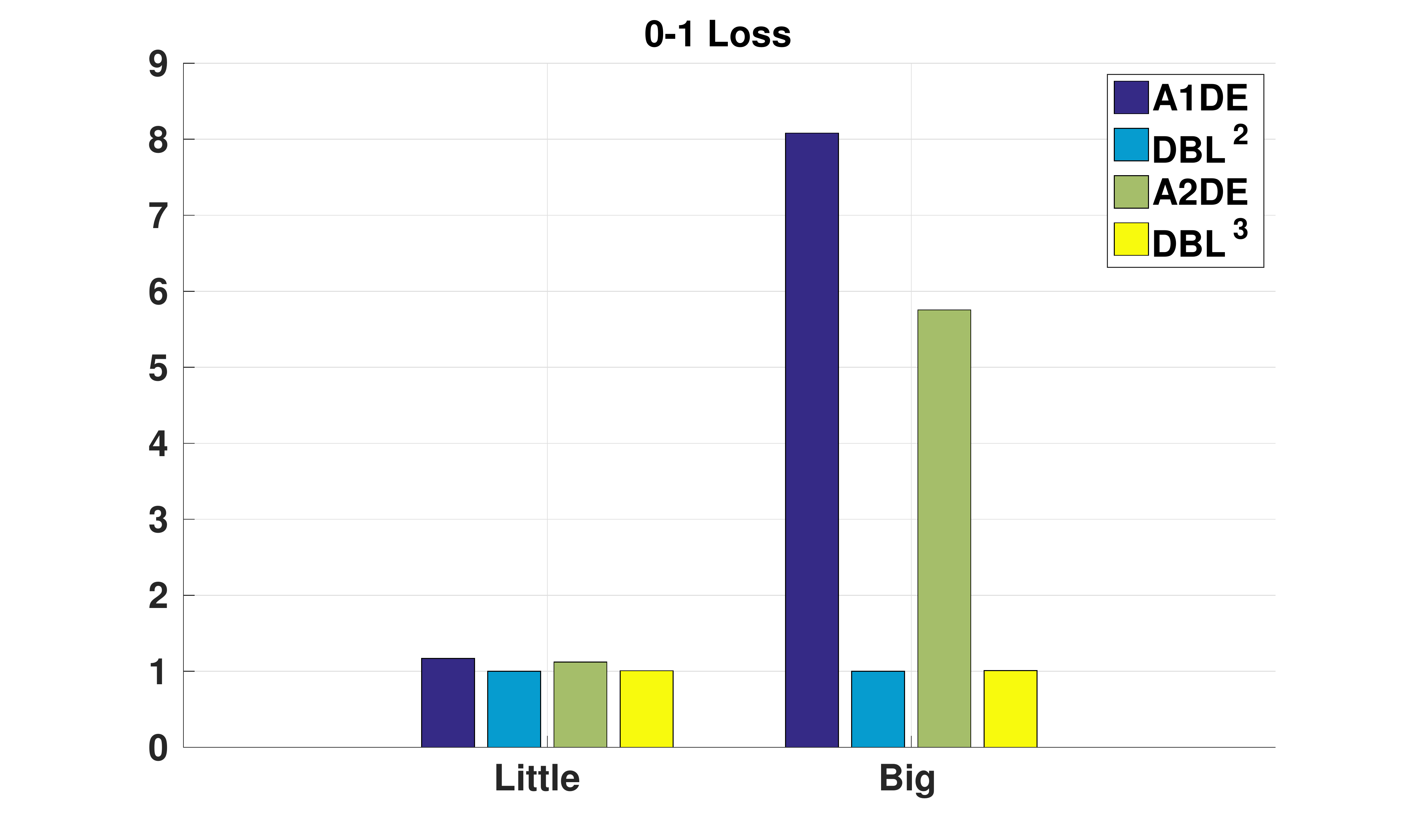}
\includegraphics[width=55mm,height=40mm]{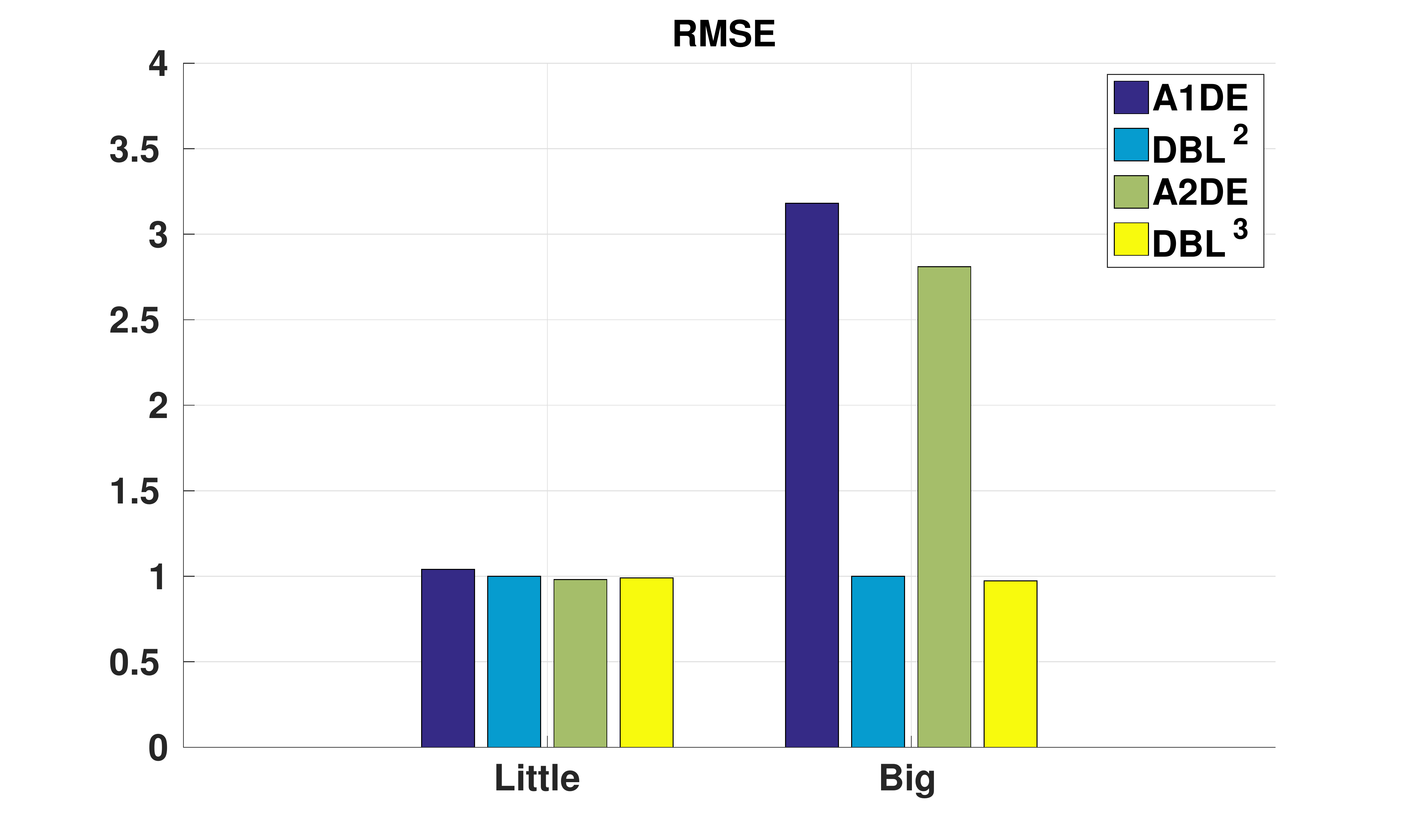}

%\includegraphics[width=55mm,height=40mm]{images2/AnDEvsALR_Bias}
%\includegraphics[width=55mm,height=40mm]{images2/AnDEvsALR_Var}
%\caption{\small Geometric mean of 0-1 Loss (Top Left), RMSE (Top Right), Bias (Bottom Left) and Variance (Bottom Right) performance of \DBL^2, A1DE, \DBL^3 and A2DE for \emph{All} and \emph{Big} datasets.}
\caption{\small Geometric mean of 0-1 Loss (Left) and RMSE (Right) performance of \DBL^2, A1DE, \DBL^3 and A2DE for \emph{Little} and \emph{Big} datasets.}
\label{fig_wAnJEvsAnDE}
\end{figure}

Normalized 0-1 Loss and RMSE are shown in Figure~\ref{fig_wAnJEvsAnDE}. 
It can be seen that the \DBL^n models have lower 0-1 Loss and RMSE than the corresponding AnDE models.
%On average, on \emph{Big} datasets, \DBL^n reduces the bias of the equivalent AnDE model by a factor of two to four.
%A similar pattern holds for 0-1 Loss. 
%The RMSE of \DBL^n is also lower than AnDE on this collection of datasets.
%We will compare and analyze results on \emph{Big} datasets individually in Appendix~\ref{sec_discussion}. 
%The improvement can be seen from W-D-L of 0-1 Loss and RMSE on \emph{Big} datasets in Table~\ref{tab_wAnJEvsAnDE}. Note, the super-script $+3$ indicates that the results could not be calculated on three datasets for both \DBL^n and AnDE models. These are three MITFaceSet datasets that is: \texttt{MITFaceSetA, MITFaceSetB, MITFaceSetC}. 
% --------------------------
% --------------------------

\begin{figure}
\centering
\includegraphics[width=55mm,height=40mm]{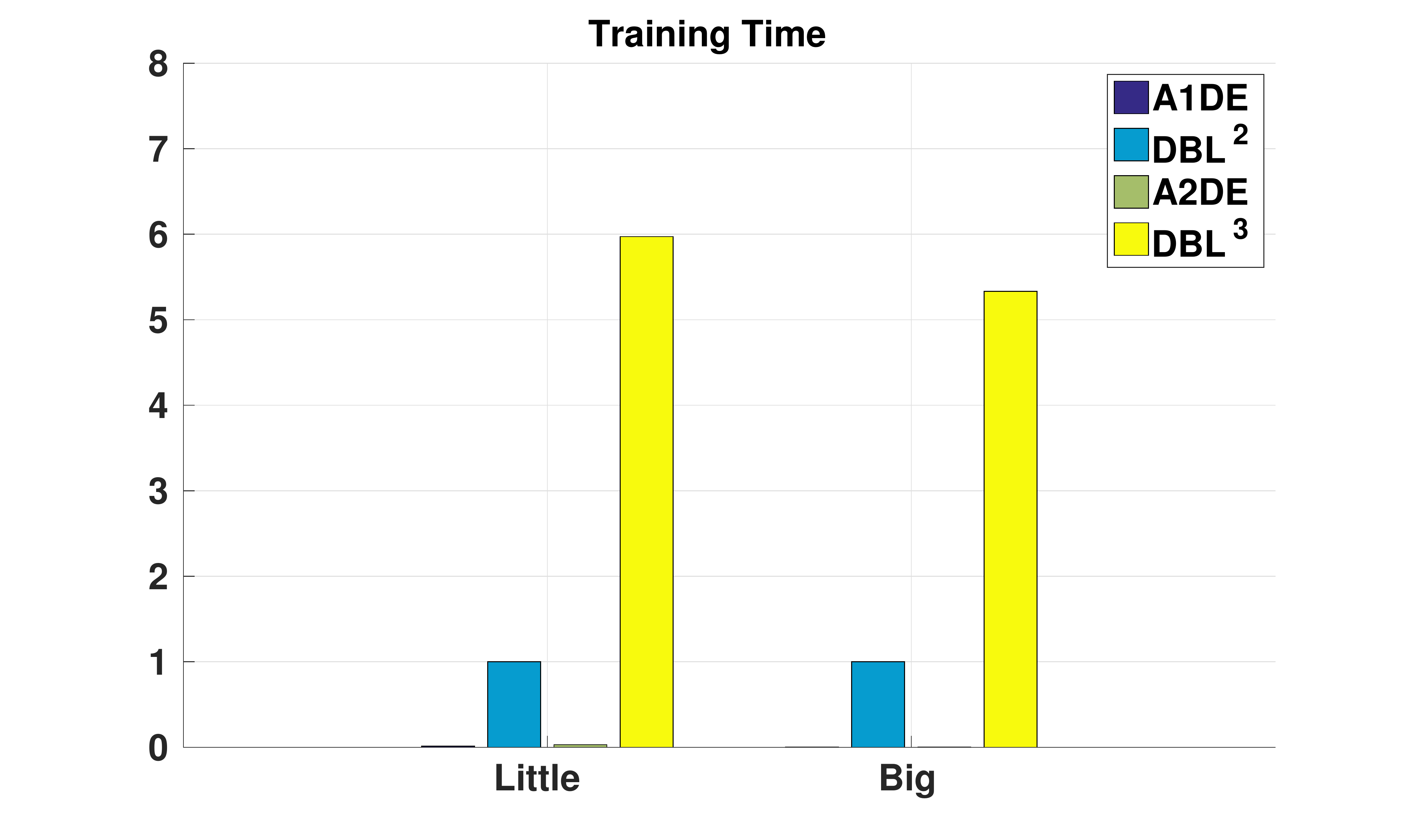}
\includegraphics[width=55mm,height=40mm]{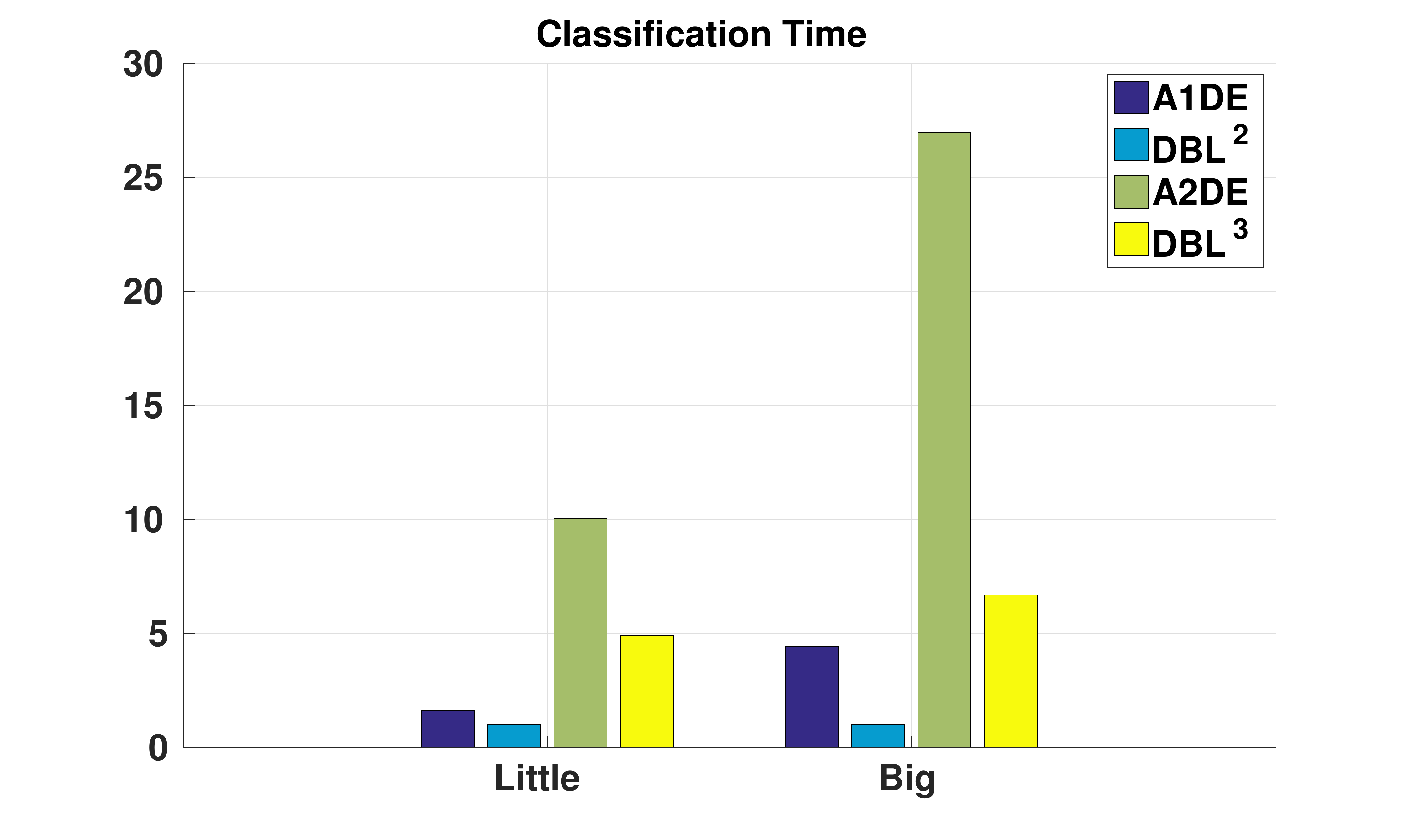}
\caption{\small Geometric mean of Training Time (Left), Classification Time (Right) of \DBL^2, A1DE, \DBL^3 and A2DE for \emph{All} and \emph{Big} datasets.}
\label{fig_wAnJEvsAnDE_Time}
\end{figure}

A comparison of the training time of \DBL^n and AnDE is given in Figure~\ref{fig_wAnJEvsAnDE_Time}. 
%As training an AnDE model is just the calculation of sufficient statistics from the data, 
As expected, due to its additional discriminative learning, \DBL^n requires substantially more training time than AnDE. 
However, AnDE does not share such a consistent advantage with respect to classification time, the relativities depending on the dimensionality of the data. For high-dimensional data the large number of permutations of attributes that AnDE must consider results in greater computation.
% --------------------------
% --------------------------

%%%%%%%%%%%%%%%%%%%%%%%%%%%%%%%%%%%%%%%%%%%%%%%%%%%%%%%%%%%%%%%%%
\subsection{\DBL^n vs. \LR^n} \label{subsec_DBLvsLR}
%%%%%%%%%%%%%%%%%%%%%%%%%%%%%%%%%%%%%%%%%%%%%%%%%%%%%%%%%%%%%%%%%
In this section, we will compare the two \DBL^n models with their equivalent \LR^n models. As discussed before, we expect to see similar bias-variance profile and a similar classification performance as the two models are re-parameterization of each other. 

\begin{figure}
\centering
\includegraphics[width=50mm,height=50mm]{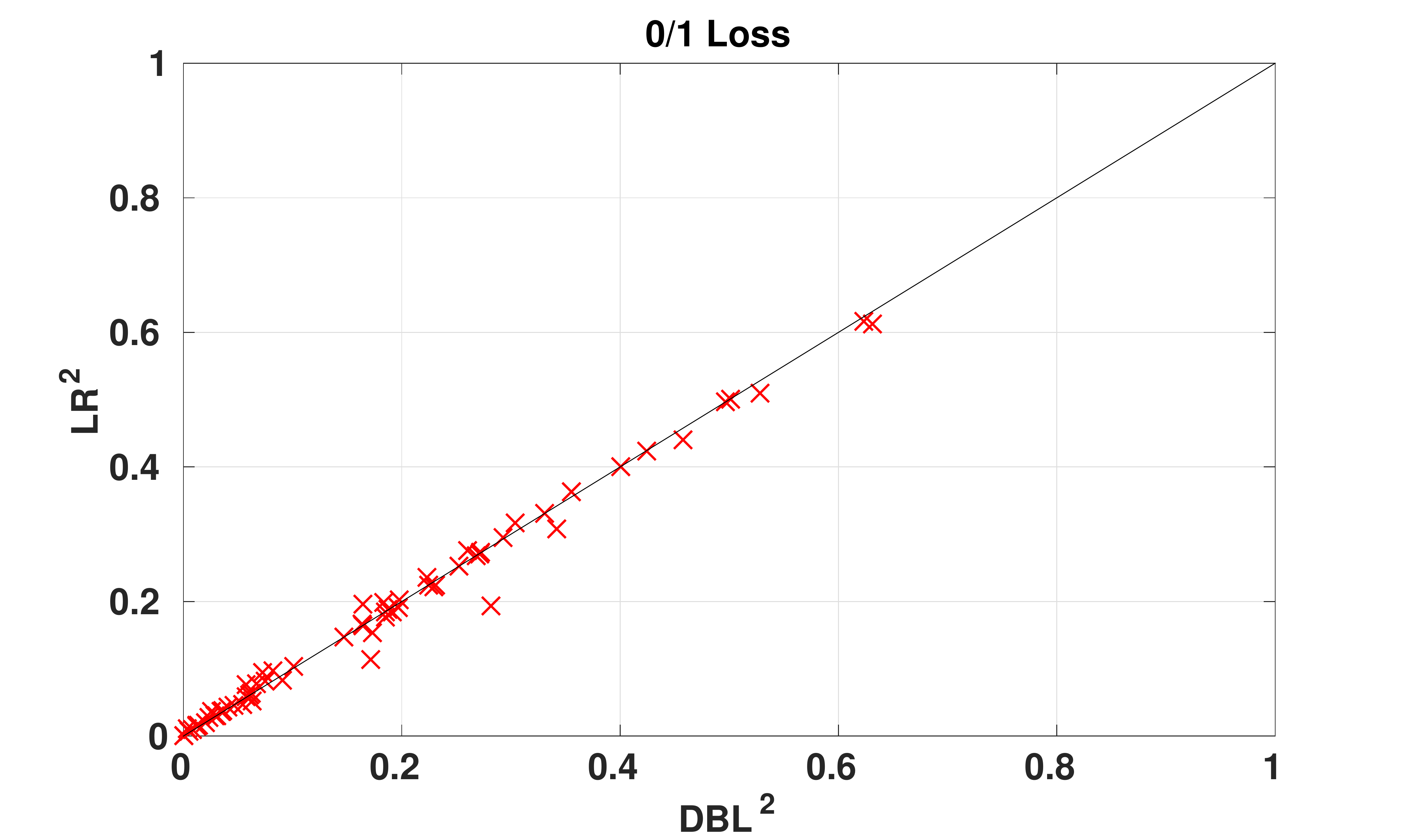}
\includegraphics[width=50mm,height=50mm]{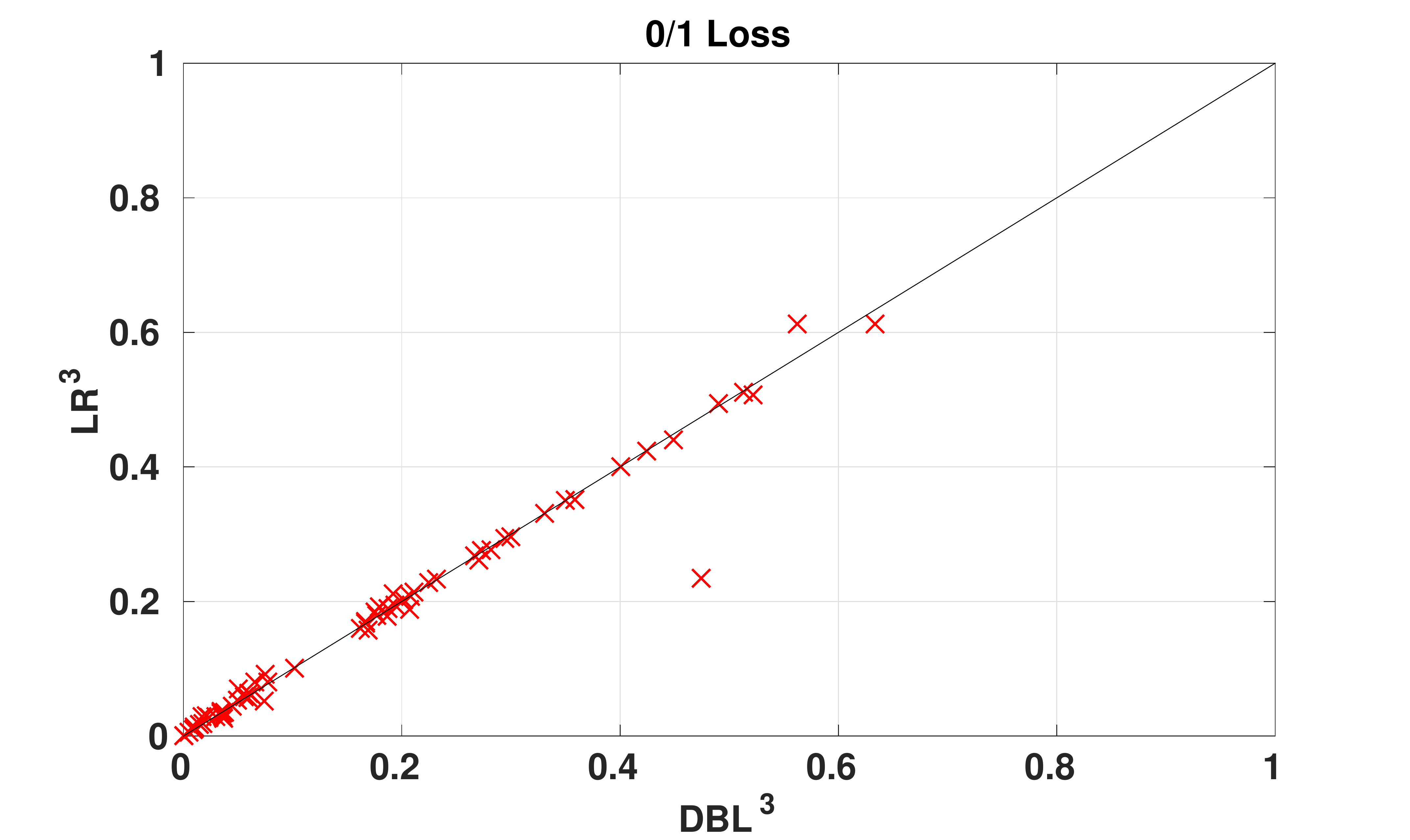}
\caption{\small Comparative scatter of 0-1 Loss of \DBL^2 and \LR^2 (Left) and \DBL^3 and \LR^3 (Right) for \emph{Little} datasets.}
\label{fig_wAnJE01Loss_Little}
\end{figure}

\begin{figure}
\centering
\includegraphics[width=50mm,height=50mm]{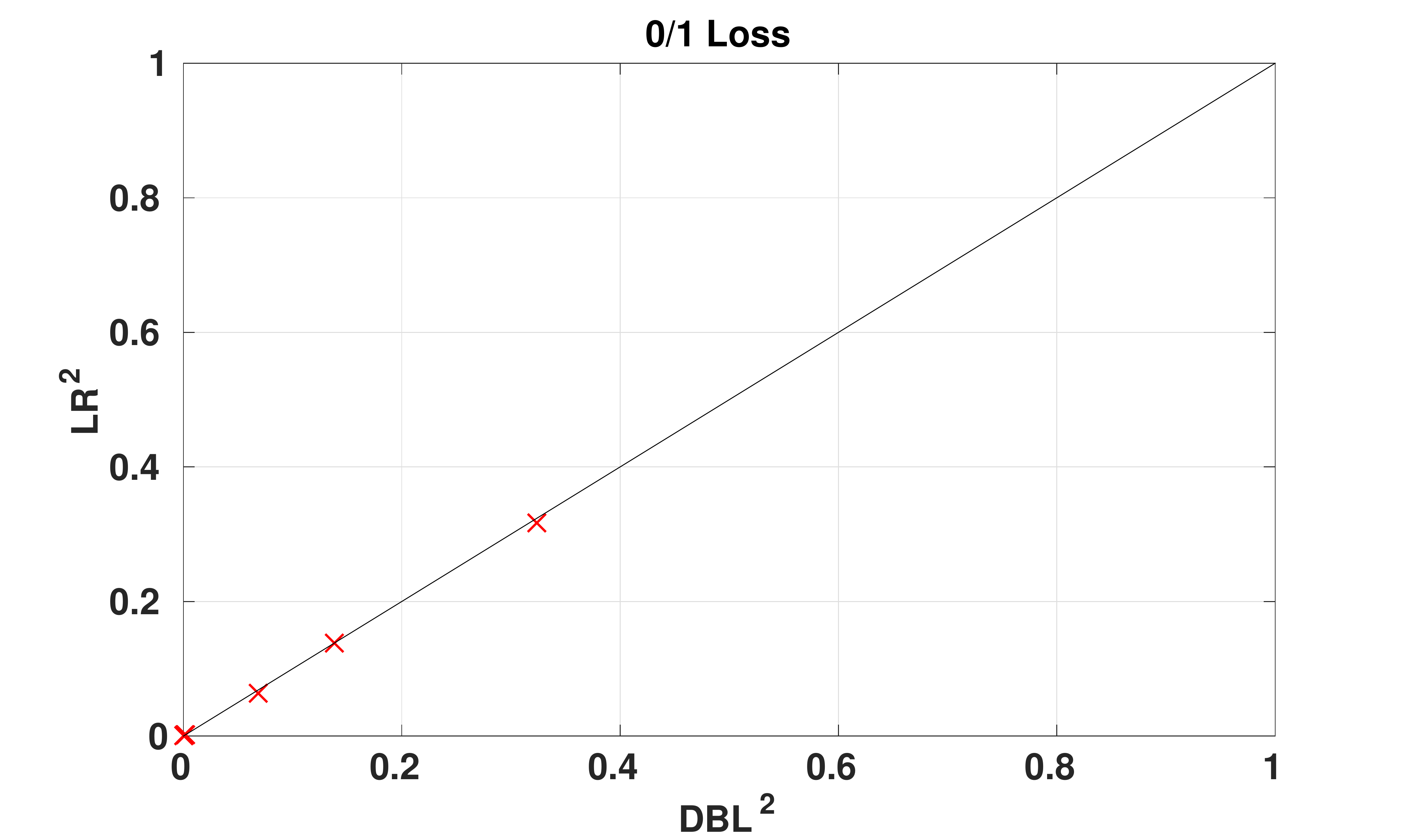}
\includegraphics[width=50mm,height=50mm]{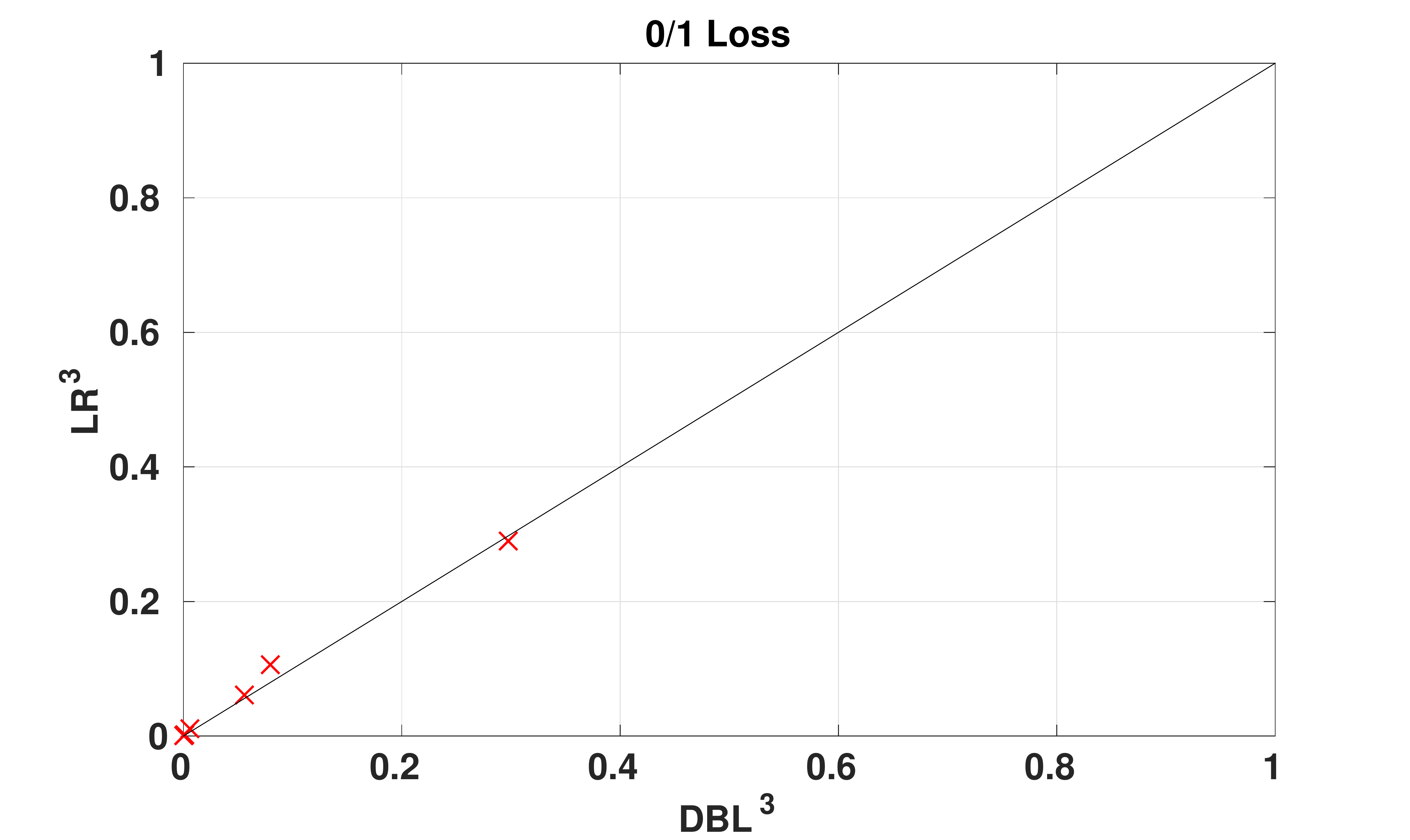}
\caption{\small Comparative scatter of 0-1 Loss of \DBL^2 and \LR^2 (Left) and \DBL^3 and \LR^3 (Right) for \emph{Big} datasets.}
\label{fig_wAnJE01Loss_Big}
\end{figure}

\begin{figure}
\centering
\includegraphics[width=50mm,height=50mm]{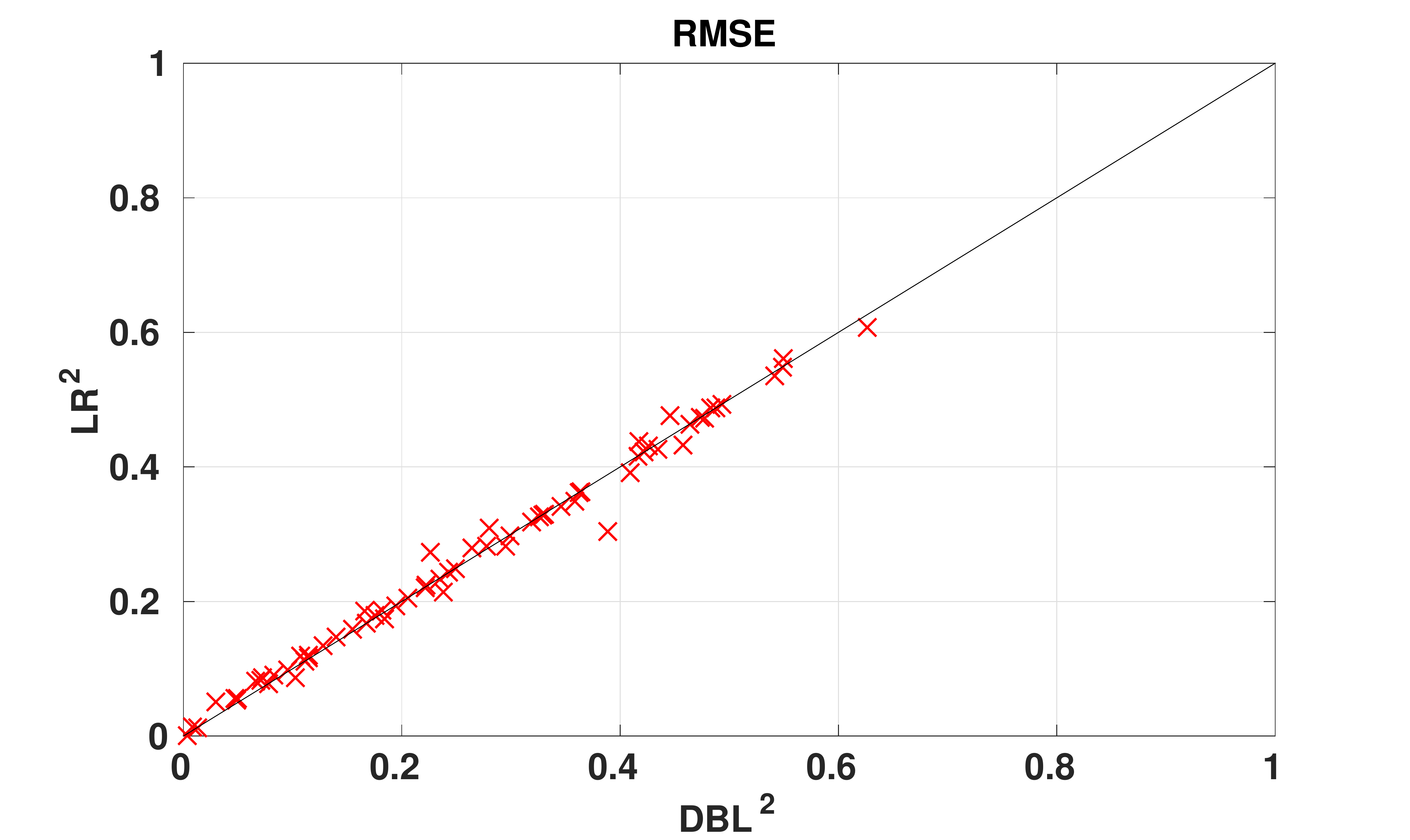}
\includegraphics[width=50mm,height=50mm]{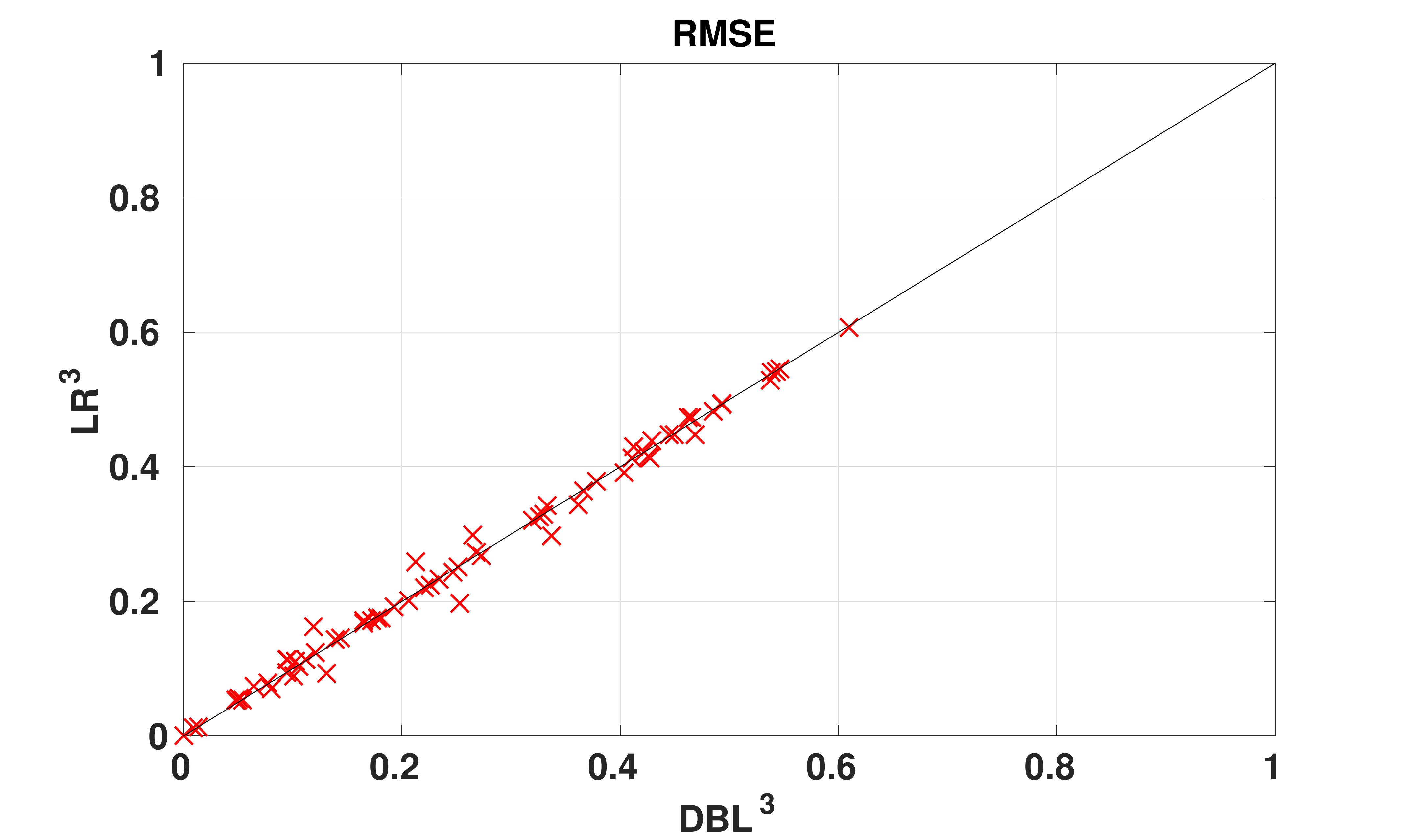}
\caption{\small Comparative scatter of RMSE of \DBL^2 and \LR^2 (Left) and \DBL^3 and \LR^3 (Right) for \emph{Little} datasets.}
\label{fig_wAnJERMSE_Little}
\end{figure}

\begin{figure}
\centering
\includegraphics[width=50mm,height=50mm]{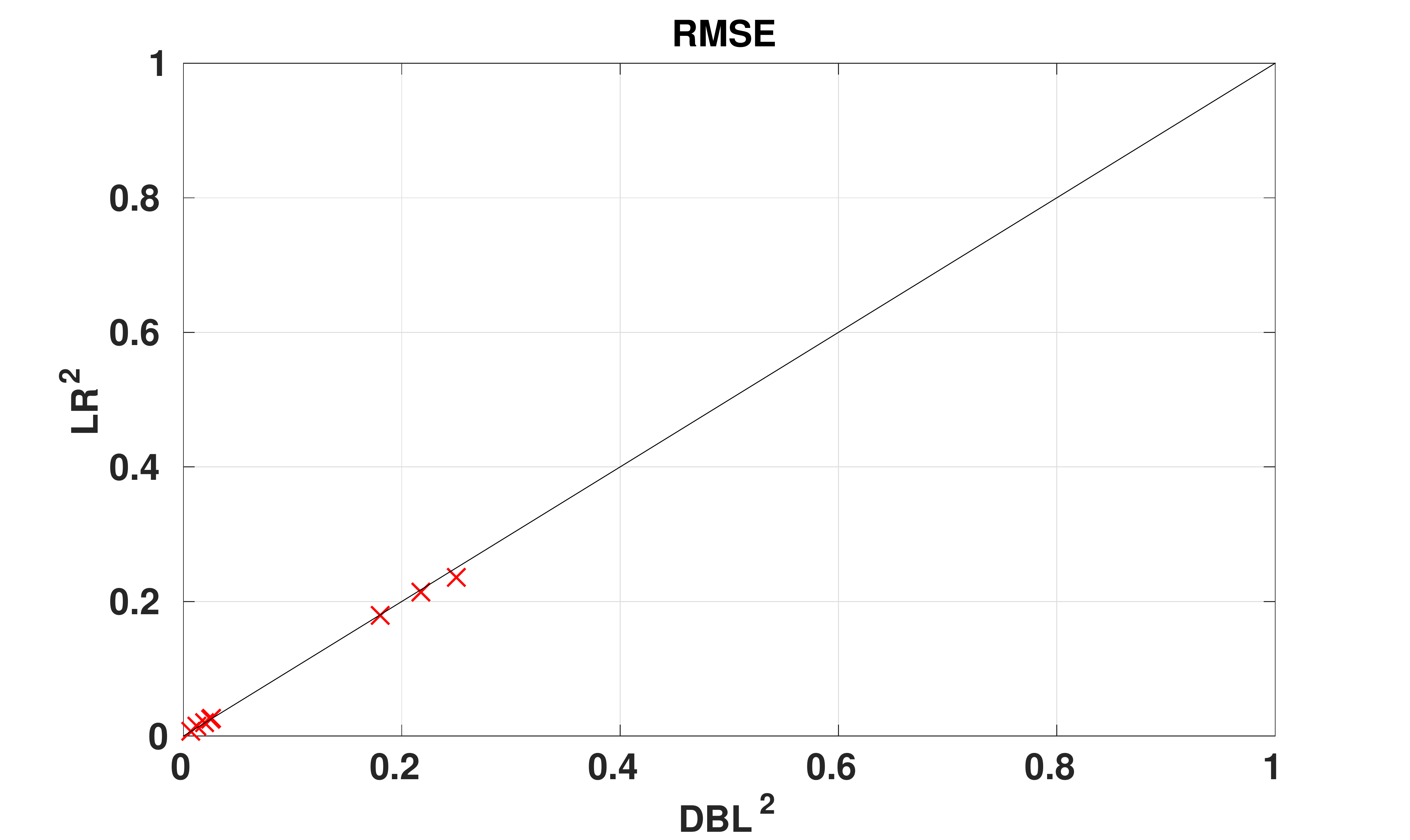}
\includegraphics[width=50mm,height=50mm]{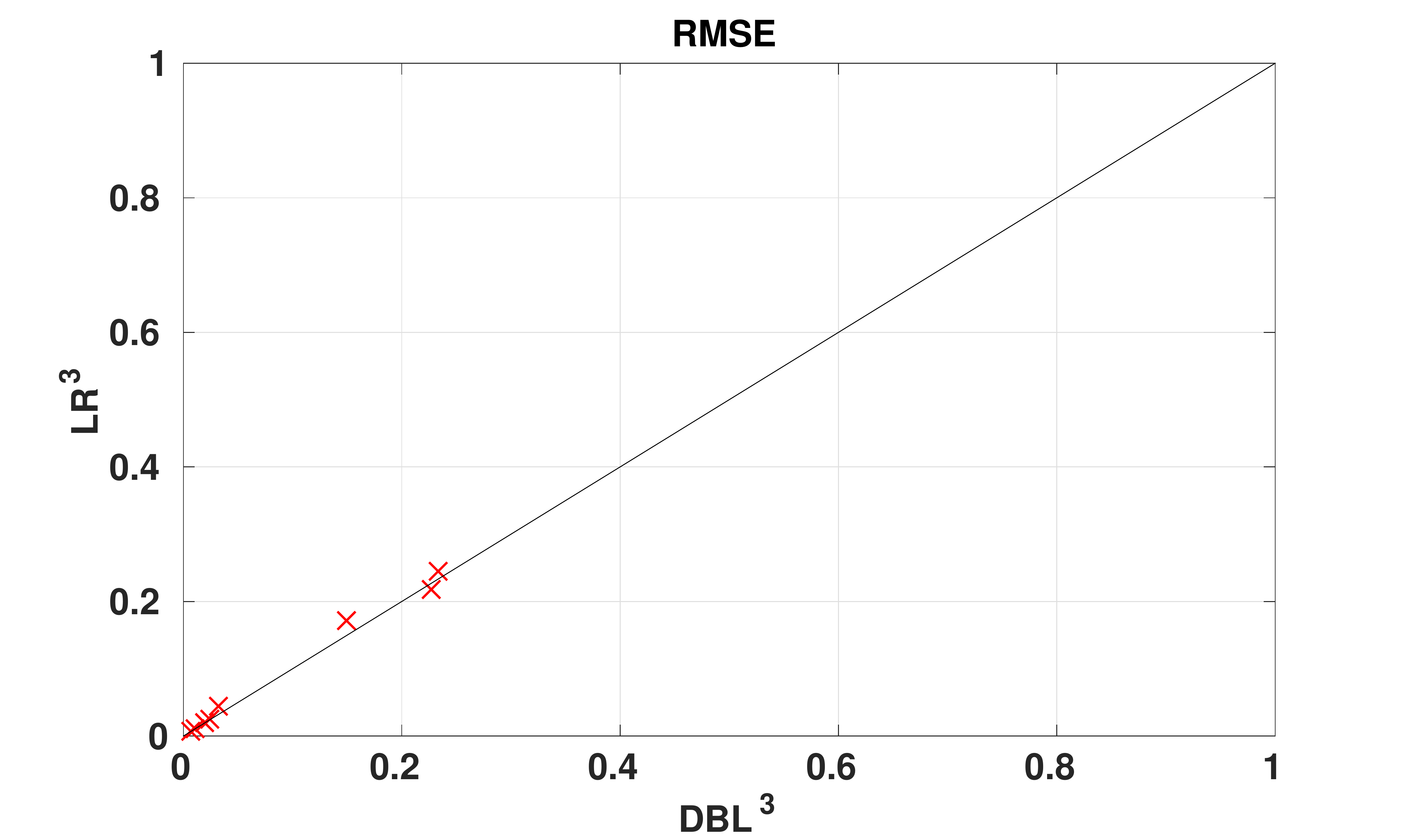}
\caption{\small Comparative scatter of RMSE of \DBL^2 and \LR^2 (Left) and \DBL^3 and \LR^3 (Right) for \emph{Big} datasets.}
\label{fig_wAnJERMSE_Big}
\end{figure}

We compare the two parameterizations in terms of the scatter of their 0-1 Loss and RMSE values on \emph{Little} datasets in Figure~\ref{fig_wAnJE01Loss_Little},~\ref{fig_wAnJERMSE_Little} respectively, and on \emph{Big} datasets in Figure~\ref{fig_wAnJE01Loss_Big},~\ref{fig_wAnJERMSE_Big} respectively.
It can be seen that the two parameterizations (with an exception of one dataset, that is: \texttt{wall-following}) have a similar spread of 0-1 Loss and RMSE values for both $n=2$ and $n=3$.
% --------------------------
% --------------------------
% --------------------------
% --------------------------
% --------------------------
% --------------------------
% --------------------------
% --------------------------

\begin{figure}
\centering
\includegraphics[width=50mm,height=50mm]{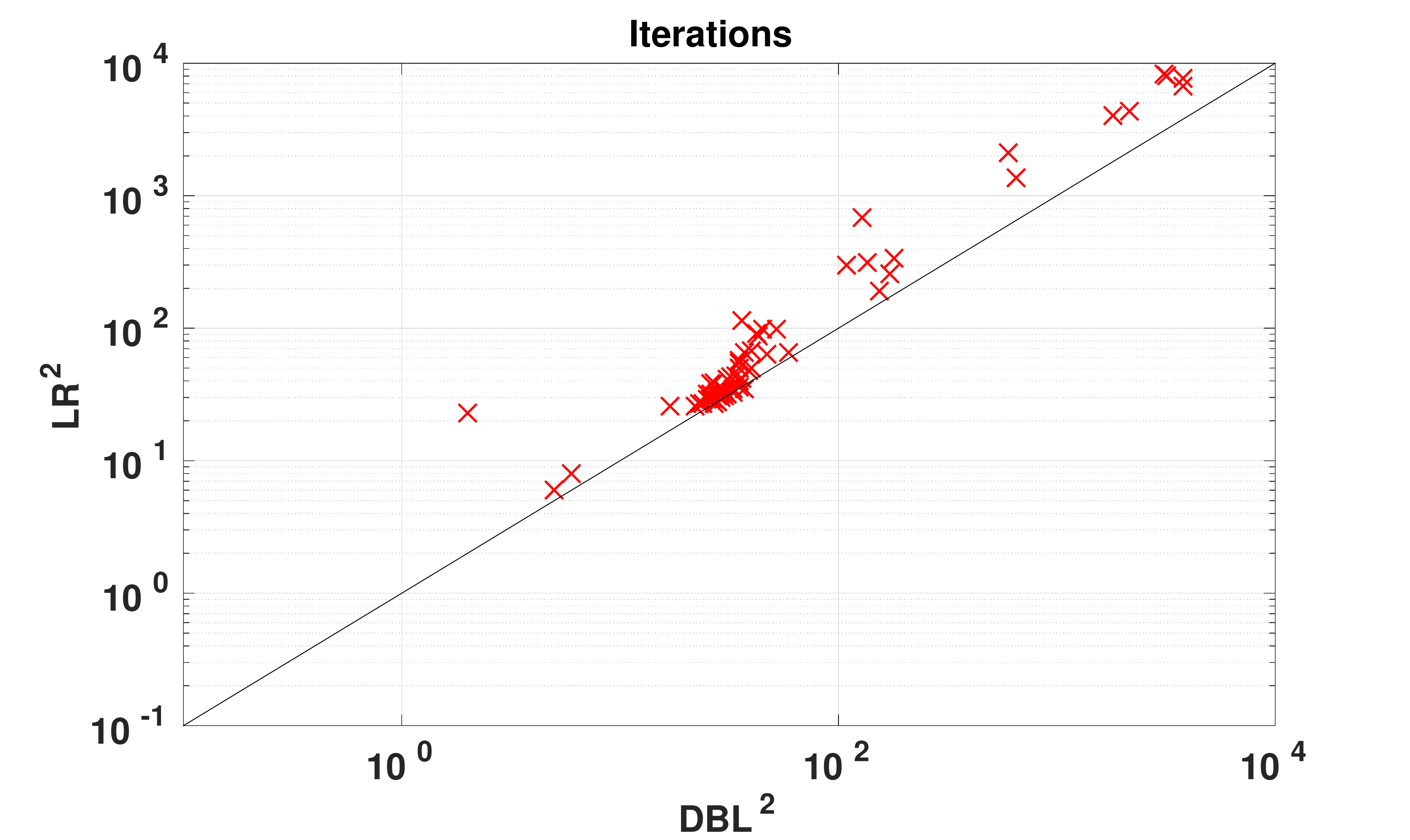}
\includegraphics[width=50mm,height=50mm]{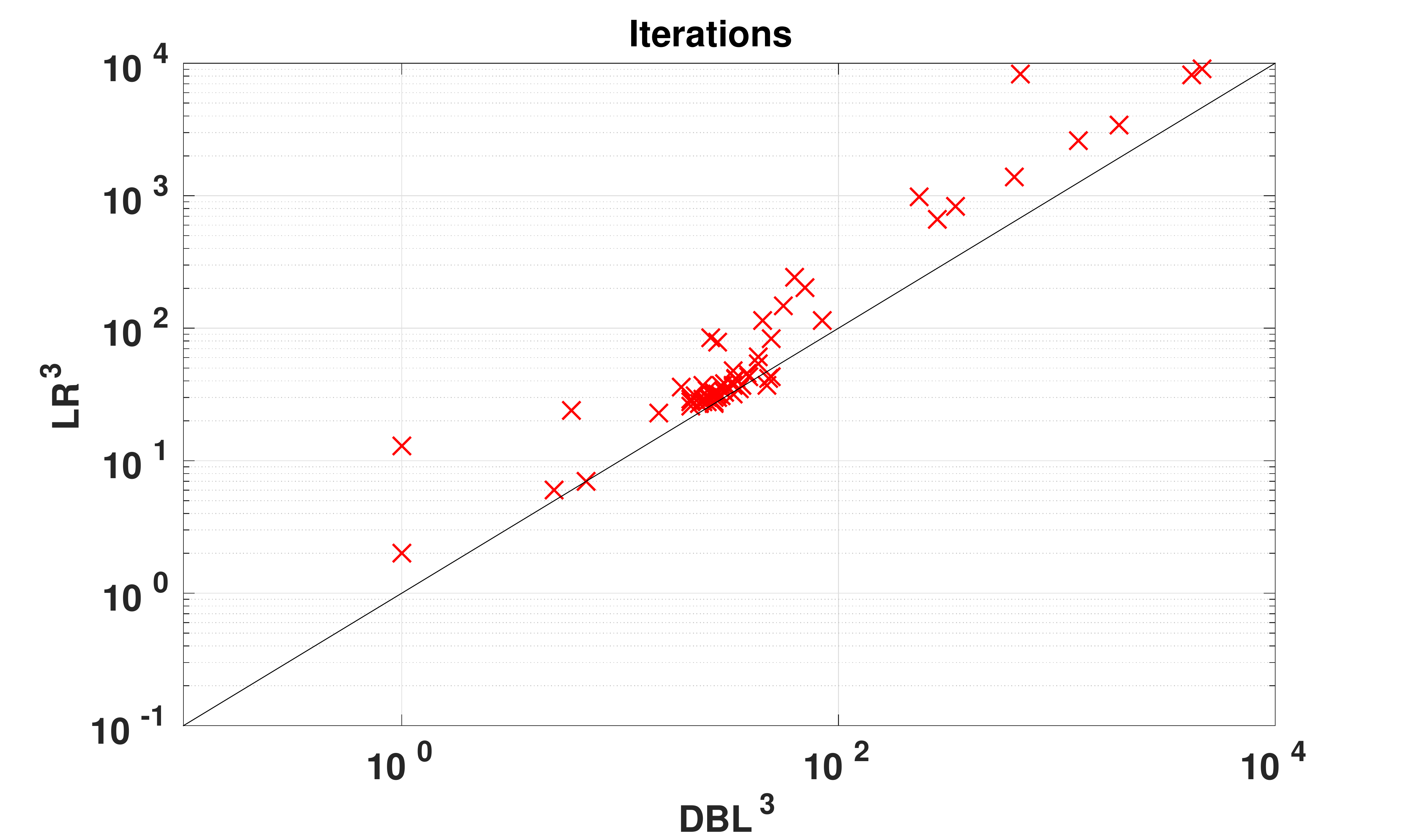}
\caption{\small Comparative scatter of number of iterations of \DBL^2 and \LR^2 (Left) and \DBL^3 and \LR^3 (Right) for \emph{Little} datasets.}
\label{fig_wAnJEIterations_Little}
\end{figure}

\begin{figure}
\centering
\includegraphics[width=50mm,height=50mm]{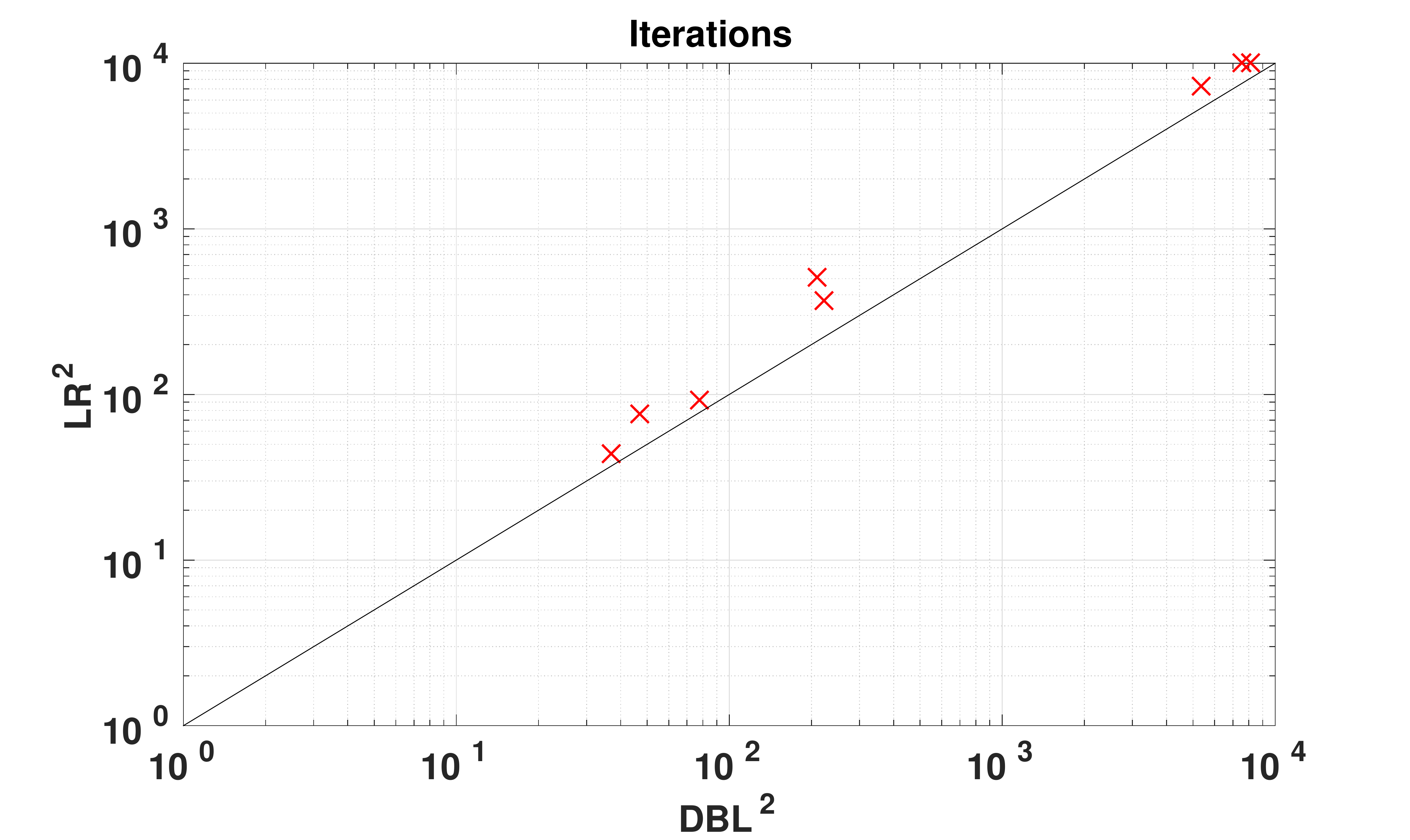}
\includegraphics[width=50mm,height=50mm]{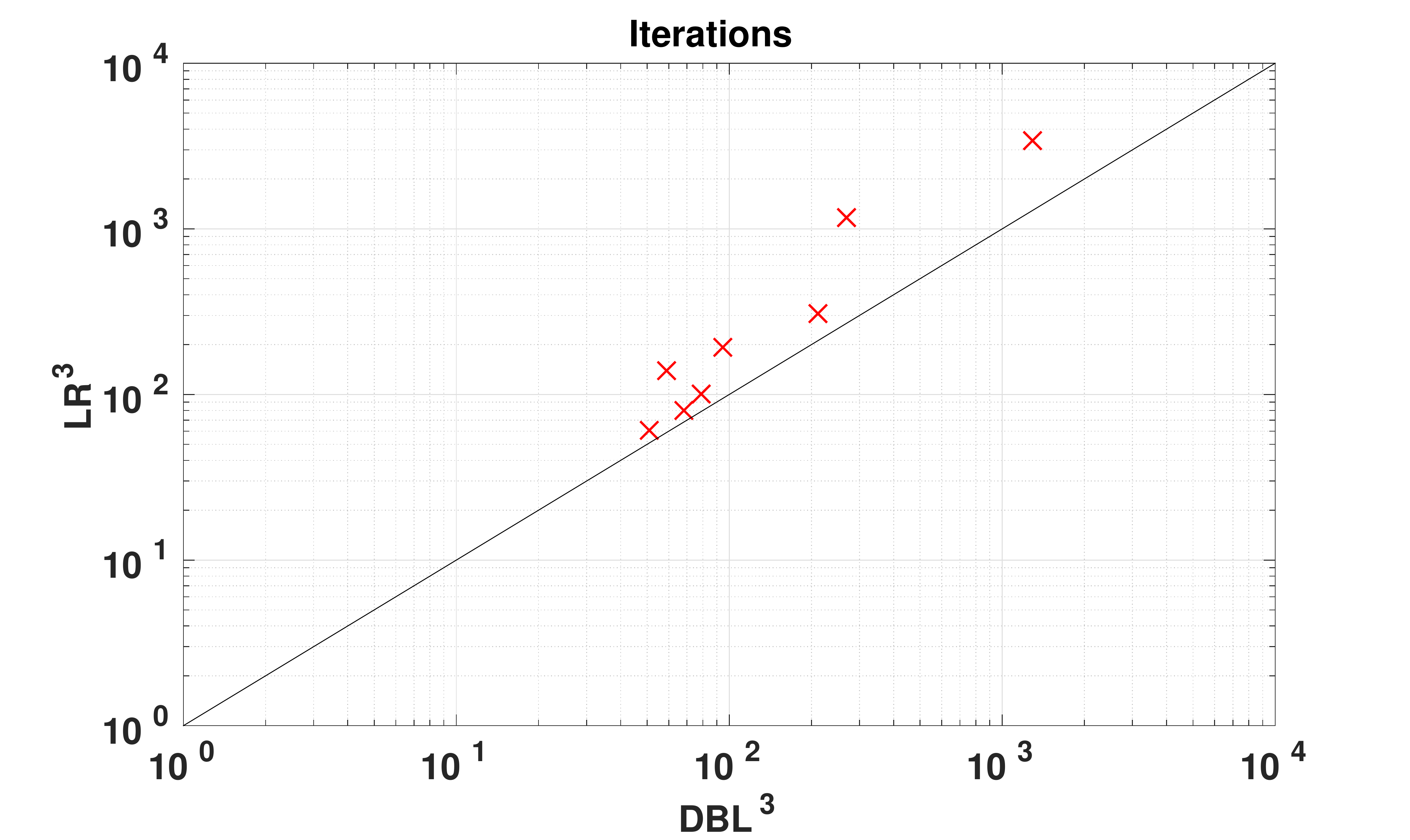}
\caption{\small Comparative scatter of iterations of \DBL^2 and \LR^2 (Left) and \DBL^3 and \LR^3 (Right) for \emph{Big} datasets.}
\label{fig_wAnJEIterations_Big}
\end{figure}

\begin{figure}
\centering
\includegraphics[width=50mm,height=50mm]{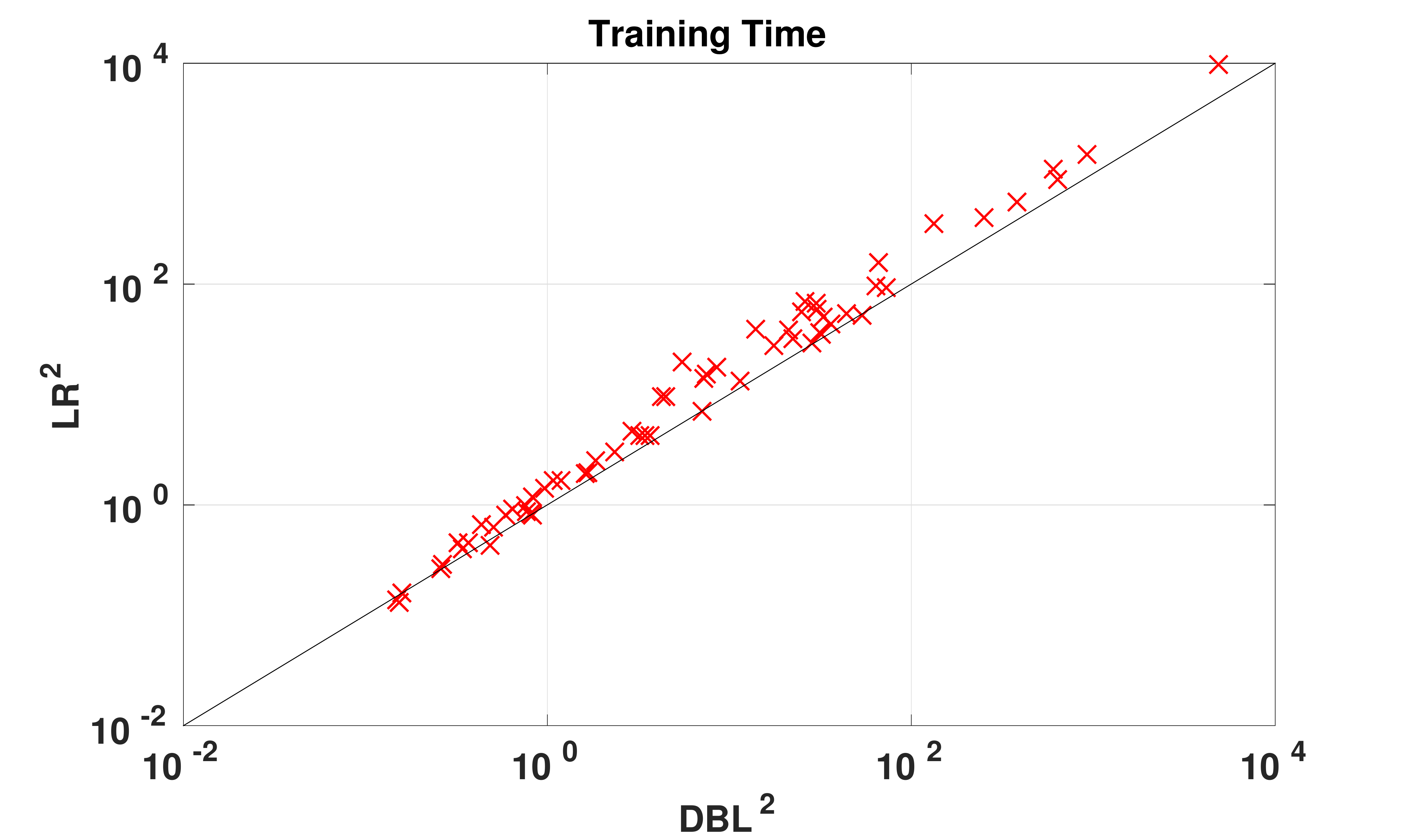}
\includegraphics[width=50mm,height=50mm]{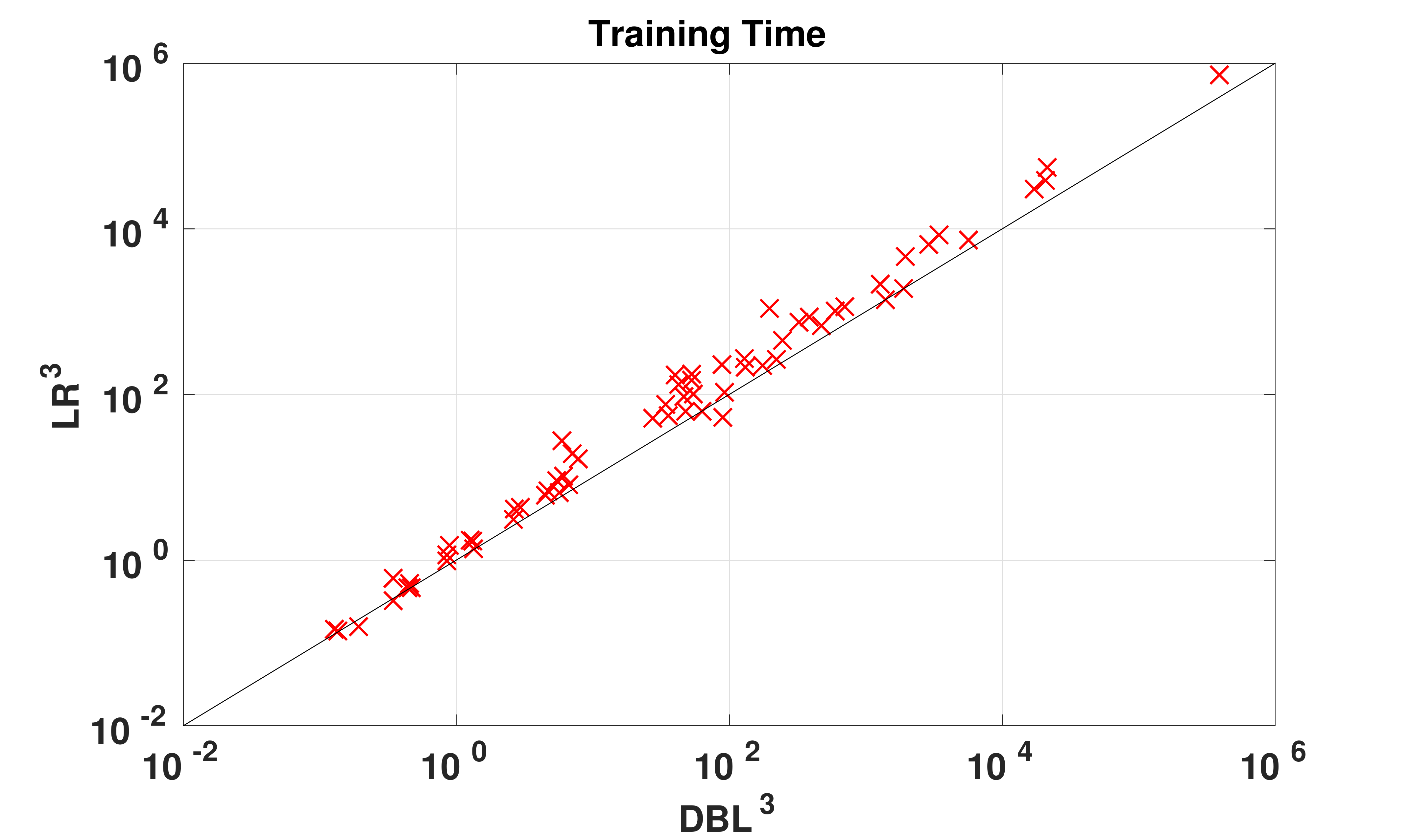}
\caption{\small Comparative scatter of Training time of \DBL^2 and \LR^2 (Left) and \DBL^3 and \LR^3 (Right) for \emph{Little} datasets.}
\label{fig_wAnJETT_Little}
\end{figure}

\begin{figure}
\centering
\includegraphics[width=50mm,height=50mm]{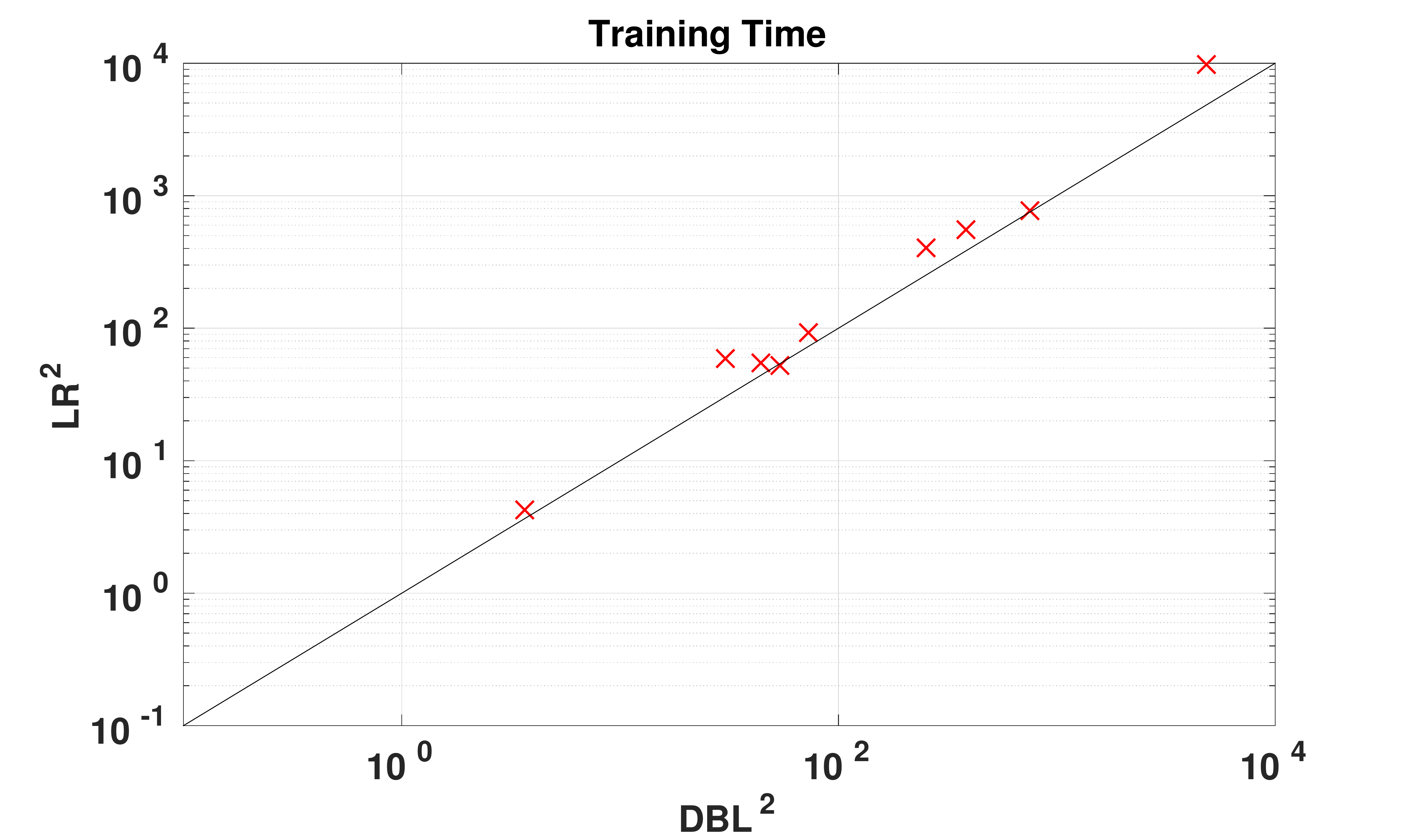}
\includegraphics[width=50mm,height=50mm]{LR2vsDBL2_TT_Big}
\caption{\small Comparative scatter of Training time of \DBL^2 and \LR^2 (Left) and \DBL^3 and \LR^3 (Right) for \emph{Big} datasets.}
\label{fig_wAnJETT_Big}
\end{figure}

The comparative scatter of the number of iterations each parameterization takes to converge is shown in Figure~\ref{fig_wAnJEIterations_Little} and~\ref{fig_wAnJEIterations_Big} for \emph{Little} and \emph{Big} datasets respectively. It can be seen that the number of iterations for \DBL^n are far fewer than \LR^n. 
With a similar spread of 0-1 Loss and RMSE values, it is very encouraging to see that \DBL^n converges in far fewer iterations. 
%In fact, there is not a single dataset, where \LR^n converges faster than \DBL^n for either $n=2$ or $n=3$.
% --------------------------
% --------------------------
% --------------------------
% --------------------------
The number of iterations to converge plays a major part in determining an algorithm's training time. The training time of the two parameterizations is shown in Figure~\ref{fig_wAnJETT_Little} and~\ref{fig_wAnJETT_Big} for \emph{Little} and \emph{Big} datasets, respectively. It can be seen that \DBL^n models are much faster than equivalent \LR^n models.
% --------------------------
% --------------------------
% --------------------------
% --------------------------

\begin{figure}[h] 
\centering
\includegraphics[width=47mm,height=35mm]{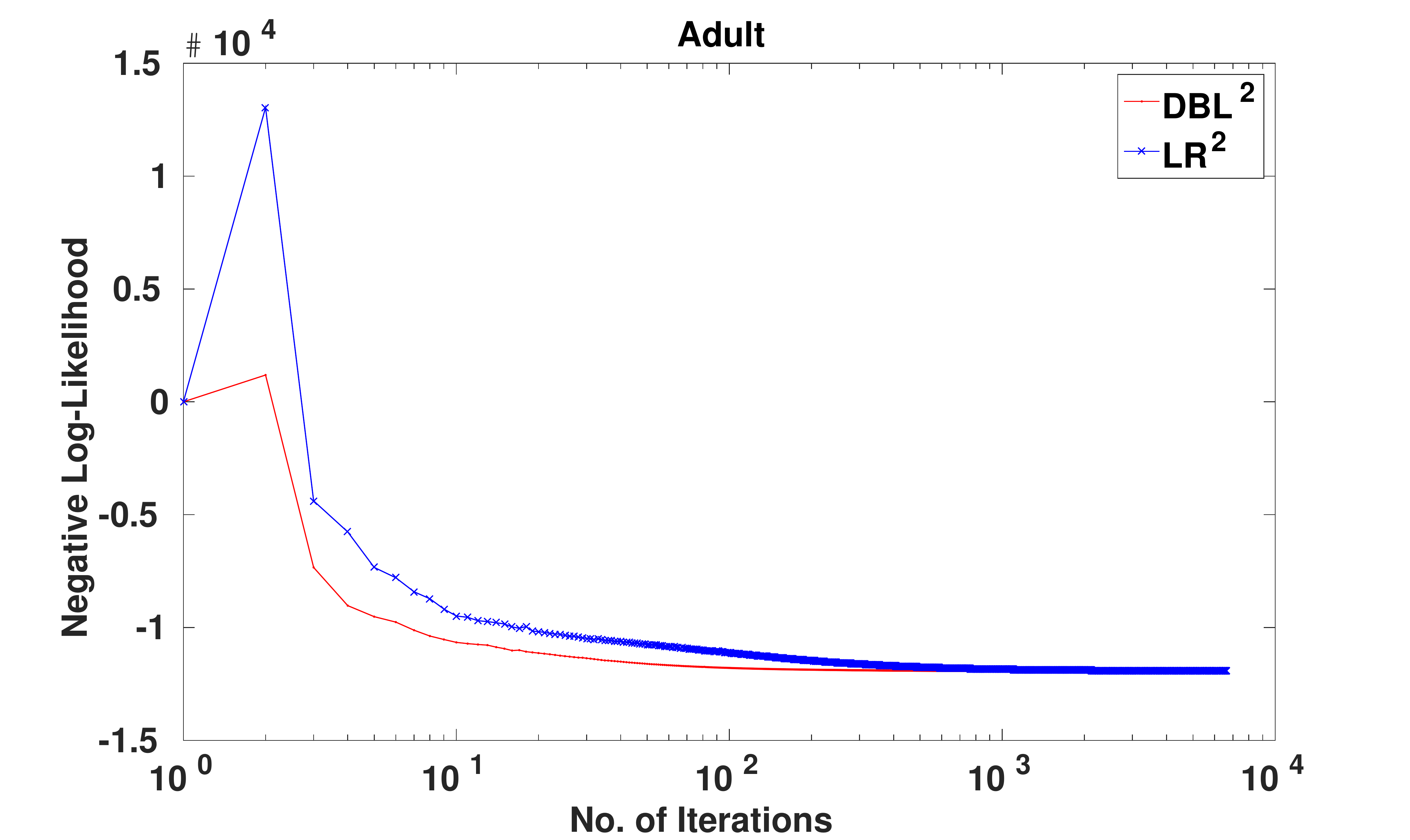}
\includegraphics[width=47mm,height=35mm]{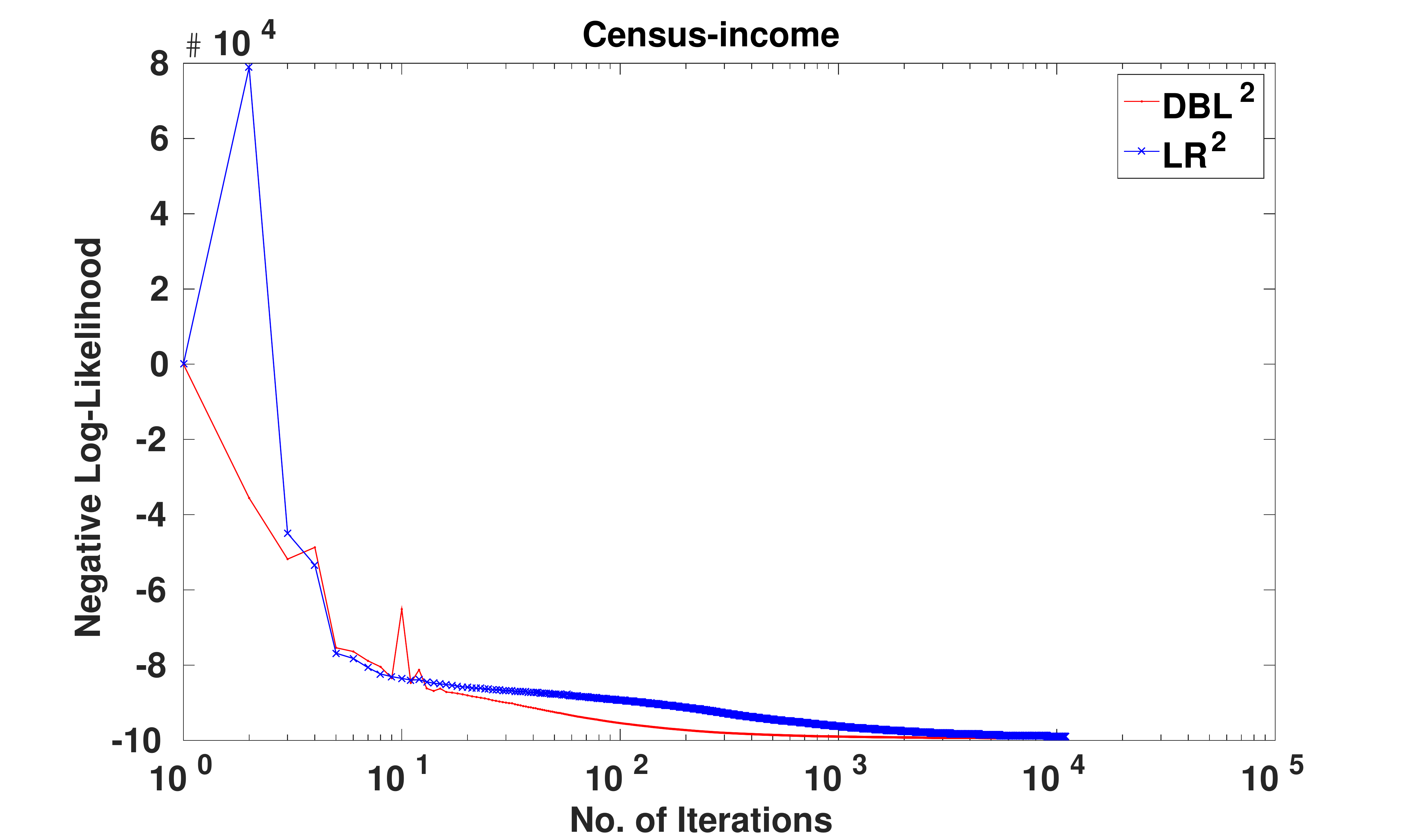}
\includegraphics[width=47mm,height=35mm]{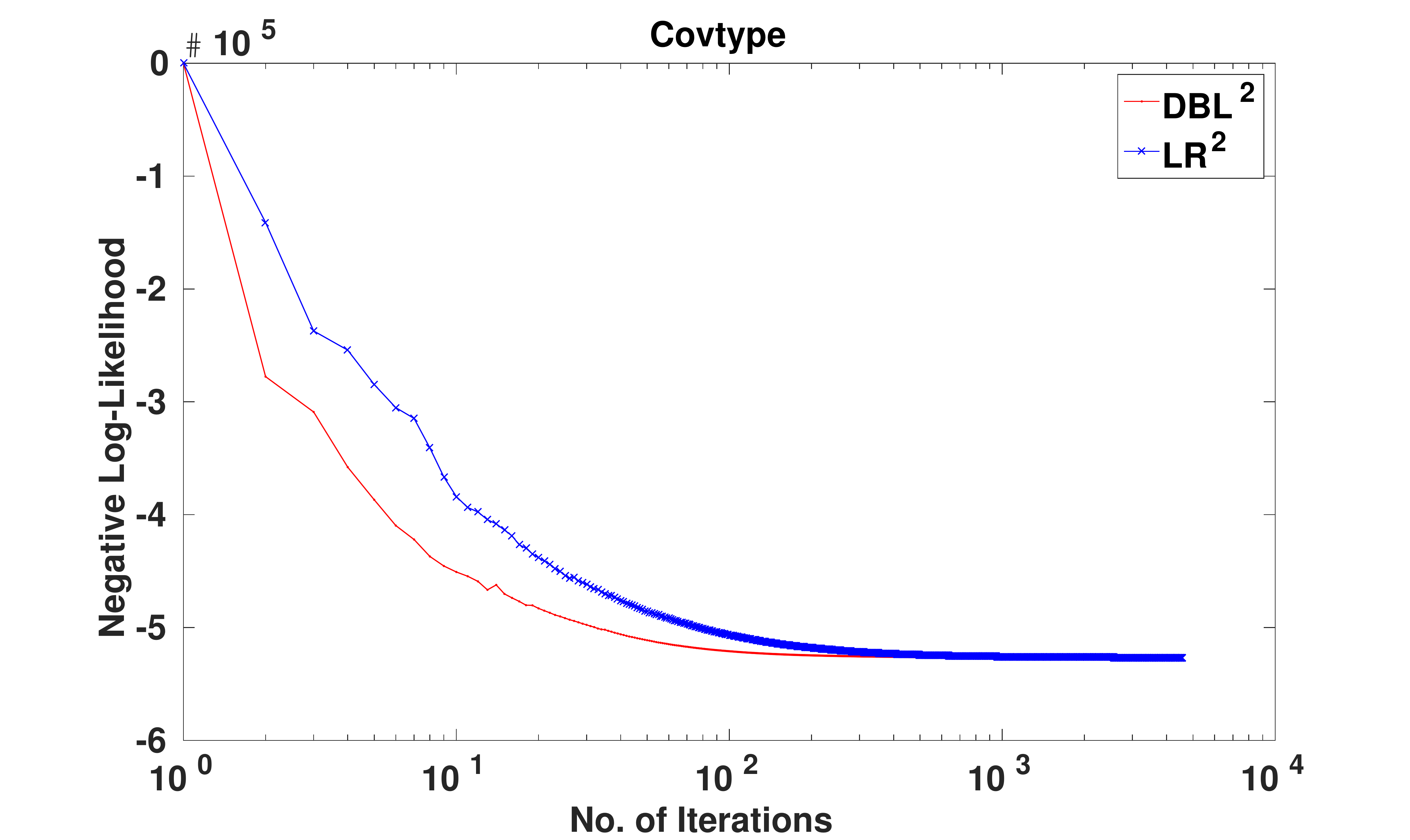}

\includegraphics[width=47mm,height=35mm]{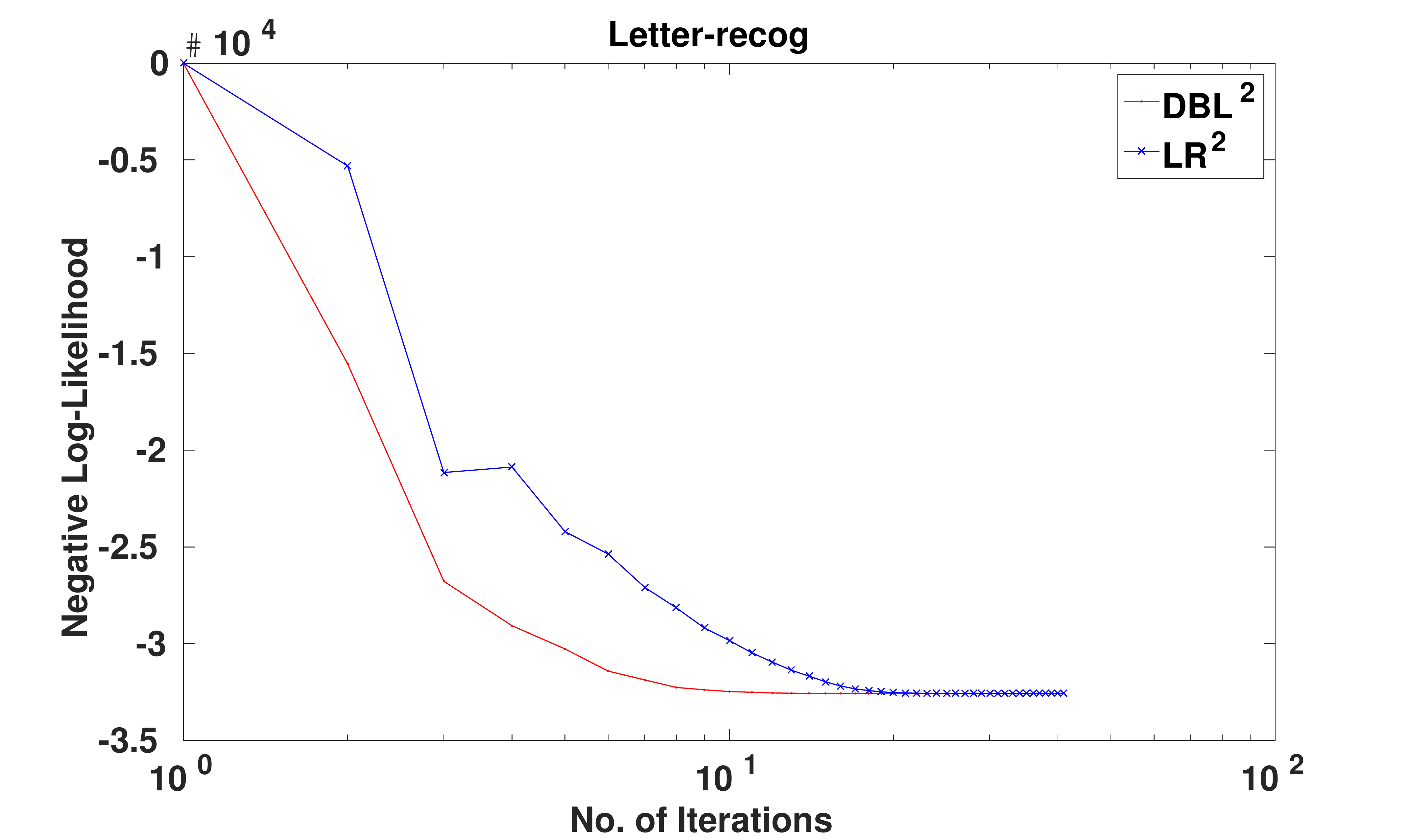}
\includegraphics[width=47mm,height=35mm]{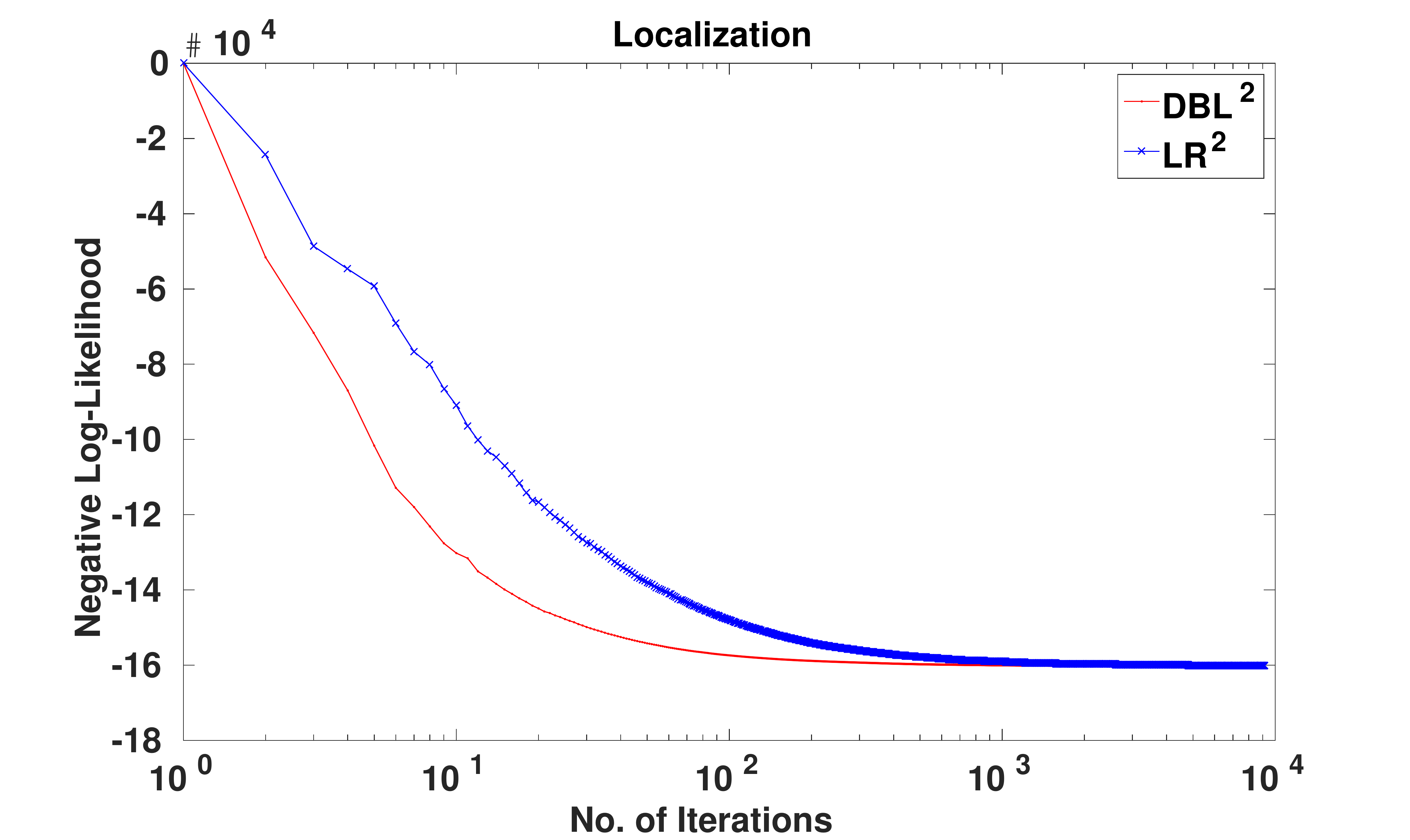}
\includegraphics[width=47mm,height=35mm]{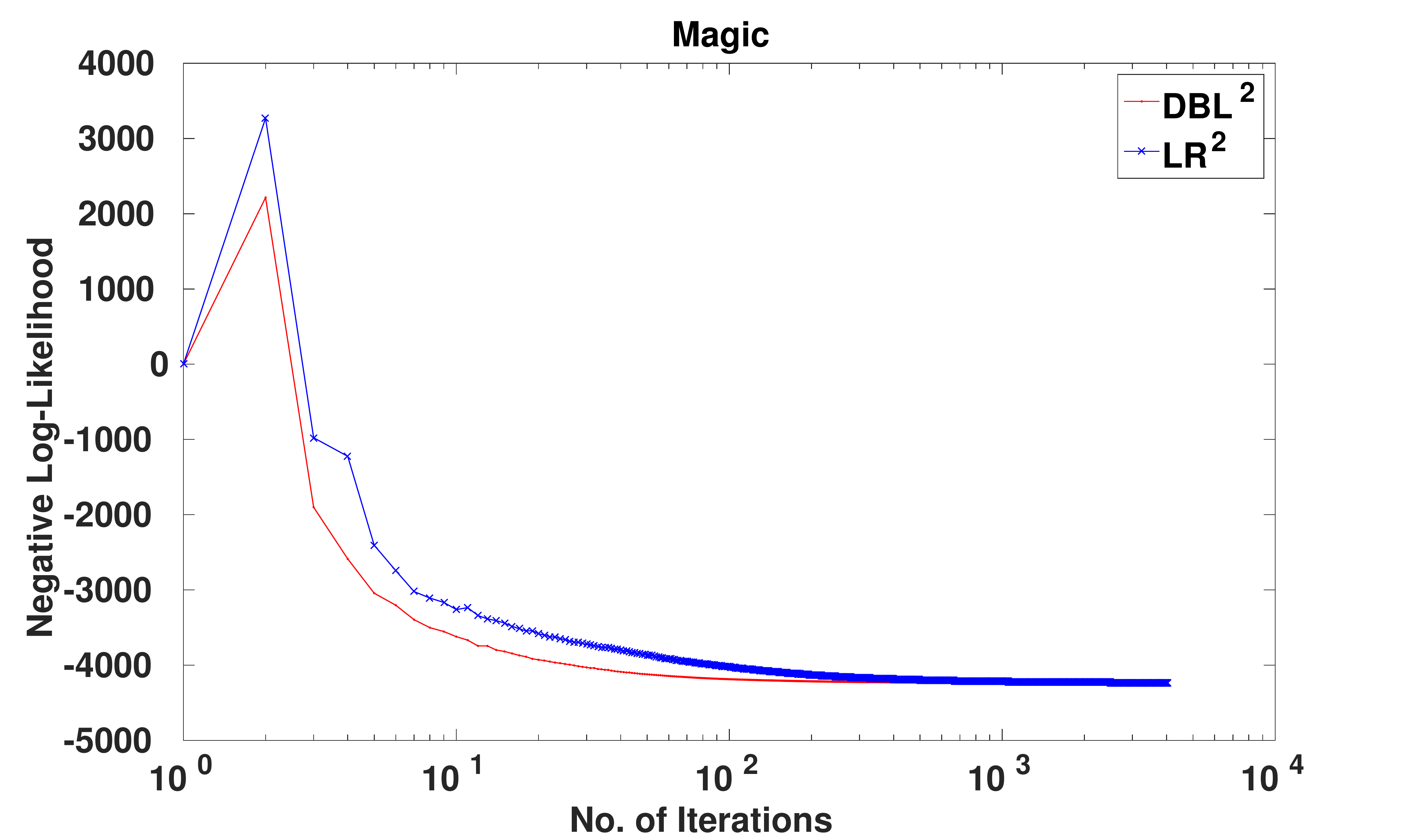}

\includegraphics[width=47mm,height=35mm]{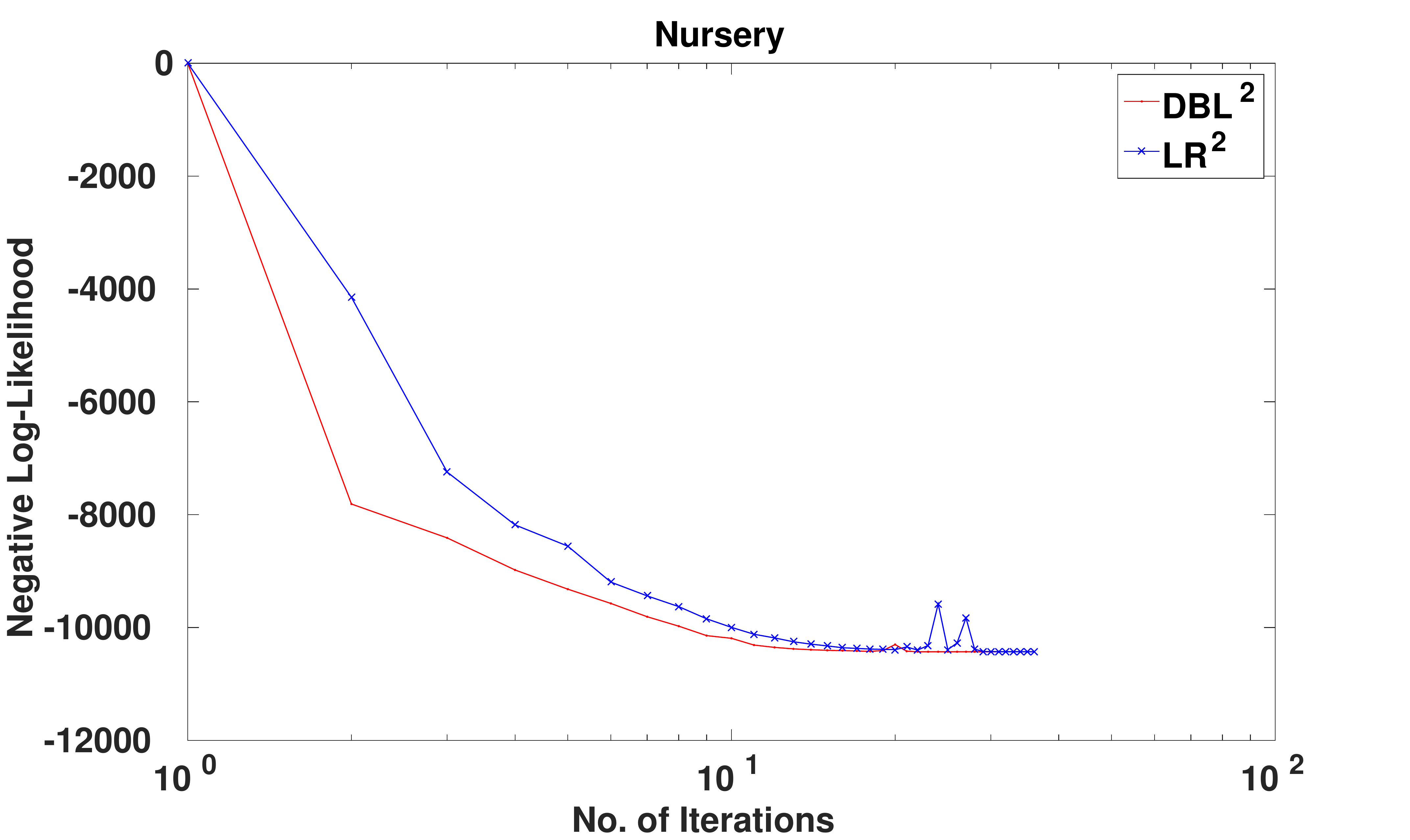}
\includegraphics[width=47mm,height=35mm]{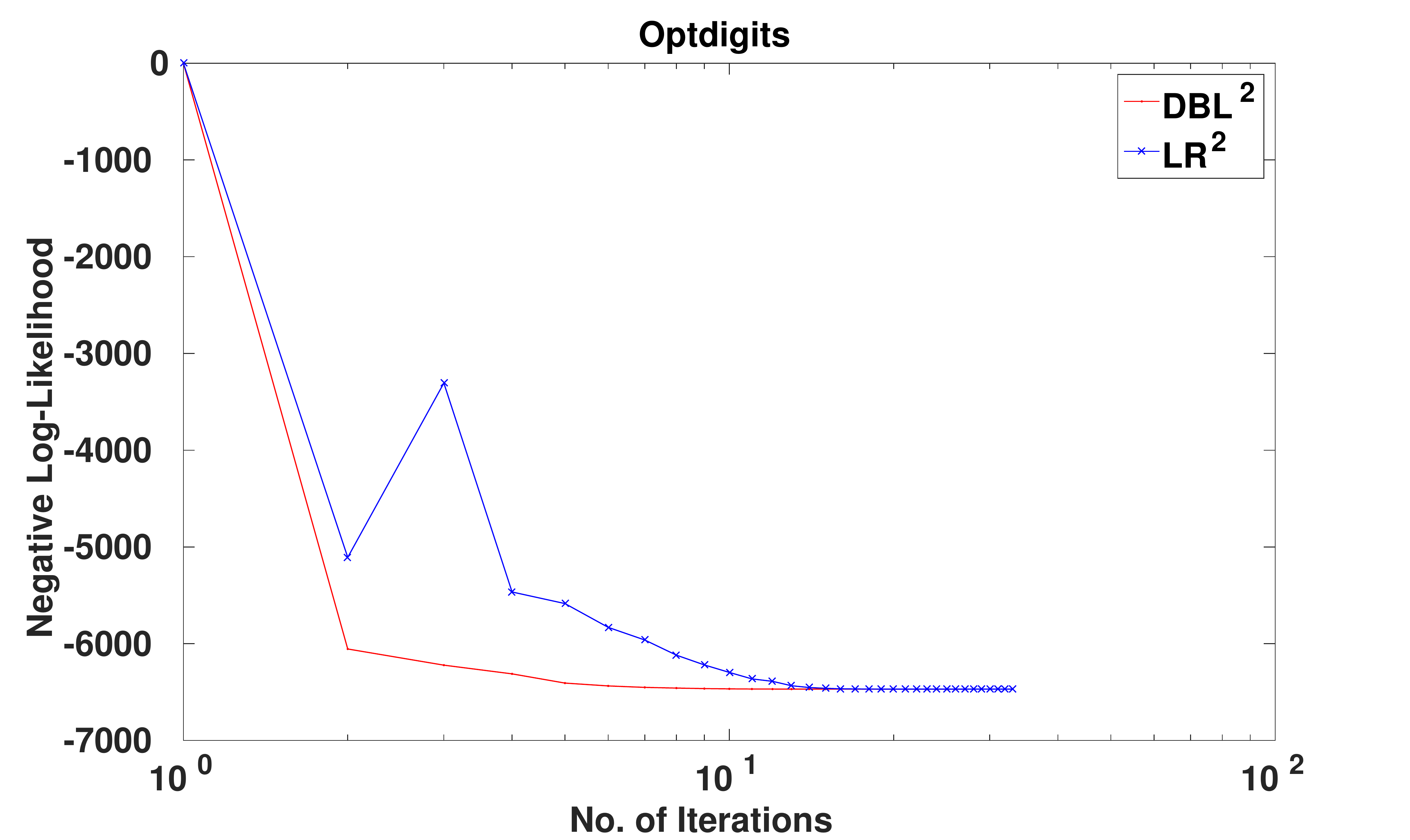}
\includegraphics[width=47mm,height=35mm]{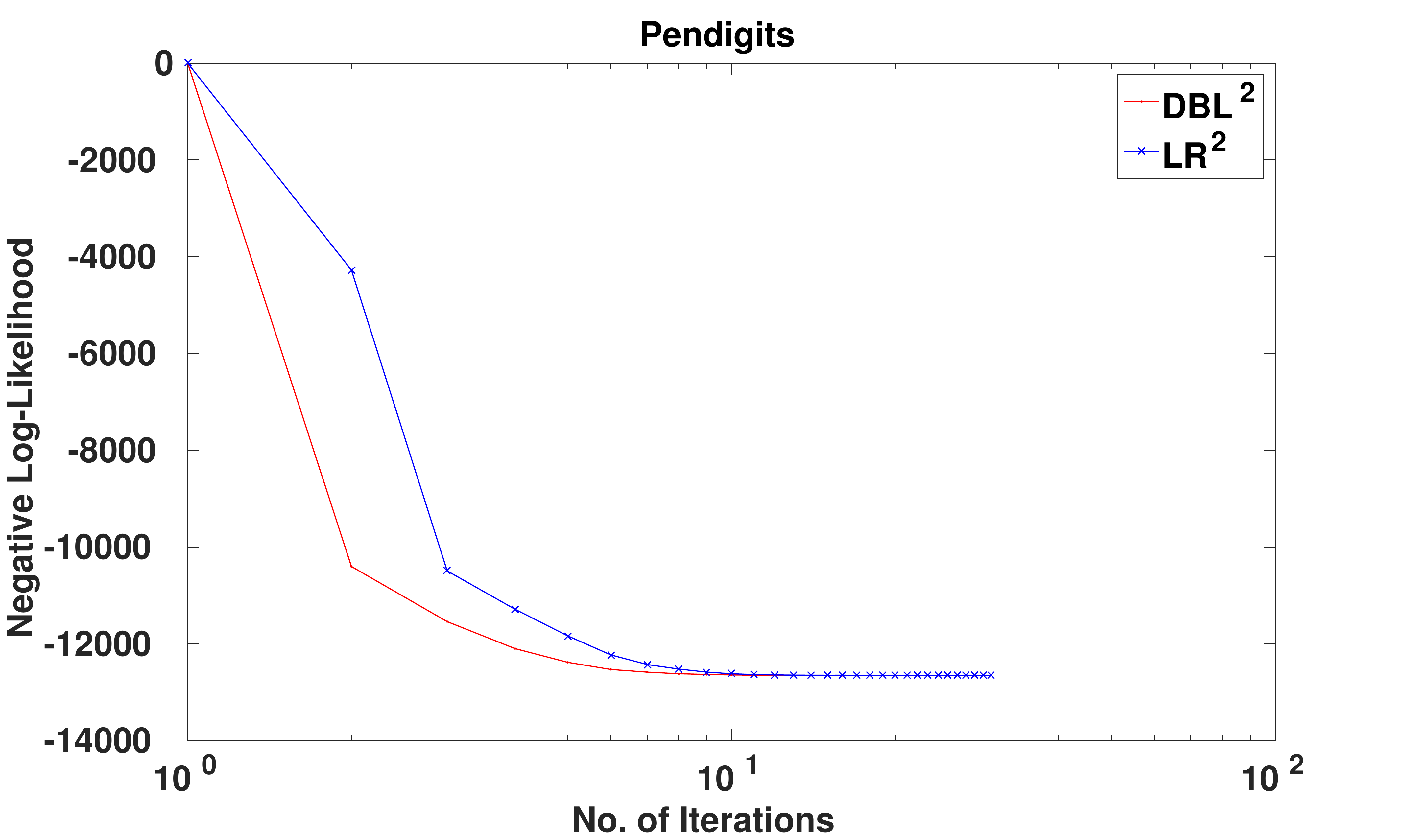}

\includegraphics[width=47mm,height=35mm]{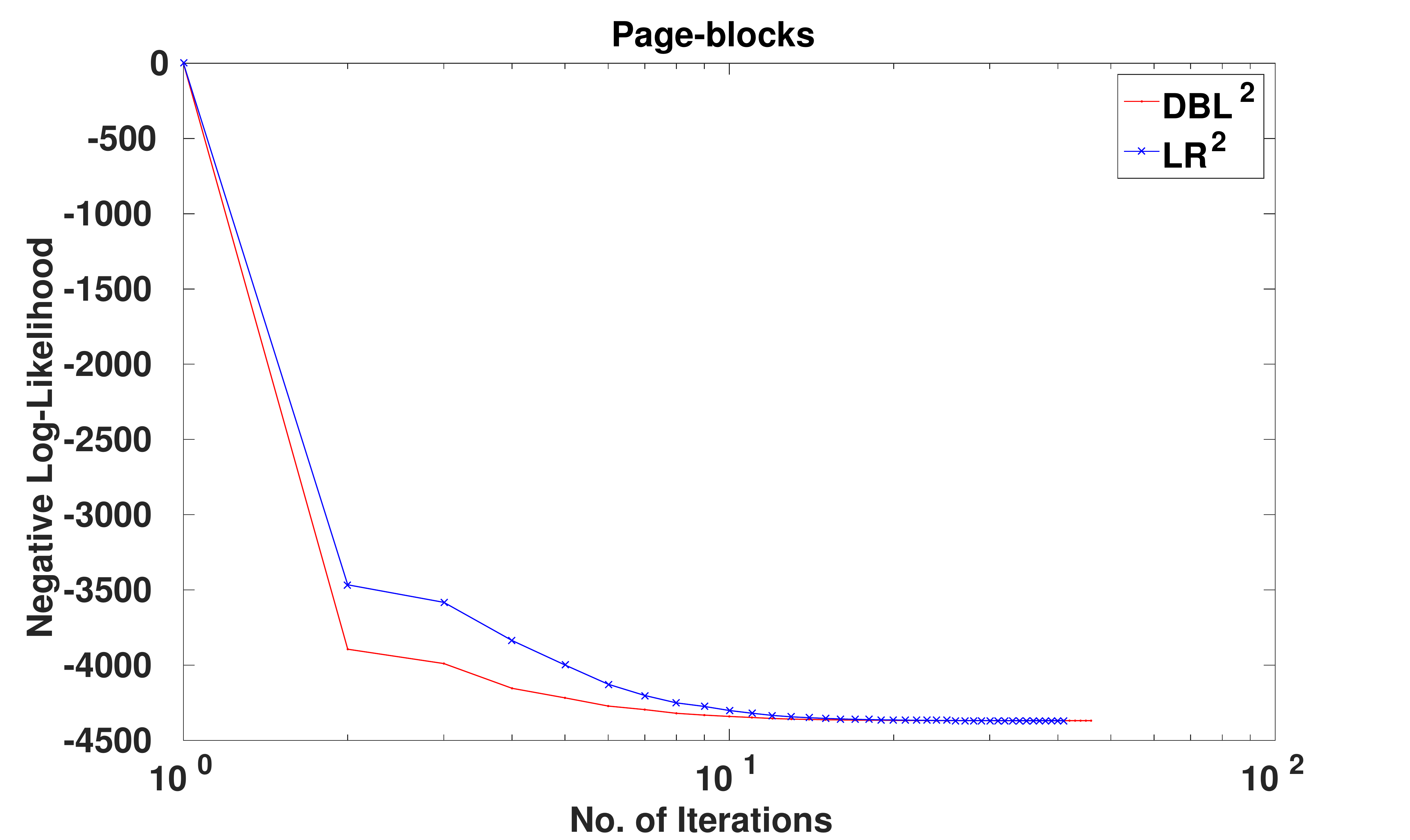}
\includegraphics[width=47mm,height=35mm]{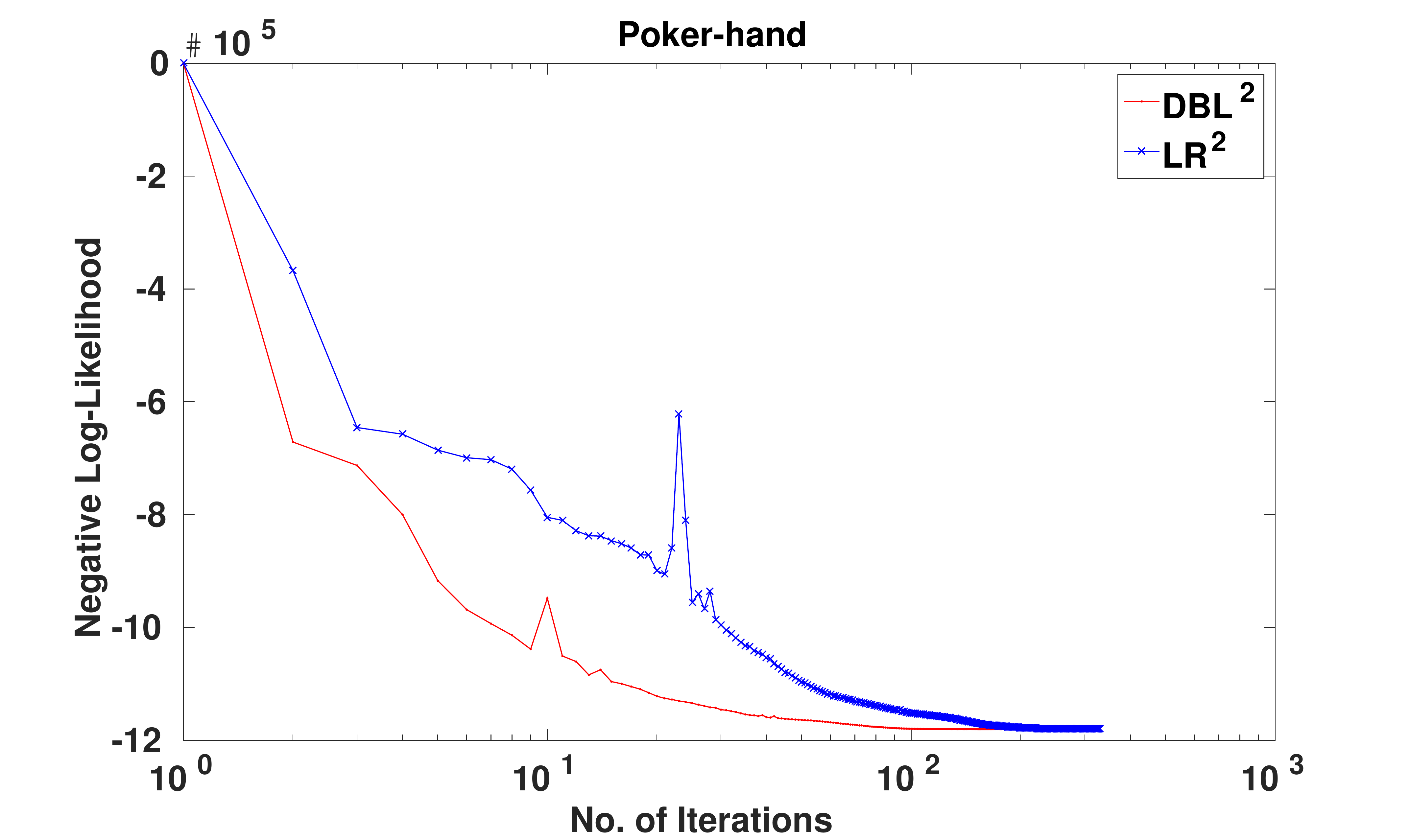}
\includegraphics[width=47mm,height=35mm]{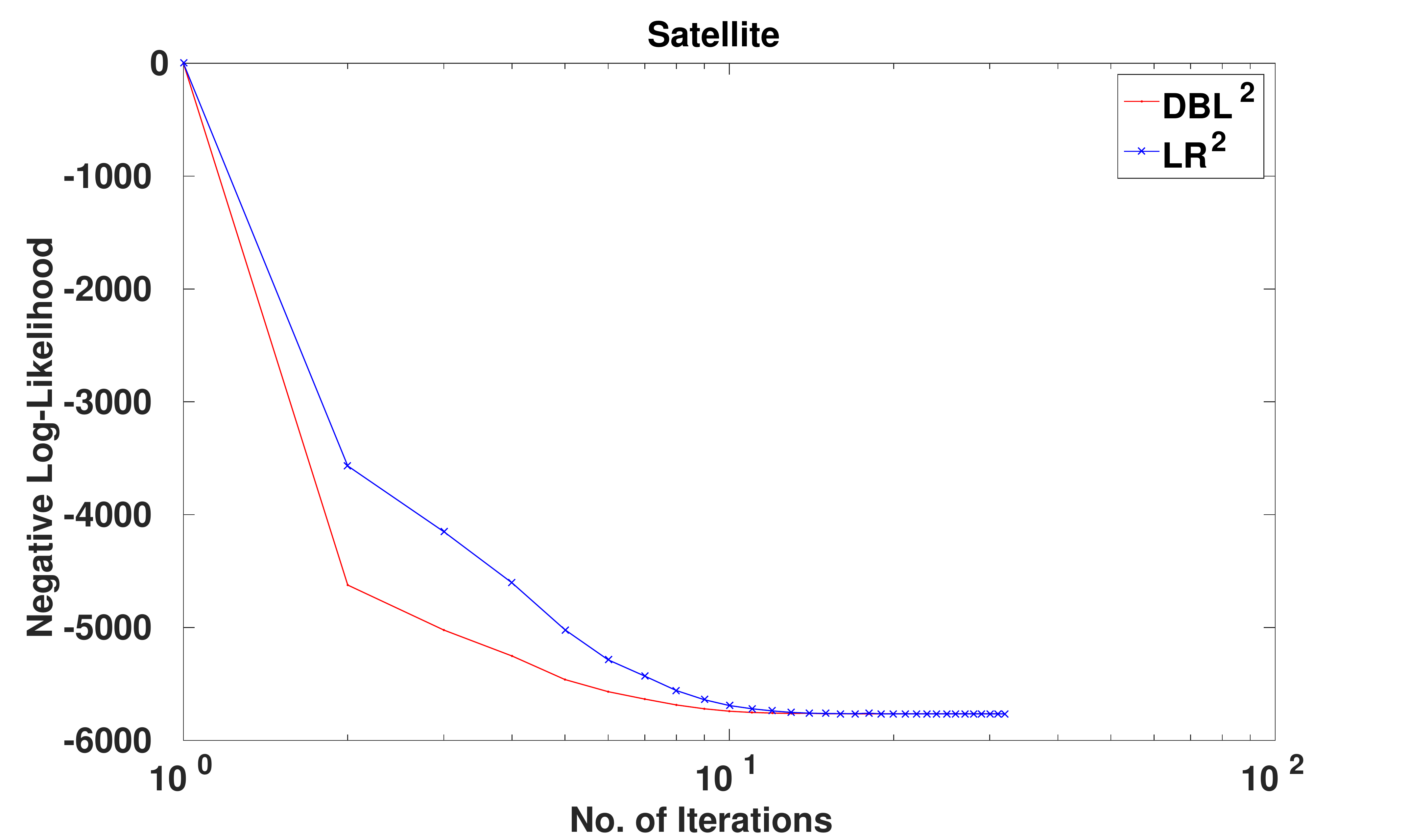}

\includegraphics[width=47mm,height=35mm]{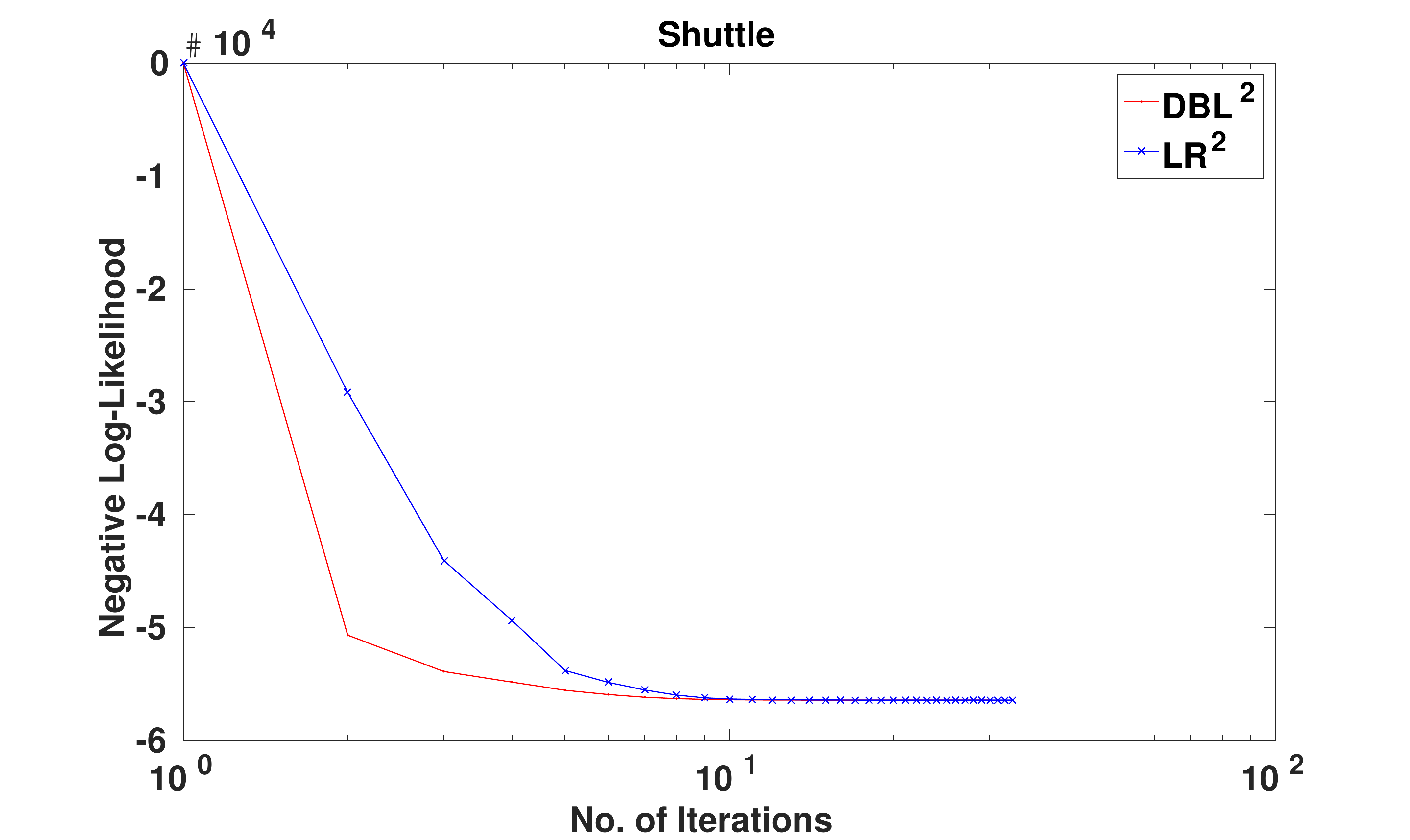}
\includegraphics[width=47mm,height=35mm]{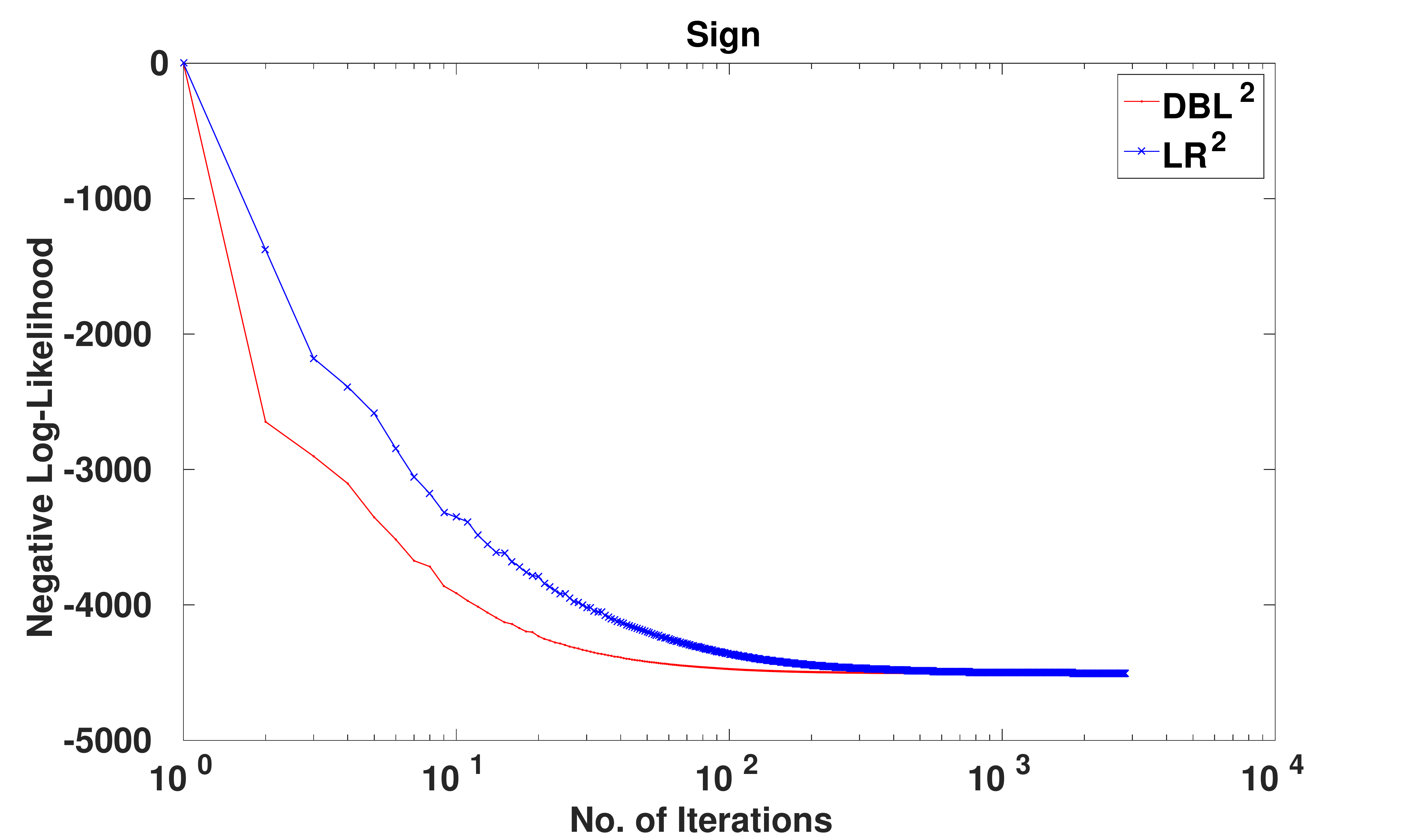}
\includegraphics[width=47mm,height=35mm]{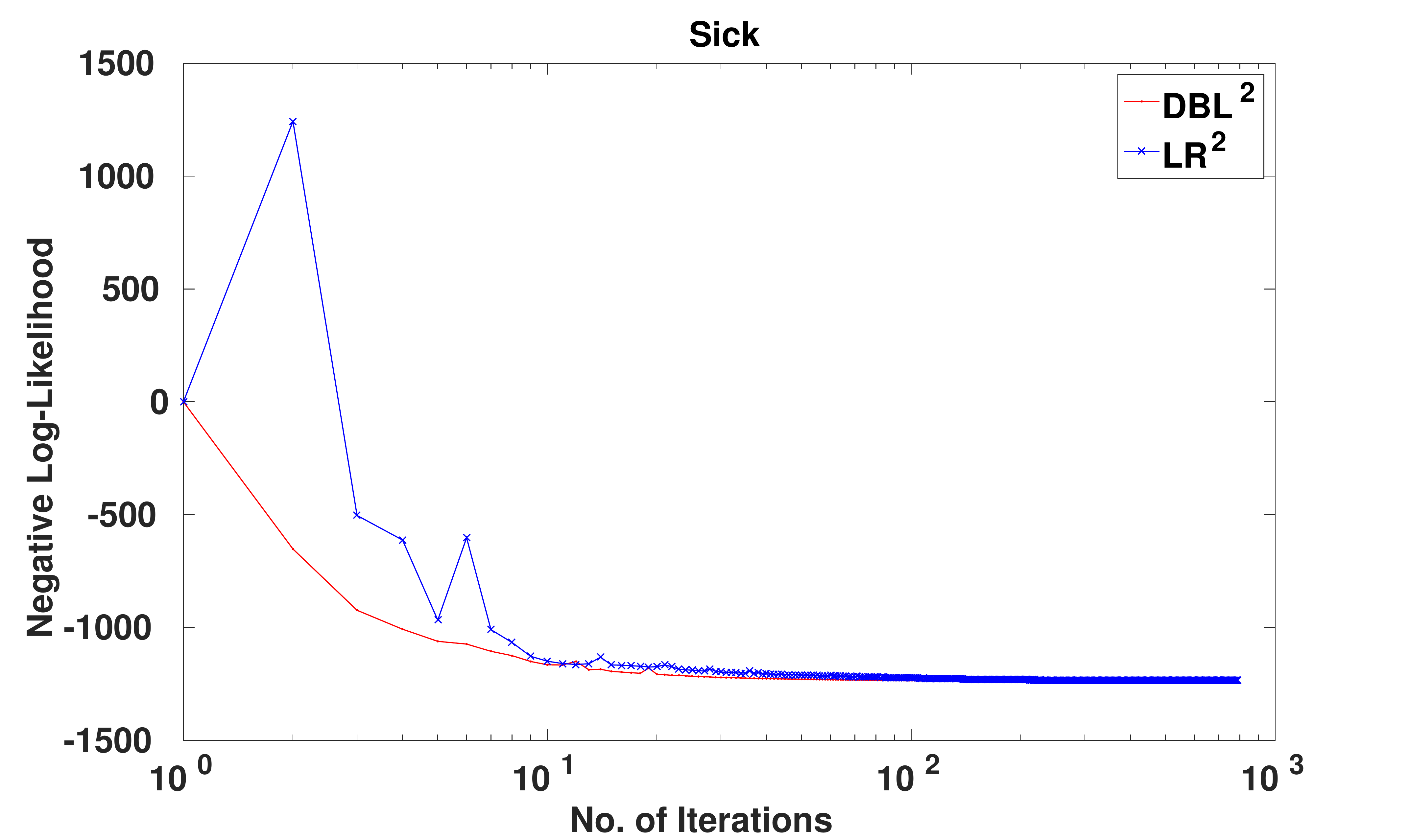}

\caption{\small Comparison of rate of convergence of \DBL^2 and \LR^2 on several datasets. The X-axis (No. of iterations) is on log scale.}
\label{fig_CCwvsdA2JE}
\end{figure}

\begin{figure}[h] 
\centering
\includegraphics[width=47mm,height=35mm]{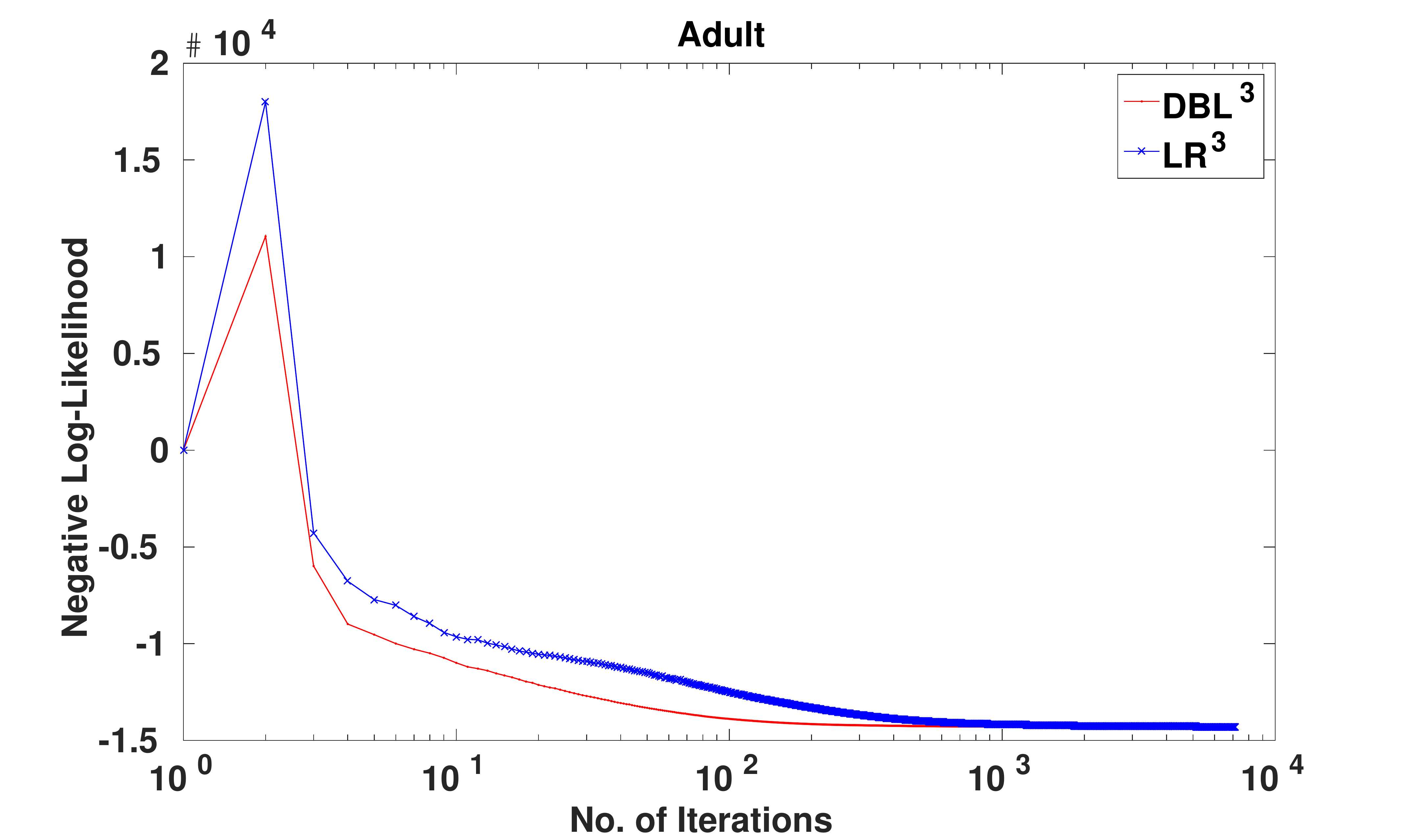}
\includegraphics[width=47mm,height=35mm]{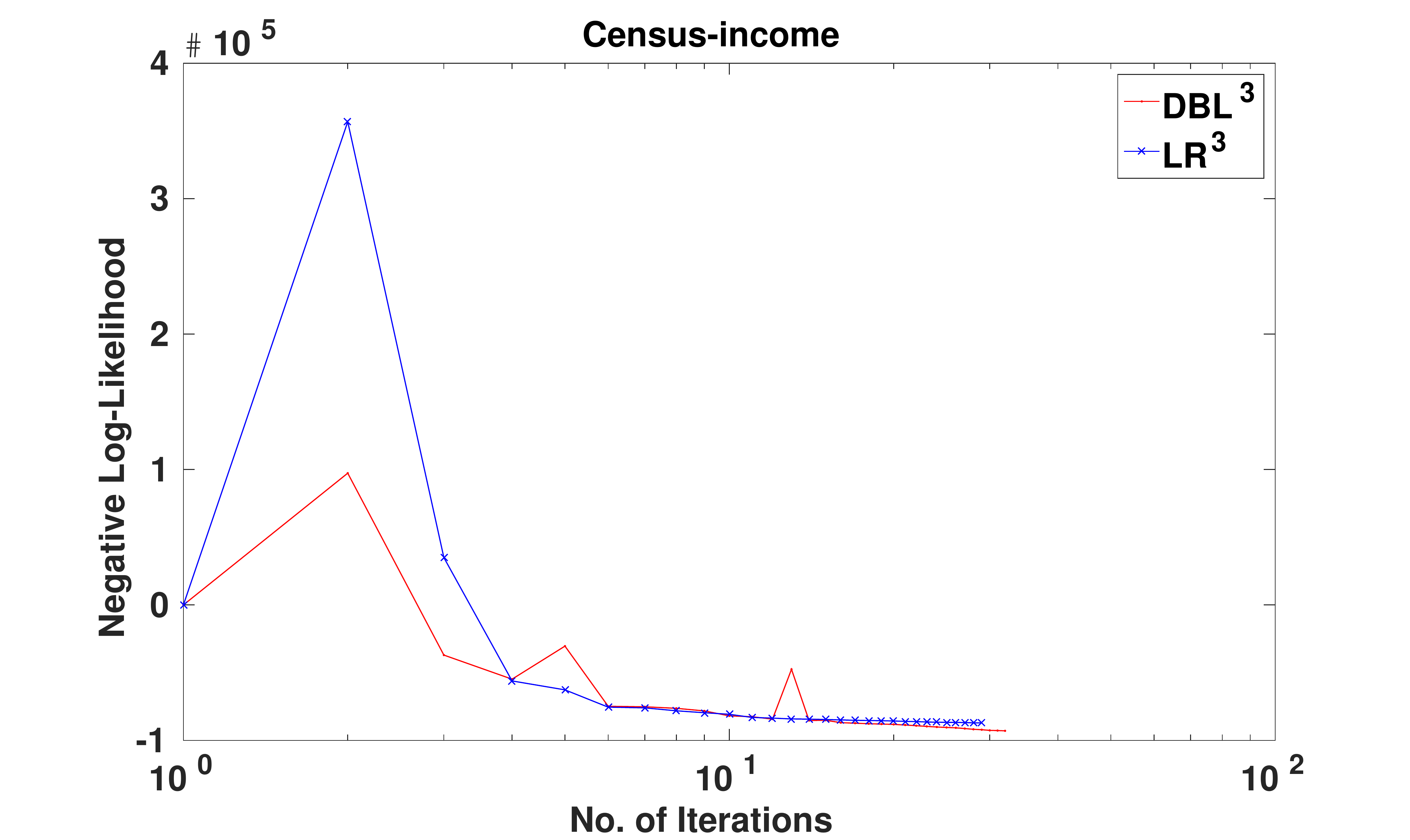}
\includegraphics[width=47mm,height=35mm]{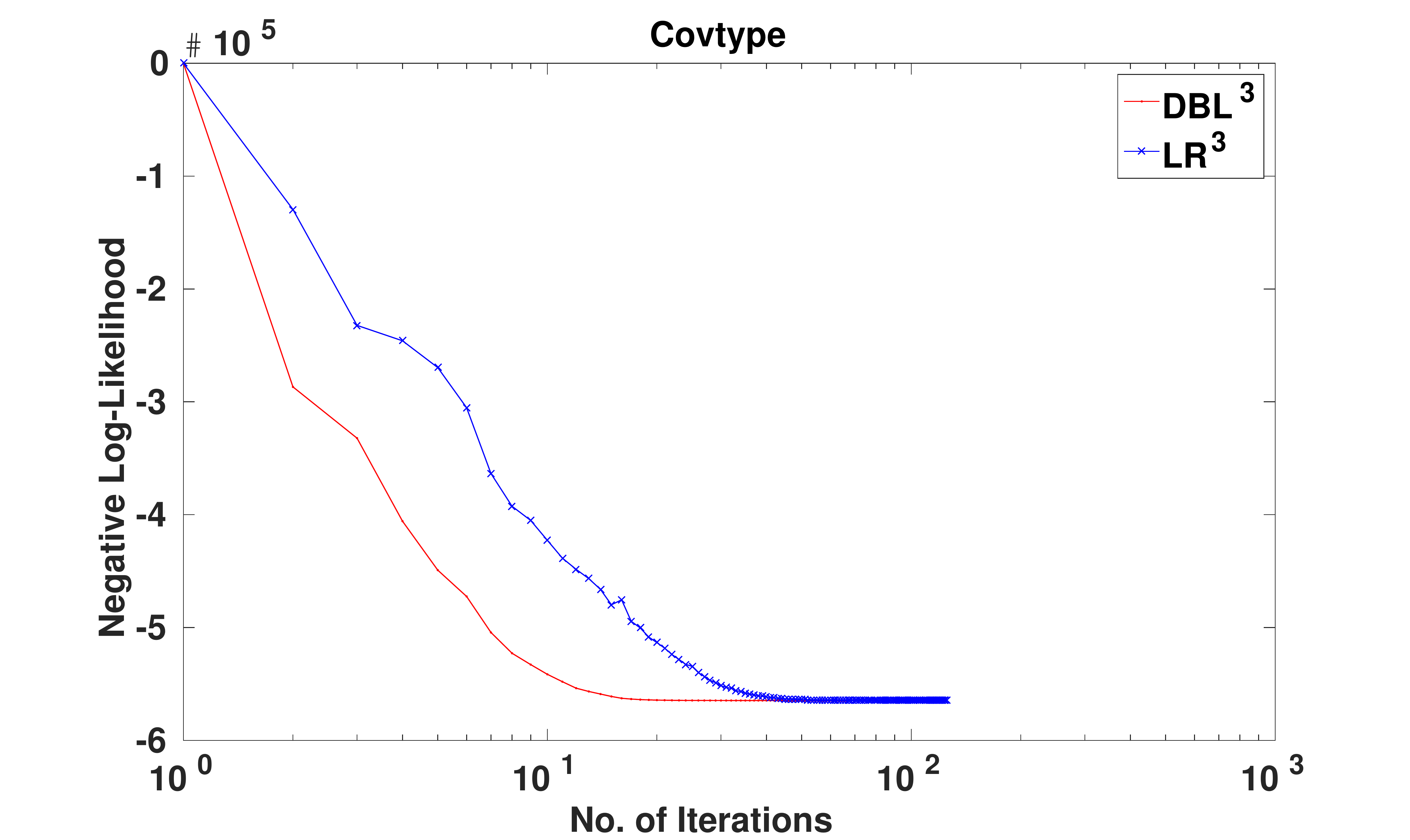}

\includegraphics[width=47mm,height=35mm]{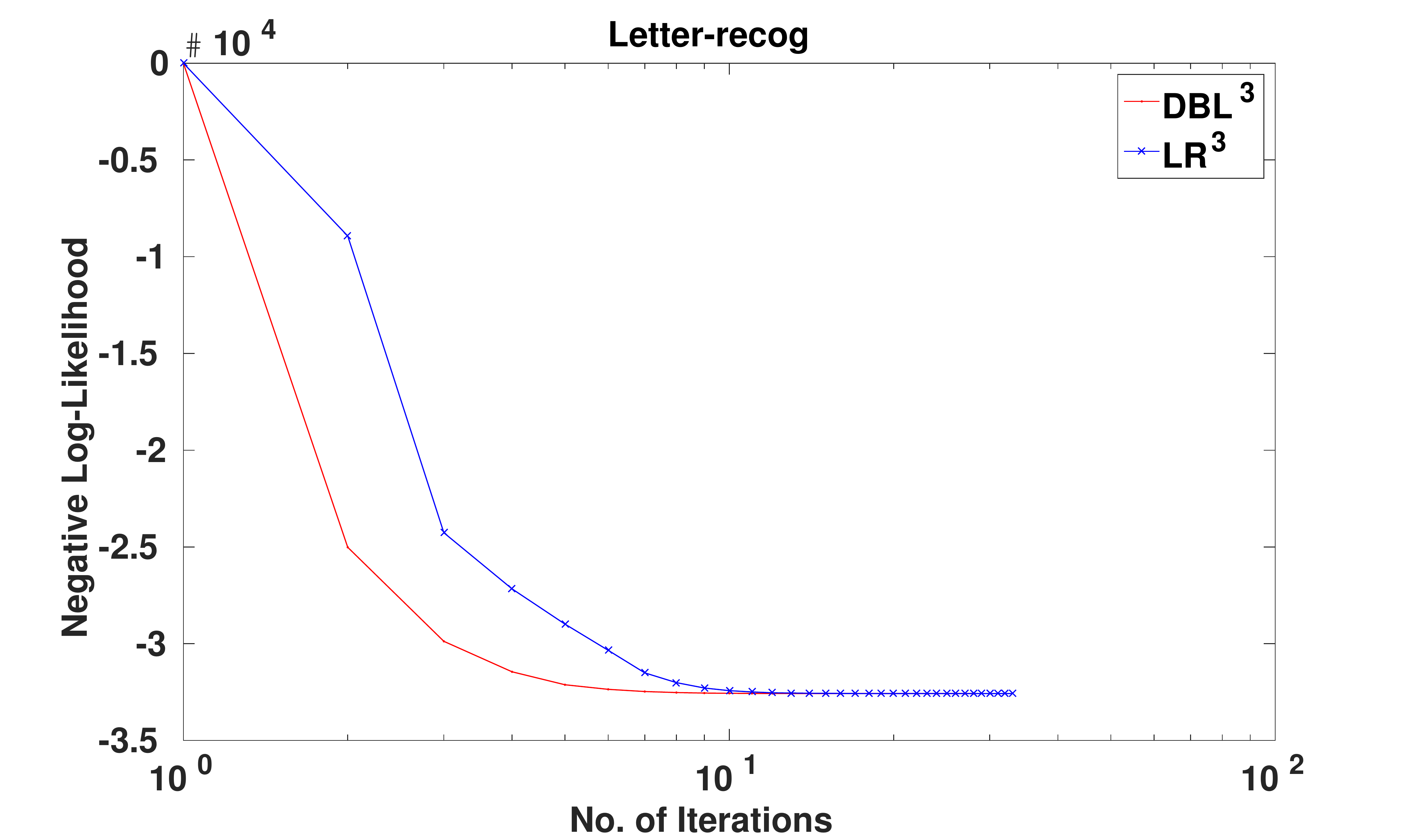}
\includegraphics[width=47mm,height=35mm]{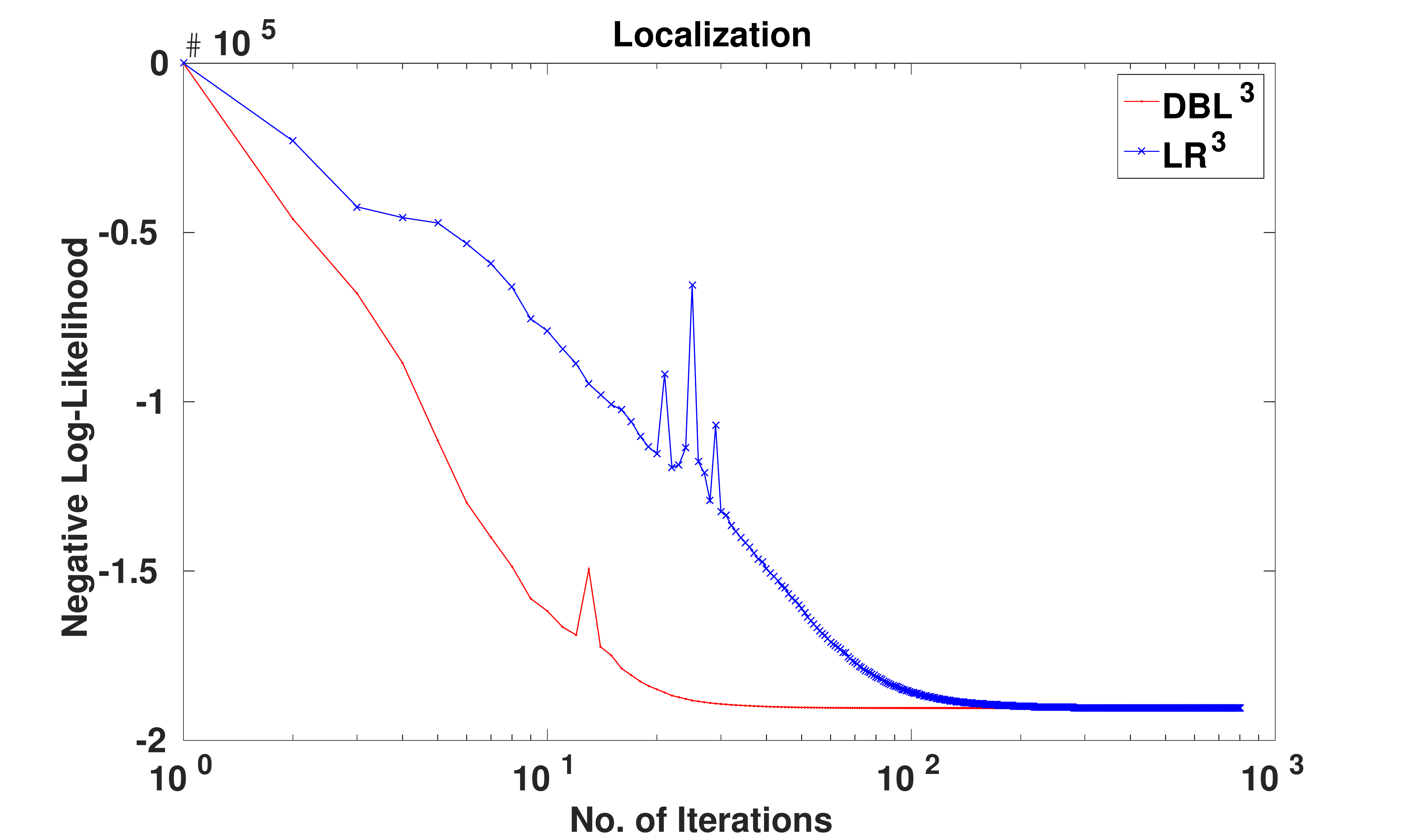}
\includegraphics[width=47mm,height=35mm]{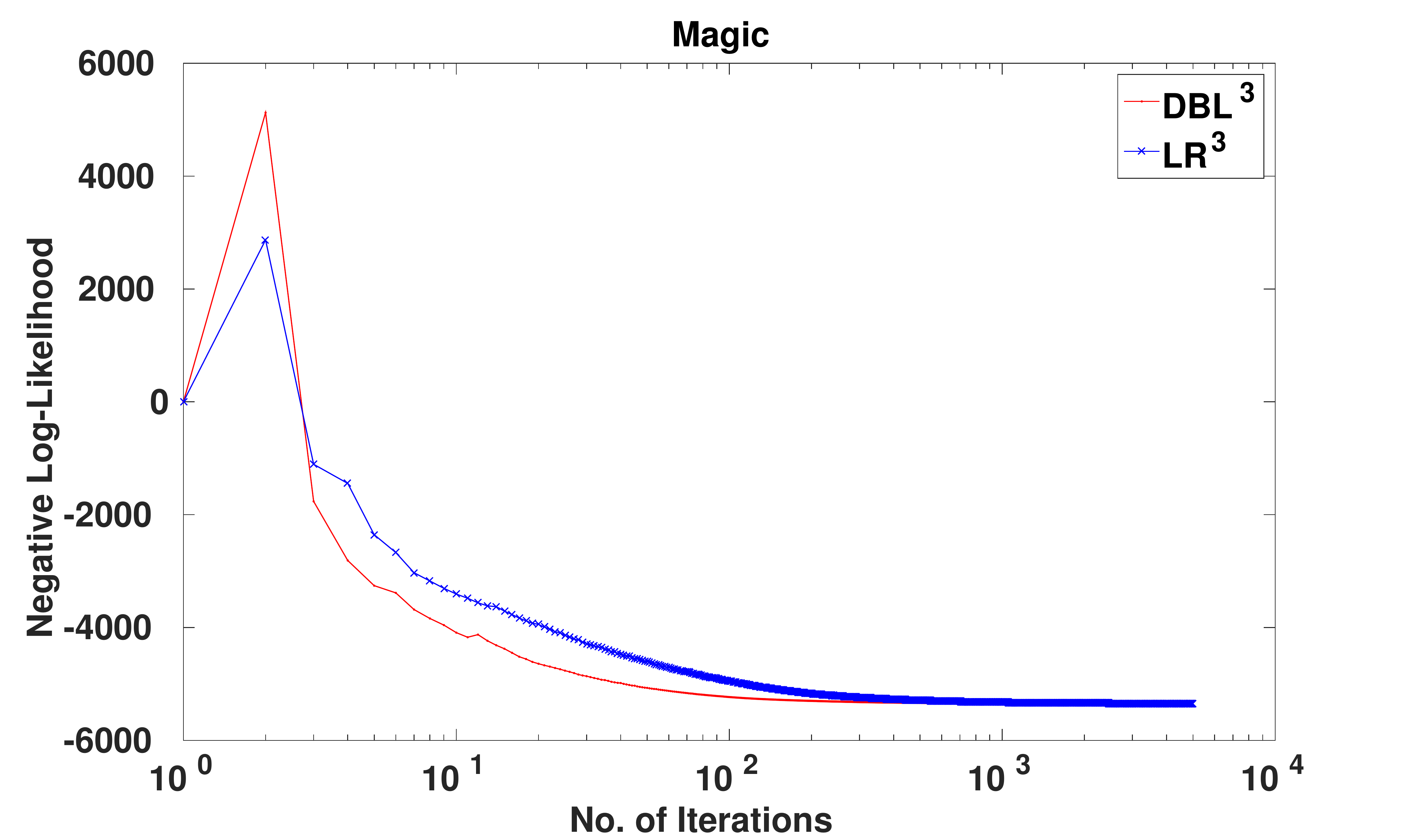}

\includegraphics[width=47mm,height=35mm]{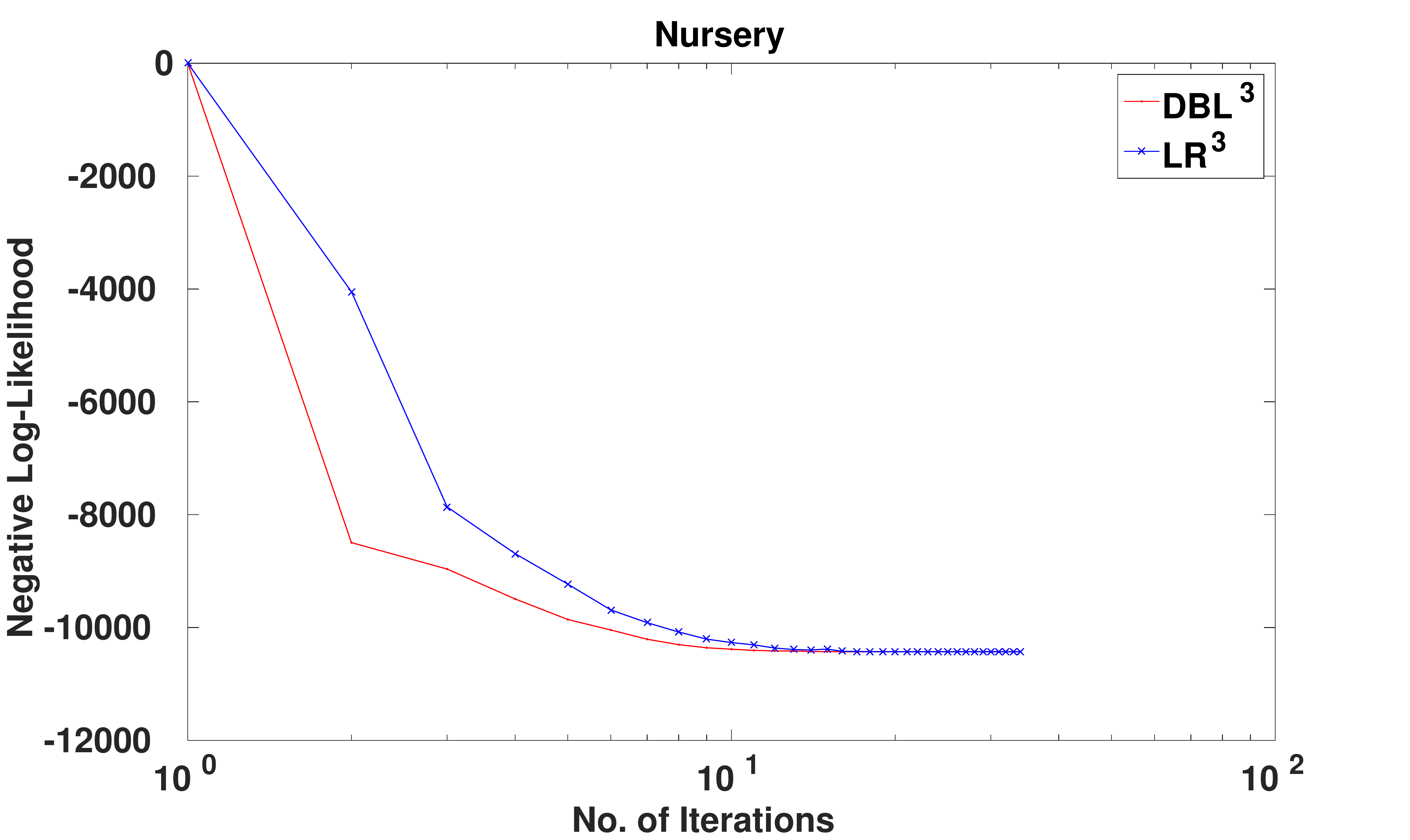}
\includegraphics[width=47mm,height=35mm]{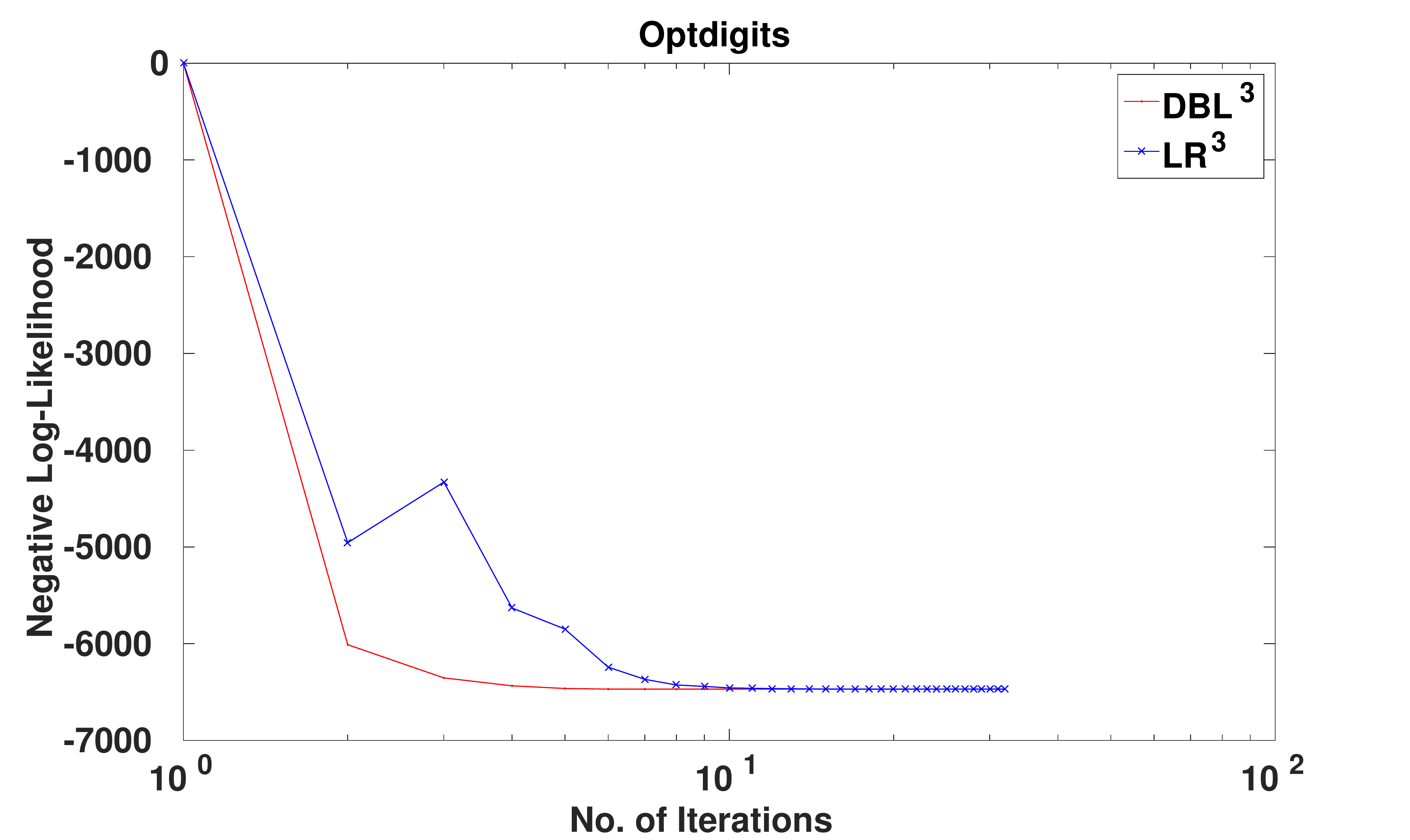}
\includegraphics[width=47mm,height=35mm]{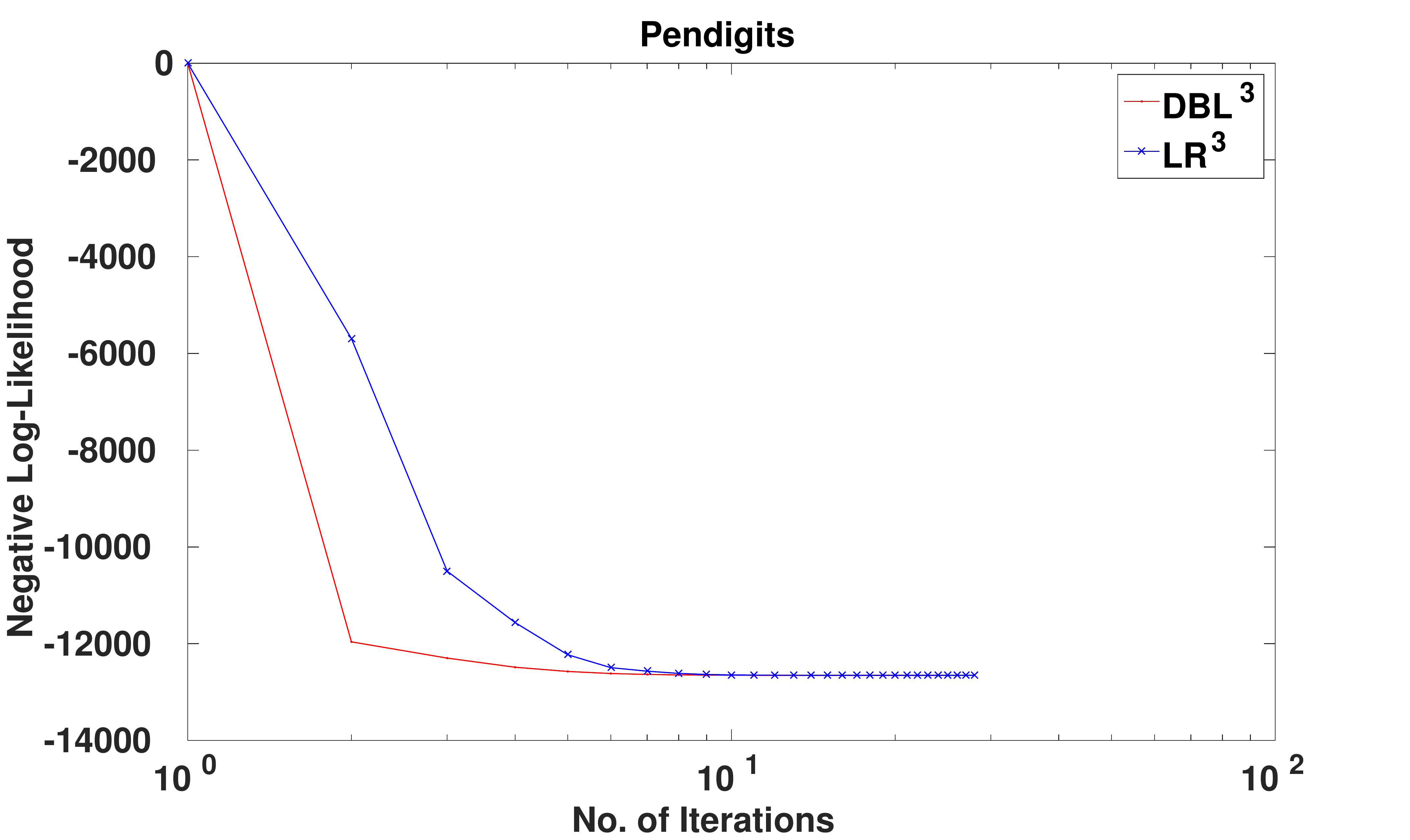}

\includegraphics[width=47mm,height=35mm]{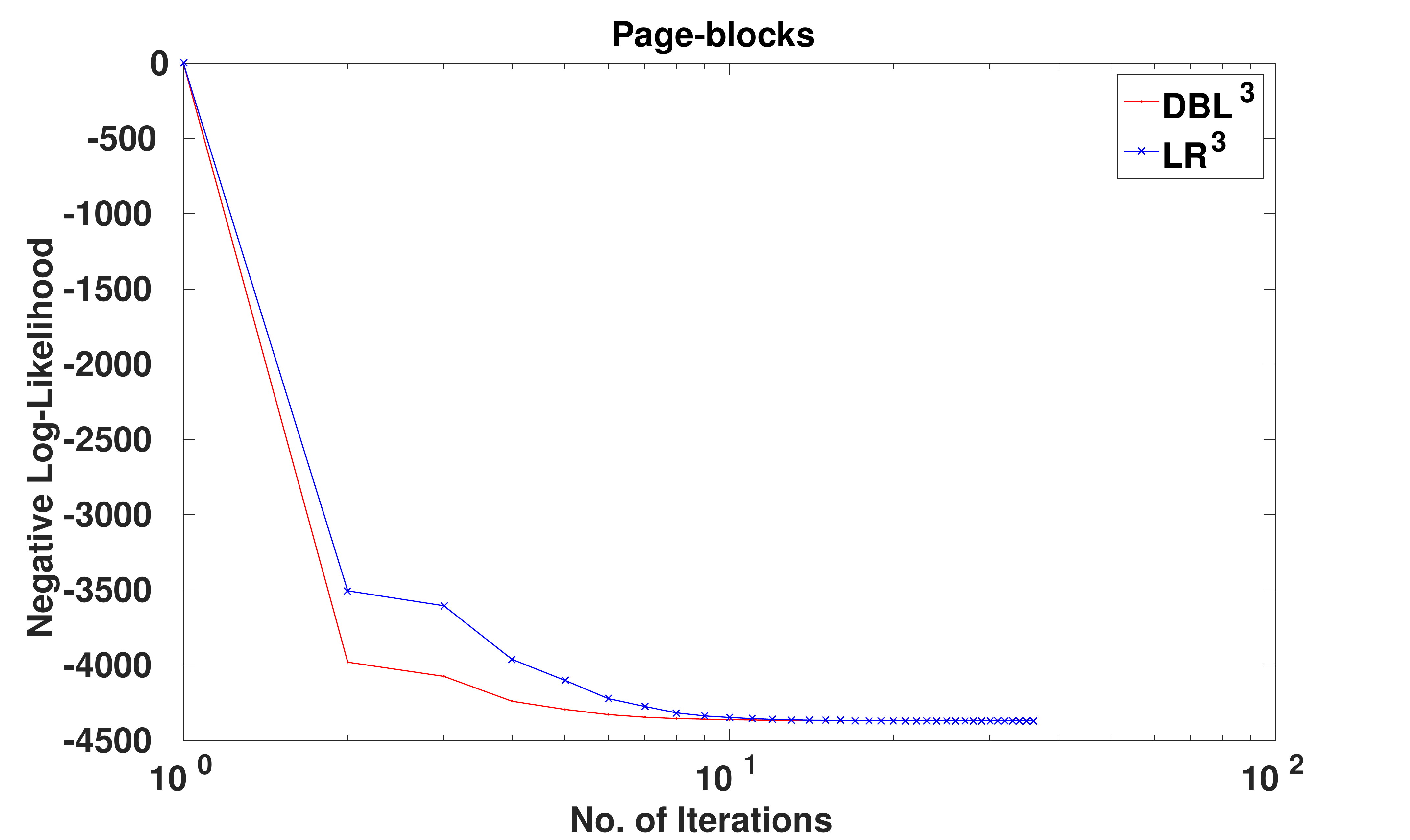}
\includegraphics[width=47mm,height=35mm]{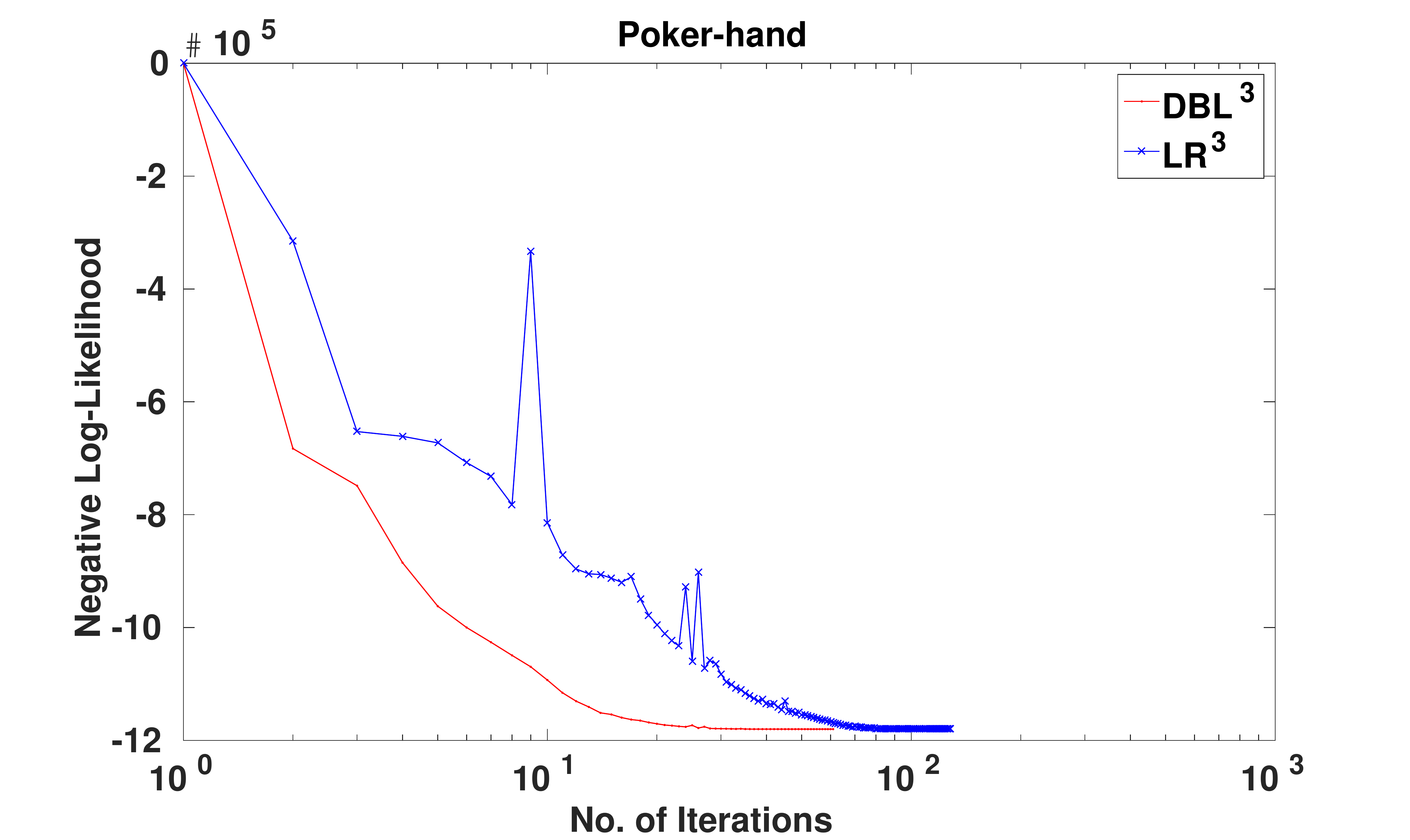}
\includegraphics[width=47mm,height=35mm]{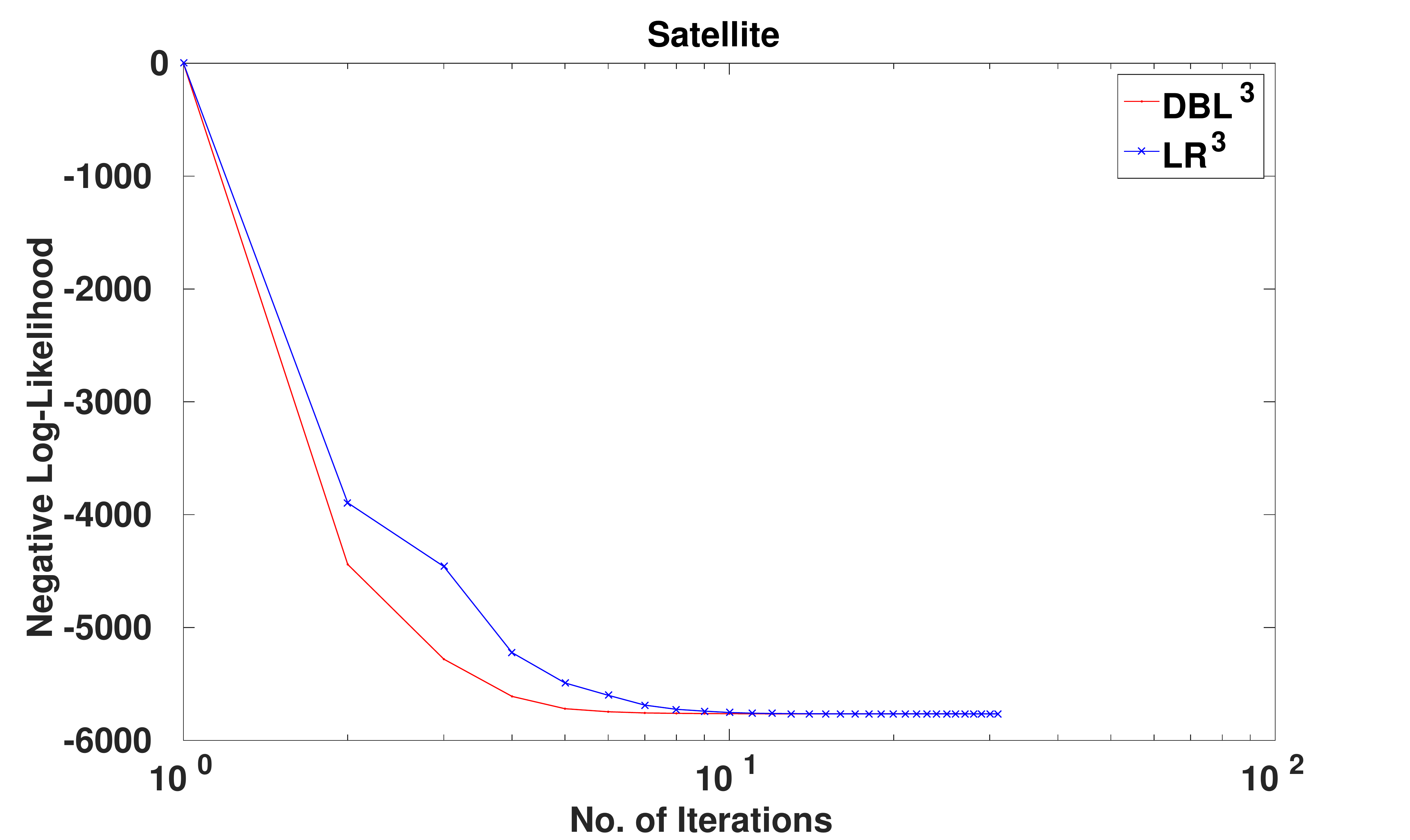}

\includegraphics[width=47mm,height=35mm]{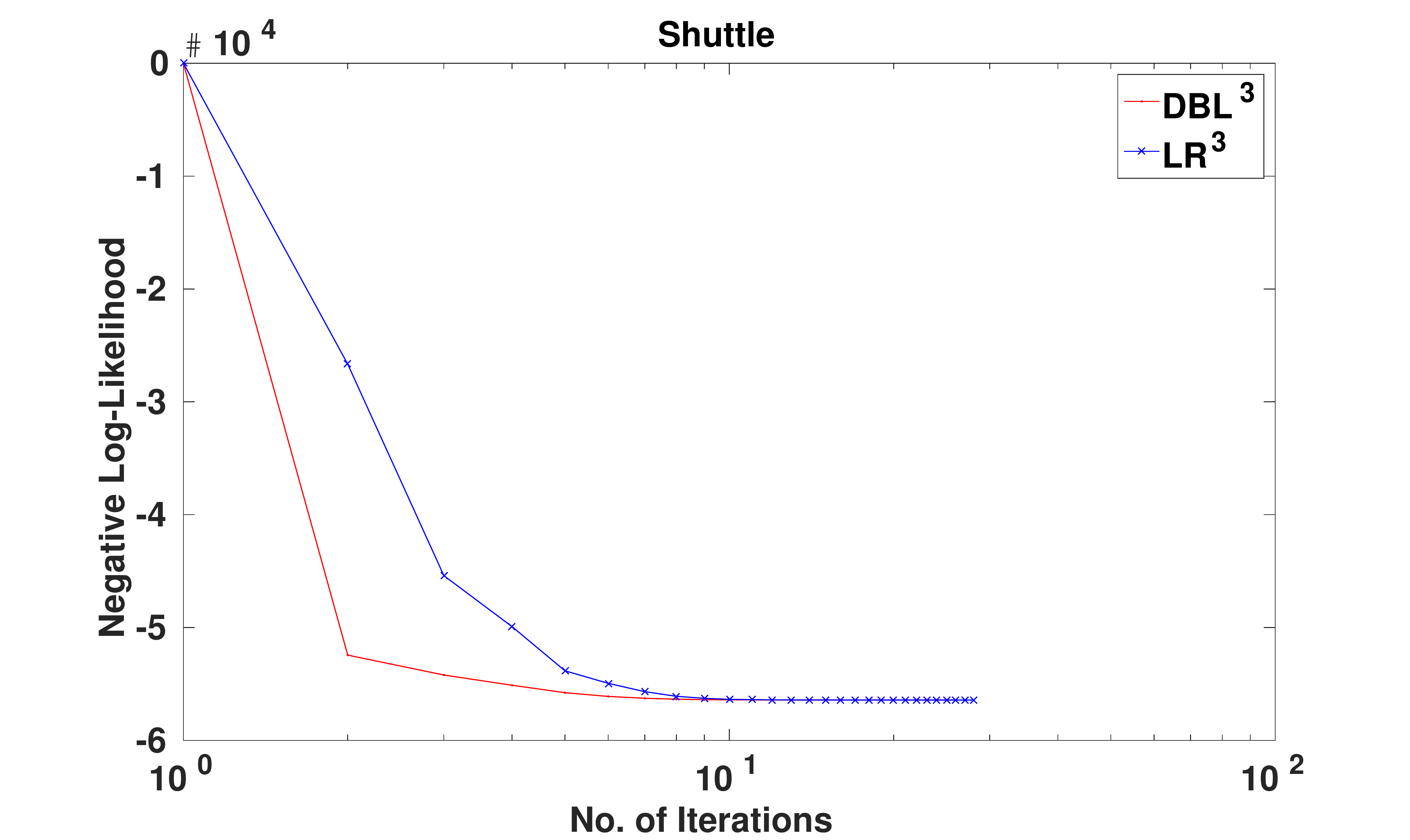}
\includegraphics[width=47mm,height=35mm]{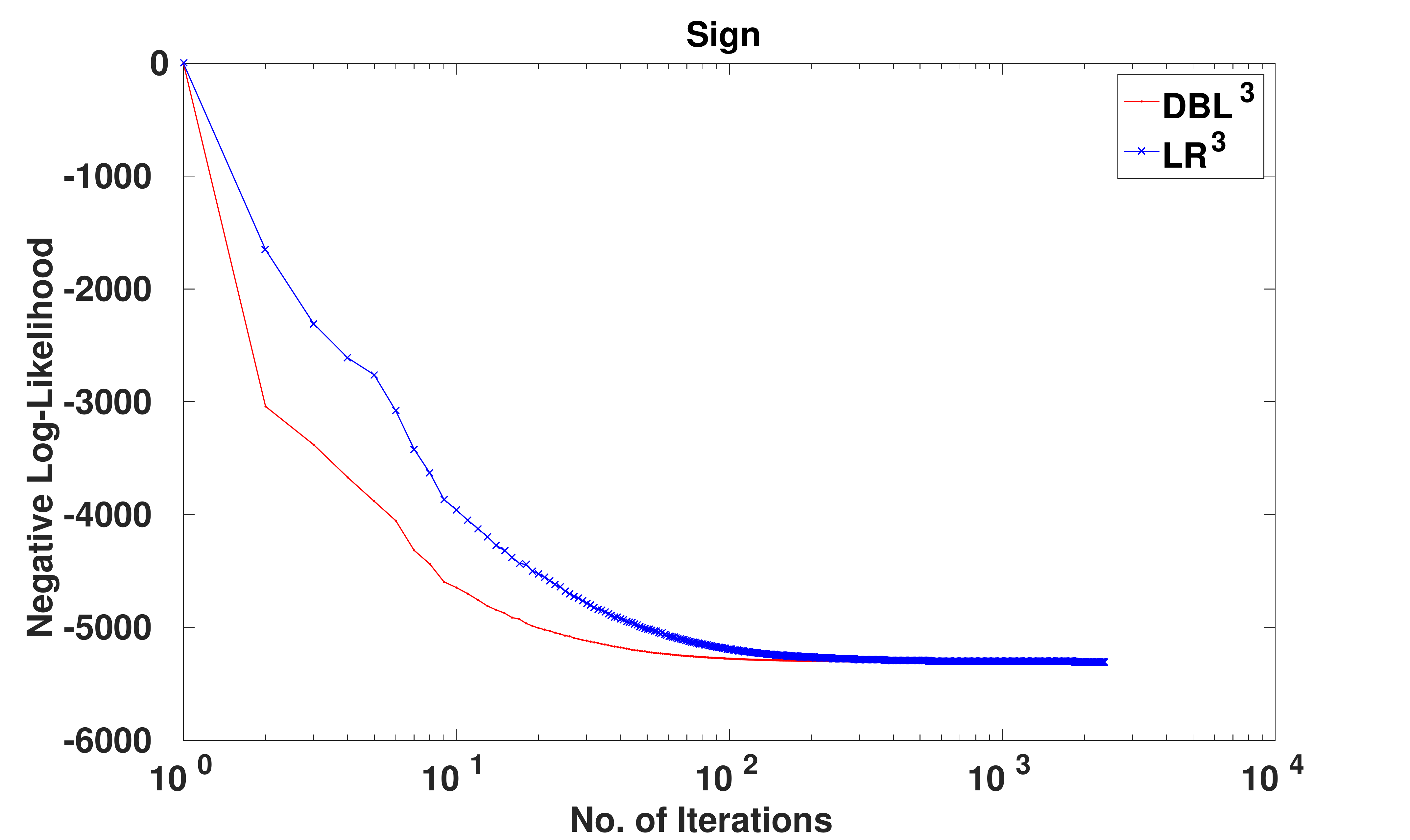}
\includegraphics[width=47mm,height=35mm]{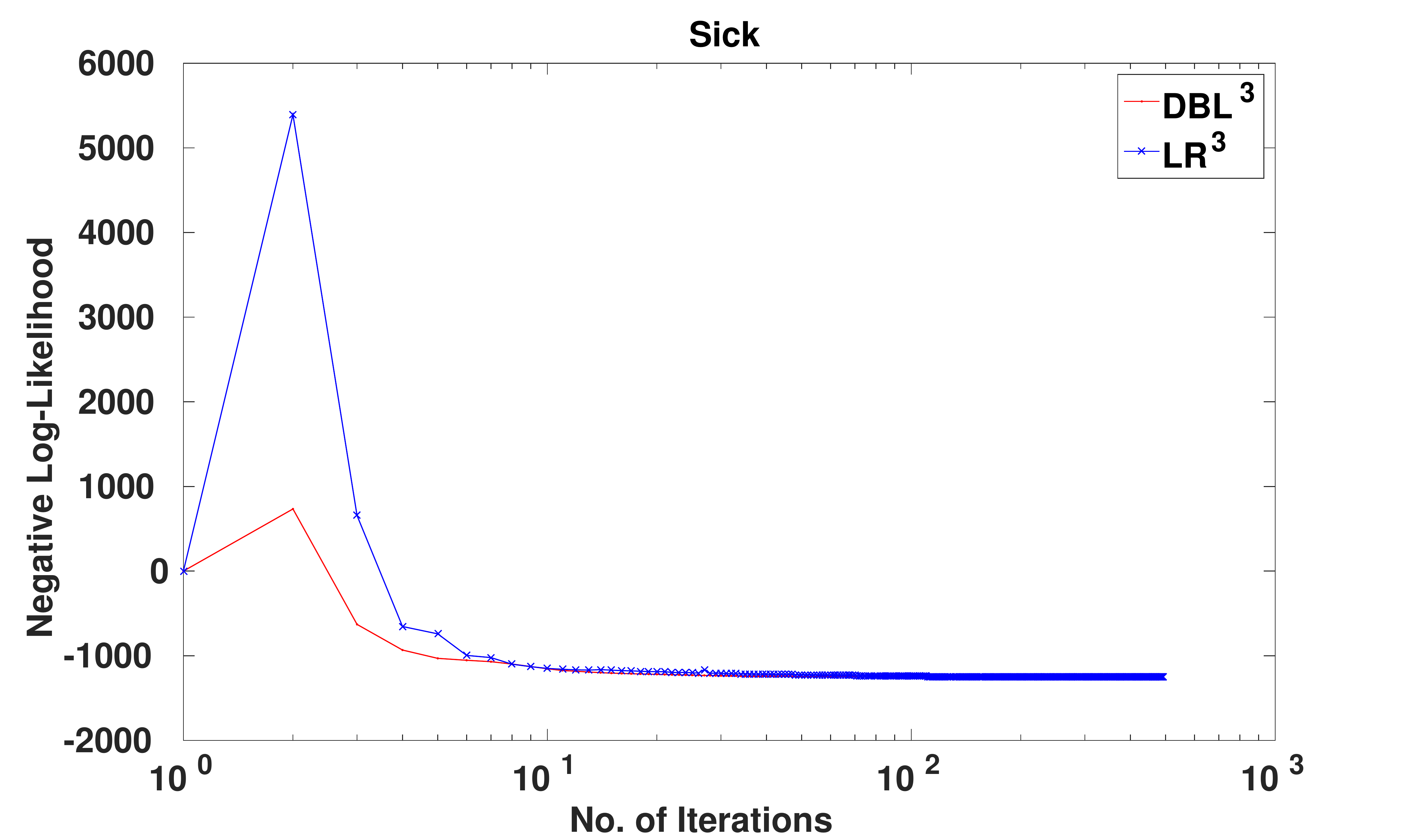}

\caption{\small Comparison of rate of convergence of \DBL^3 and \LR^3 on several datasets. The X-axis (No. of iterations) is on log scale.}
\label{fig_CCwvsdA3JE}
\end{figure}

A comparison of rate of convergence of Negative-Log-Likelihood (NLL) of \DBL^2 and \LR^2 parameterization on some sample datasets is shown in Figure~\ref{fig_CCwvsdA2JE}. 
%These results are averaged over different rounds of cross-validation.
It can be seen that, \DBL^2 has a steeper curve, asymptoting to its global minimum much faster. 
For example, on almost all datasets, one can see that \DBL^2 follows a steeper, hence more desirable, path toward convergence. 
This is extremely advantageous when learning from very few iterations (for example, when learning using Stochastic Gradient Descent based optimization) and, therefore, is a desirable property for scalable learning.
% --------------------------
% --------------------------
A similar trend can be seen in Figure~\ref{fig_CCwvsdA3JE} for \DBL^3 and \LR^3. 
% --------------------------

\begin{figure}
\centering
\includegraphics[width=.32\linewidth]{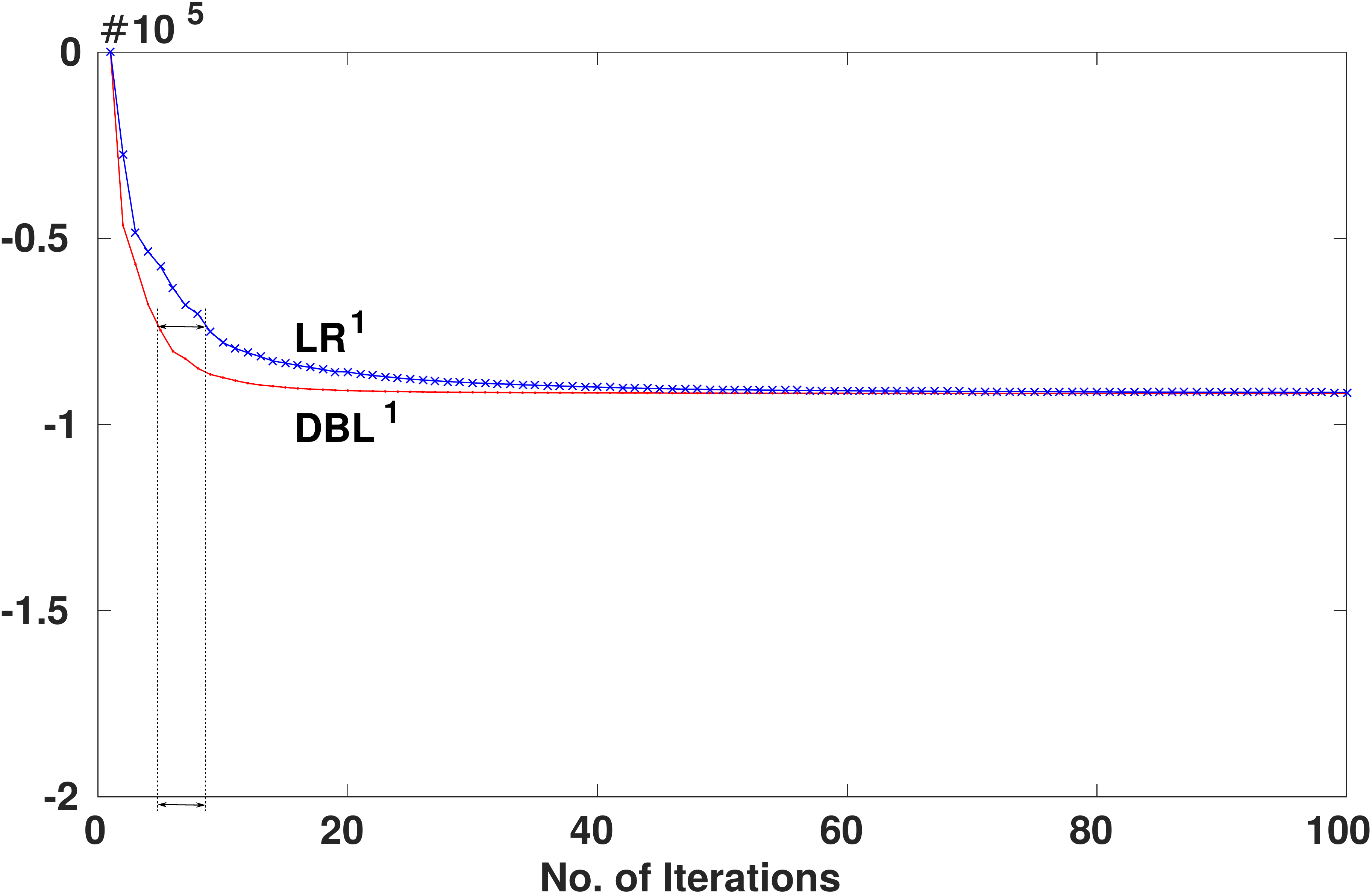}
\includegraphics[width=.32\linewidth]{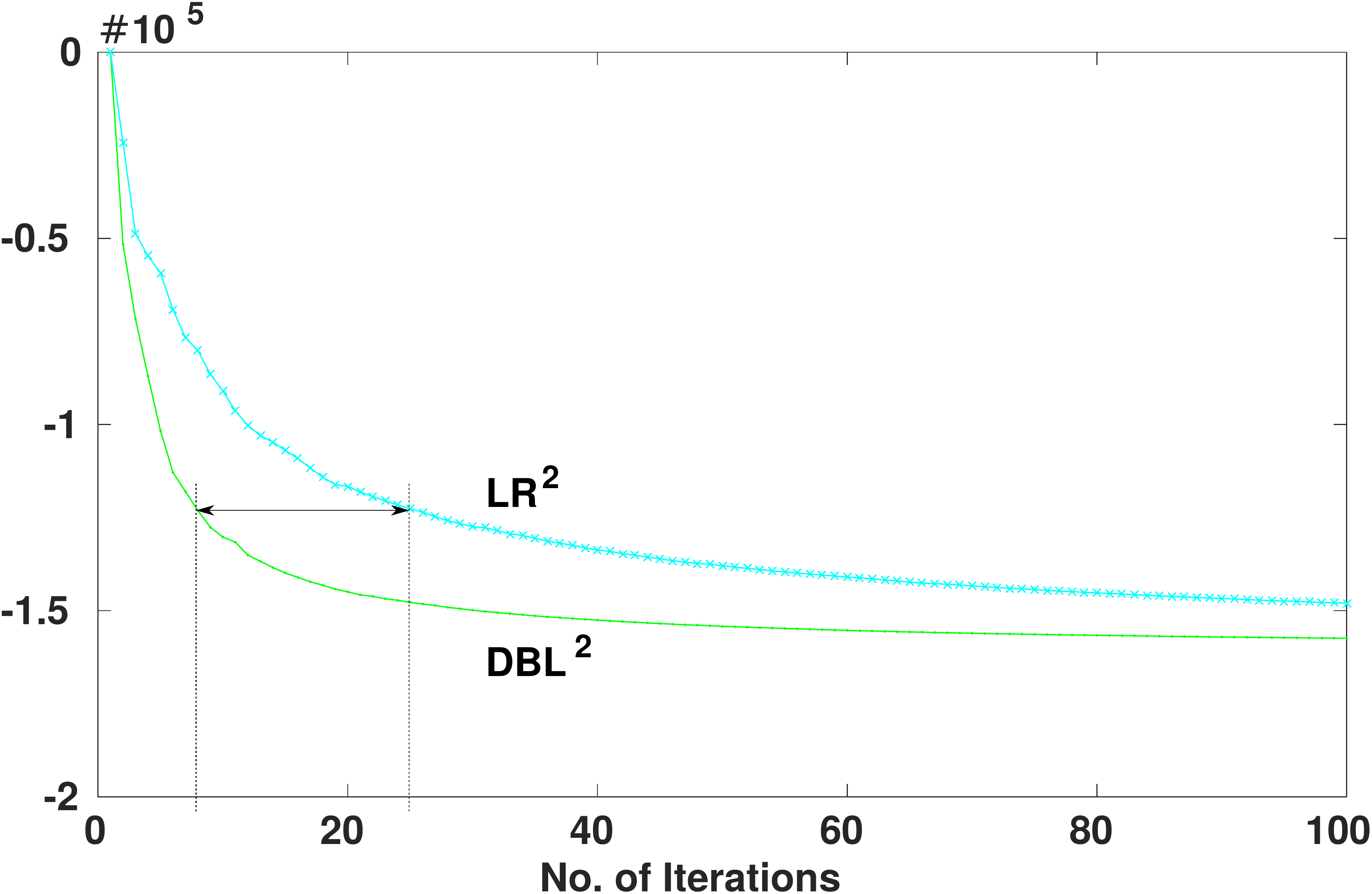}
\includegraphics[width=.32\linewidth]{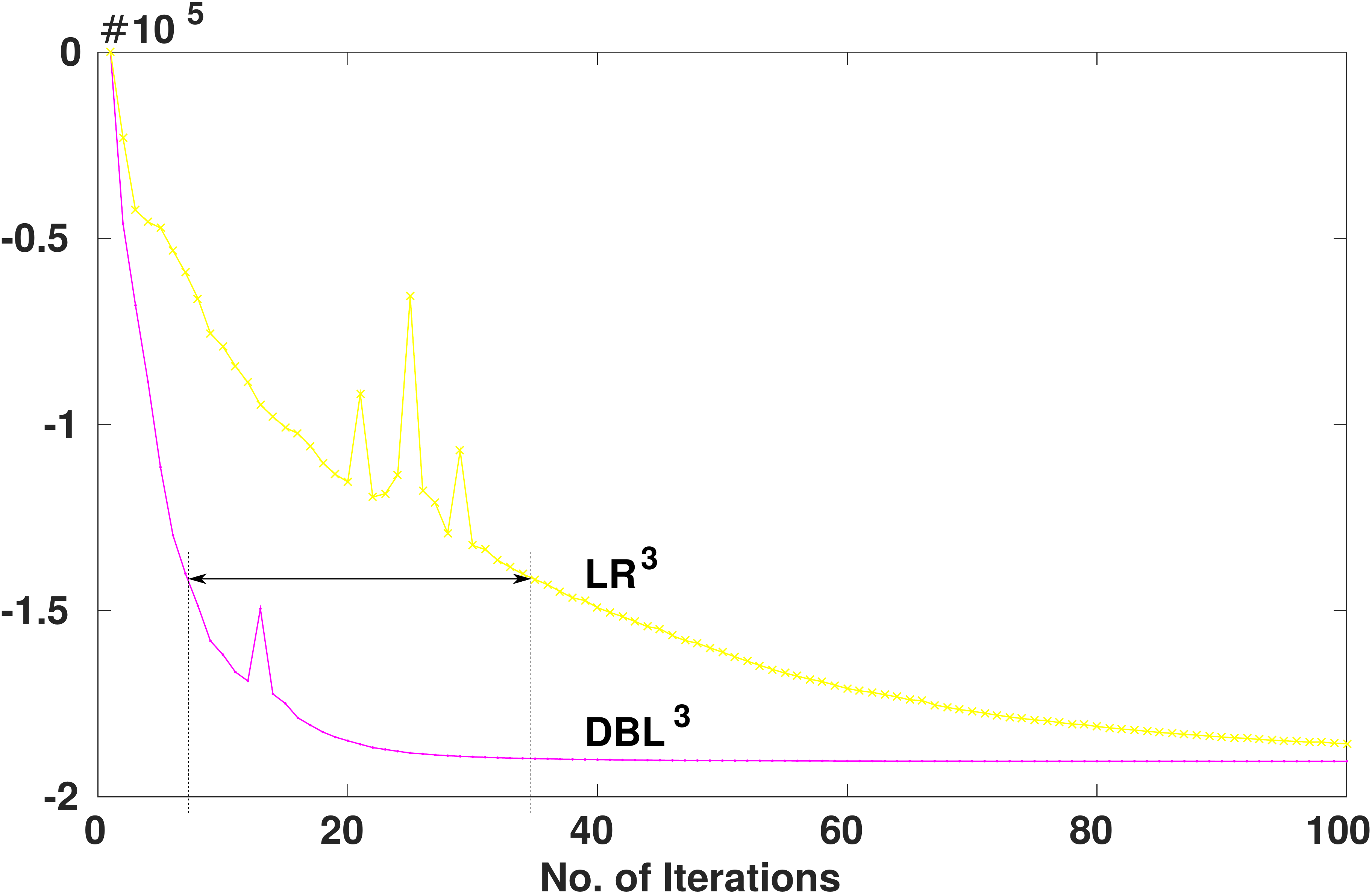}

\caption{\small Compared rates of convergence of \DBL^n vs. \LR^n, for $n=1,2,3$ on the sample \texttt{Localization} dataset. Y-axis is the negative log-likelihood.}
\label{fig_CCwvsdAnJE}
\end{figure}

% --------------------------
Finally, let us present some comparison results about the speed of convergence of \DBL^n vs.\ \LR^n as we increase $n$. 
In Figure~\ref{fig_CCwvsdAnJE}, we compare the convergence for $n=1$, $n=2$ and $n=3$ on the sample \texttt{Localization} dataset. 
It can be seen that the improvement that \DBL^n provides over \LR^n gets better as we go to deeper structures, i.e., as $n$ becomes larger. 
%For $n=1$, \DBL^1 reaches the same level of negative log-likelihood with KKK fewer iterations. 
%For $n=2$, this difference increases to KKK iterations. 
%For $n=3$, this difference increases to KKK iterations. 
The similar behaviour was observed for several datasets and, although studying rates of convergence is a complicated matter and is outside the scope of this work, we anticipate this phenomenon to be an interesting venue of investigation for future work. 
%This is especially true for large datasets because they require low-bias algorithms and hence high $n$. 
% --------------------------
% --------------------------

%%%%%%%%%%%%%%%%%%%%%%%%%%%%%%%%%%%%%%%%%%%%%%%%%%%%%%%%%%%%%%%%%
\subsection{\DBL^n vs. Random Forest}
%%%%%%%%%%%%%%%%%%%%%%%%%%%%%%%%%%%%%%%%%%%%%%%%%%%%%%%%%%%%%%%%%

The two \DBL^n models are compared in terms of W-D-L of 0-1 Loss, RMSE, bias and variance with Random Forest in Table~\ref{tab_wAnJEvsRF}.  
On \emph{Little} datasets, it can be seen that \DBL^n has significantly lower bias than RF. The variance of \DBL^3 is significantly higher than RF, whereas, difference in the variance is not significant for \DBL^2 and RF. 
0-1 Loss results of \DBL^n and RF are similar. However, RF has better RMSE results than \DBL^n on \emph{Little} datasets. 
On \emph{Big} datasets, \DBL^n wins on majority of datasets in terms of 0-1 Loss and RMSE.
% --------------------------
\begin{table} \scriptsize
\begin{tabular}{lcccc}
\cline{1-5}
&\multicolumn{2}{c}{\bf \DBL^2 vs. RF} & \multicolumn{2}{c}{\bf \DBL^3 vs. RF}  \\
\cmidrule {2-3}\cmidrule (l){4-5} 
&W-D-L& $p$& W-D-L& $p$\\
\cline{1-5}
&\multicolumn{4}{c}{Little Datasets} \\
\cline{1-5}
Bias	&51/3/21	&\bf $<$0.001	&52/2/21&\bf $<$0.001	\\
Variance&33/3/39	& 0.556		&28/5/42&\bf 0.119	\\
0-1 Loss&40/3/32	&0.409		&37/3/35& 0.906	\\
RMSE	&26/1/48	&0.014		&27/1/47& 0.026	\\
\cline{1-5}
&\multicolumn{4}{c}{Big Datasets} \\
\cline{1-5}
0-1 Loss&5/0/3	& 0.726		&6/0/2& 0.289	\\
RMSE	&5/0/3	& 0.726		&5/0/3& 0.726	\\
\cline{1-5}
\end{tabular}
\caption{\small Win-Draw-Loss: \DBL^2 vs. RF and \DBL^3 vs RF. $p$ is two-tail binomial sign test. Results are significant if $p \leq 0.05$.}
\label{tab_wAnJEvsRF}
\end{table}
% ---------------------------------

\begin{figure}[t] 
\centering
\includegraphics[width=55mm,height=40mm]{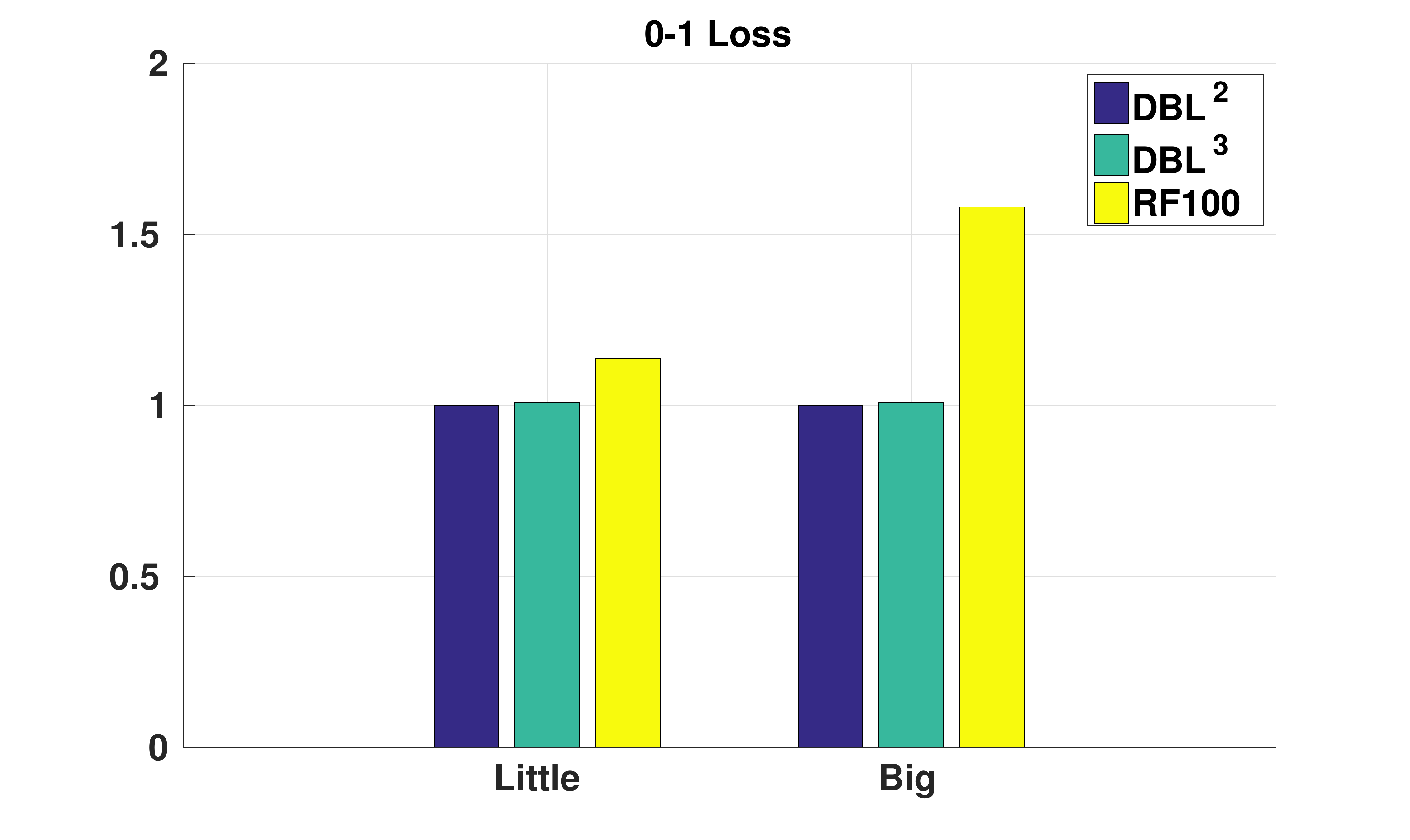}
\includegraphics[width=55mm,height=40mm]{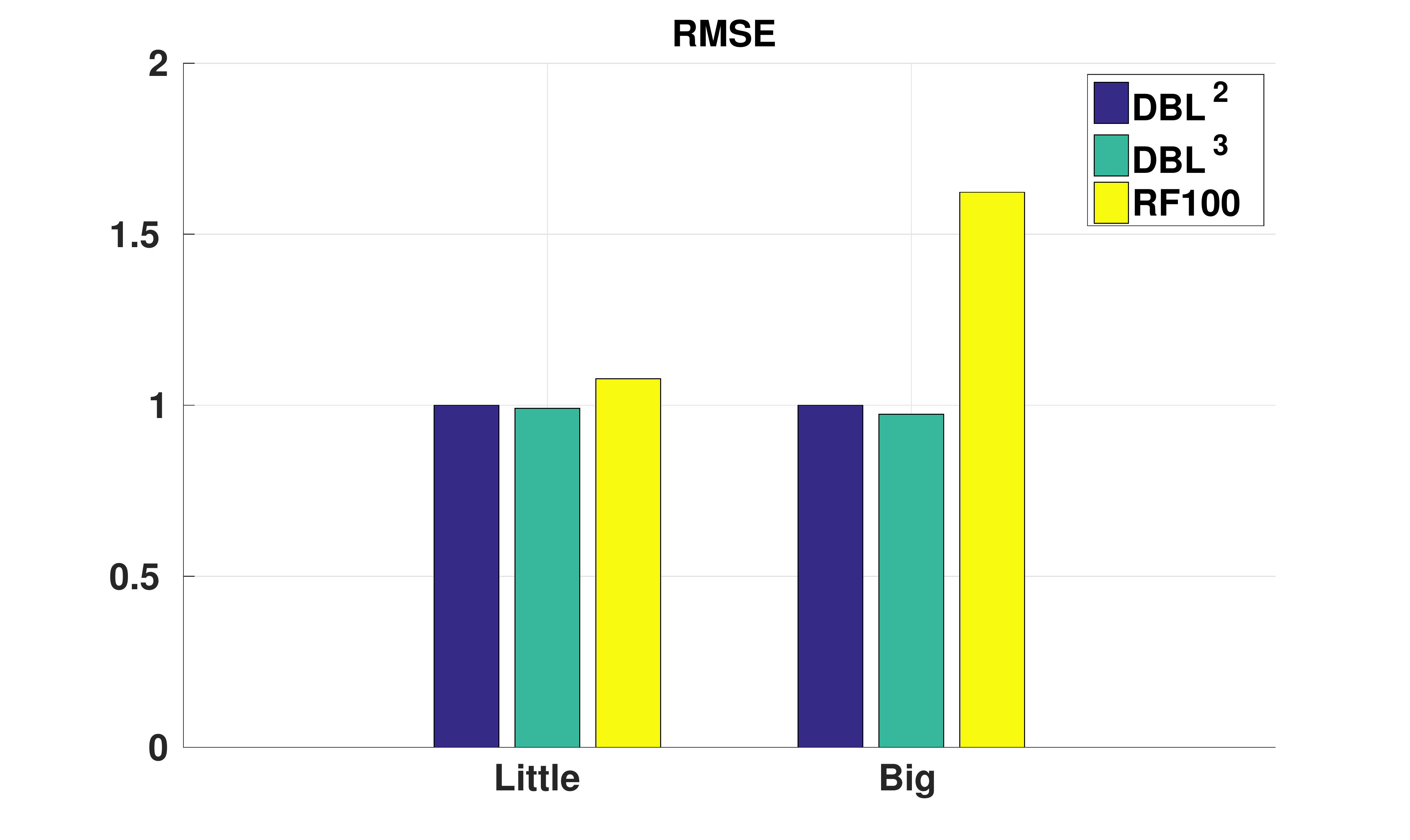}

%\includegraphics[width=55mm,height=40mm]{images2/RF100vsALR_Bias}
%\includegraphics[width=55mm,height=40mm]{images2/RF100vsALR_Var}
%\caption{\small Geometric mean of 0-1 Loss (Top Left), RMSE (Top Right), Bias (Bottom Left) and Variance (Bottom Right) performance of \DBL^2,  \DBL^3 and RF for \emph{Little} and \emph{Big} datasets.}
\caption{\small Geometric mean of 0-1 Loss (Left) and RMSE (Right) performance of \DBL^2,  \DBL^3 and RF for \emph{Little} and \emph{Big} datasets.}
\label{fig_wAnJEvsRF}
\end{figure}

The averaged 0-1 Loss and RMSE results are given in Figure~\ref{fig_wAnJEvsRF}. It can be seen that \DBL^2, \DBL^3 and RF have similar 0-1 Loss and RMSE across \emph{Little} datasets. 
However, on \emph{Big} datasets, the lower bias of \DBL^n results in much lower error than RF in terms of both 0-1 Loss and RMSE.
These averaged results also corroborate with the W-D-L results in Table~\ref{tab_wAnJEvsRF}, showing \DBL^n to be a less biased model than RF. 
% --------------------------
% --------------------------

\begin{figure}[t] 
\centering
\includegraphics[width=55mm,height=40mm]{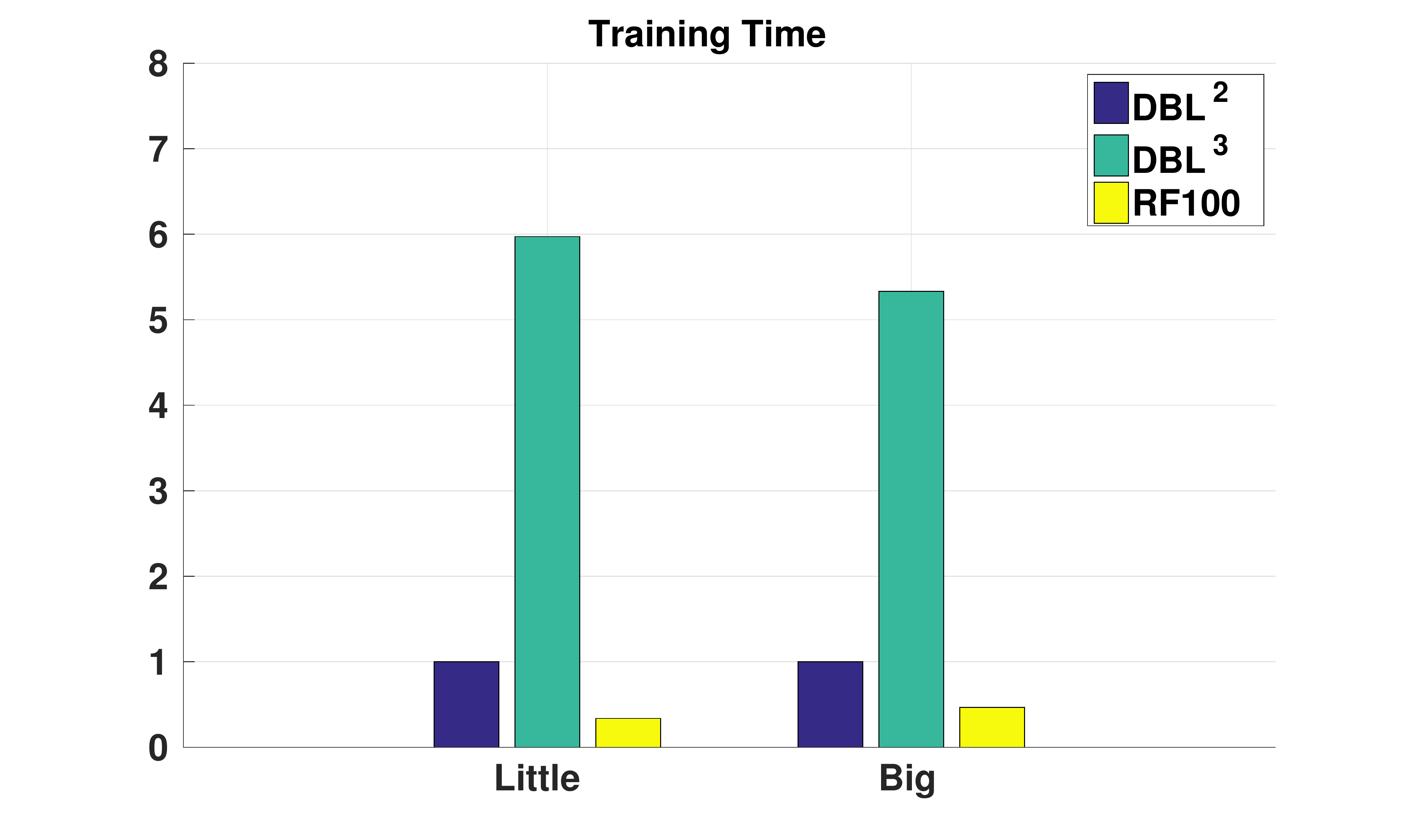}
\includegraphics[width=55mm,height=40mm]{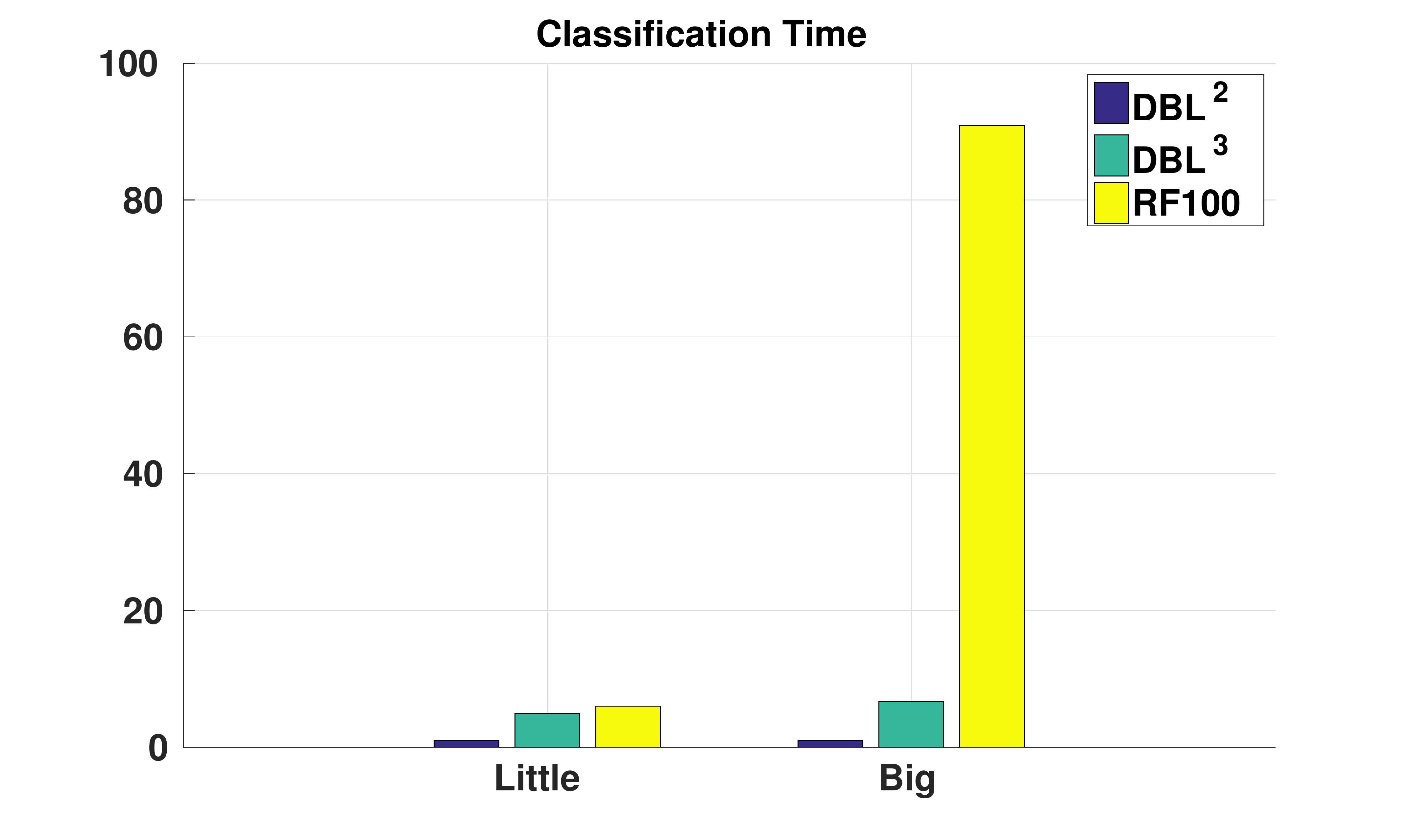}
\caption{\small Geometric average of Training Time (Left), Classification Time (Right) of \DBL^2, \DBL^3 and RF for \emph{Little} and \emph{Big} datasets.}
\label{fig_wAnJEvsRF_TT}
\end{figure}

The comparison of training and classification time of \DBL^n and RF is given in Figure~\ref{fig_wAnJEvsRF_TT}. It can be seen that \DBL^n models are worst than RF in terms of the training time but better in terms of classification time.
% --------------------------
% --------------------------
%%%%%%%%%%%%%%%%%%%%%%%%%%%%%%%%%%%%%%%%%%%%%%%%%%%%%%%%%%%%%%%%%%%%%%%%%%%%%%%%%%%%%%%%%%%%%%%%%%%%%%%%
\section{Conclusion and Future Work}\label{sec_conclusion}
%%%%%%%%%%%%%%%%%%%%%%%%%%%%%%%%%%%%%%%%%%%%%%%%%%%%%%%%%%%%%%%%%%%%%%%%%%%%%%%%%%%%%%%%%%%%%%%%%%%%%%%%
We have presented an algorithm for deep broad learning. DBL consists of parameters that are learned using both generative and discriminative training.
To obtain the generative parameterization for DB, we first developed AnJE, a generative counter-part of higher-order logistic regression. 
We showed that \DBL^n and  \LR^n learn equivalent models, but that \DBL^n is able to exploit the information gained generatively to effectively precondition the optimization process. 
\DBL^n  converges in fewer iterations, leading to its global minimum much more rapidly, resulting in faster training time.
We also compared \DBL^n with the equivalent AnJE and AnDE models and showed that \DBL^n has lower bias than both AnJE and AnDE models.
We compared \DBL^n with state of the art classifier Random Forest and showed that \DBL^n models are indeed lower biased than RF and on bigger datasets \DBL^n often obtains lower 0-1 loss than RF. 

There are a number of exciting new directions for future work.
\begin{itemize}
\item %\gw{I think that we need to qualify the claim that there is only one tunable parameter.  There is also the $m$ used in smoothing.} 
We have showed that \DBL^n is a low bias classifier with minimal tuning parameters and has the ability to handle multiple classes. The obvious extension is to make it out-of-core. We argue that \DBL^n is greatly suited for stochastic gradient descent based methods as it can converge to global minimum very quickly.

\item It may be desirable to utilize a hierarchical DBL, such that $h\DBL^n = \{\DBL^1 \ldots \DBL^n\}$,  incorporating all the parameters up till $n$.  This may be useful for smoothing the parameters. For example, if a certain interaction does not occur in the training data, at classification time one can resort to lower values of $n$. 

\item In this work, we have constrained the values of $n$ to two and three. Scaling-up \DBL^n to higher values of $n$ is greatly desirable. One can exploit the fact that many interactions at higher values of $n$ will not occur in the data and hence can develop sparse implementations of \DBL^n models.

\item Exploring other objective functions such as Mean-Squared-Error or Hinge Loss can result in improving the performance and has been left as a future work.

\item The preliminary version of DBL that we have developed is restricted to categorical data and hence requires that numeric data be discretized.  While our results  show that this is often highly competitive with random forest using local cut-points, on some datasets it is not.  In consequence, there is much scope for investigation of deep broad techniques for numeric data.

\item DBL presents a credible path towards deep broad learning for big data.  We have demonstrated very competitive error on big data and expect future refinements to deliver even more efficient and effective outcomes.
\end{itemize}

%%%%%%%%%%%%%%%%%%%%%%%%%%%%%%%%%%%%%%%%%%%%%%%%%%%%%%%%%%%%%%%%%%%%%%%%%%%%%%%%%%%%%%%%%%%%%%%%%%%%%%%%
\section{Code and Datasets} \label{sec_code}
%%%%%%%%%%%%%%%%%%%%%%%%%%%%%%%%%%%%%%%%%%%%%%%%%%%%%%%%%%%%%%%%%%%%%%%%%%%%%%%%%%%%%%%%%%%%%%%%%%%%%%%%

Code with running instructions can be download from \url{https://www.dropbox.com/sh/iw33mgcku9m2quc/AABXwYewVtm0mVE6KoyMPEVFa?dl=0}.

%%%%%%%%%%%%%%%%%%%%%%%%%%%%%%%%%%%%%%%%%%%%%%%%%%%%%%%%%%%%%%%%%%%%%%%%%%%%%%%%%%%%%%%%%%%%%%%%%%%%%%%%
\section{Acknowledgments}
%%%%%%%%%%%%%%%%%%%%%%%%%%%%%%%%%%%%%%%%%%%%%%%%%%%%%%%%%%%%%%%%%%%%%%%%%%%%%%%%%%%%%%%%%%%%%%%%%%%%%%%%
This research has been supported by the Australian Research Council (ARC) under grants DP140100087, DP120100553, DP140100087 and
Asian Office of Aerospace Research and Development, Air Force Office of Scientific Research under contracts FA2386-12-1-4030, FA2386-15-1-4017 and FA2386-15-1-4007.

%%%%%%%%%%%%%%%%%%%%%%%%%%%%%%%%%%%%%%%%%%%%%%%%%%%%%%%%%%%%%%%%%%%%%%%%%%%%%%%%%%%%%%%%%%%%%%%%%%%%%%%%
\appendix
%%%%%%%%%%%%%%%%%%%%%%%%%%%%%%%%%%%%%%%%%%%%%%%%%%%%%%%%%%%%%%%%%%%%%%%%%%%%%%%%%%%%%%%%%%%%%%%%%%%%%%%%

%%%%%%%%%%%%%%%%%%%%%%%%%%%%%%%%%%%%%%%%%%%%%%%%%%%%%%%%%%%%%%%%%%%%%%%%%%%%%%%%%%%%%%%%%%%%%%%%%%%%%%%%
\section{Detailed Results}~\label{sec_discussion}
%%%%%%%%%%%%%%%%%%%%%%%%%%%%%%%%%%%%%%%%%%%%%%%%%%%%%%%%%%%%%%%%%%%%%%%%%%%%%%%%%%%%%%%%%%%%%%%%%%%%%%%%

In this appendix, we compare the 0-1 Loss and RMSE results of \DBL^n, AnDE and RF. The goal here is to assess the performance of each model on \emph{Big} datasets. Therefore, results on 8 big datasets are reported only in Table~\ref{tab_01LossResults} and~\ref{tab_RMSEResults} for 0-1 Loss and RMSE respectively.
We also compare results with AnJE. Note A1JE is naive Bayes. Also \DBL^1 results are also compared. Note,~\DBL^1 is WANBIA-C~\citep{zaidi13a}.

The best results are shown in bold font.

% -----------------------------------------------------------------------
\begin{table}[h] \center \scriptsize
\setlength{\topmargin}{2pt} \setlength{\tabcolsep}{2.5pt}\renewcommand{\arraystretch}{1}
\caption{\small 0-1 Loss values of A1JE, A2JE, A3JE, A1DE, A2DE, \DBL^1, \DBL^2, \DBL^3, RF and RF (numeric) on \emph{Big} datasets.} 
\begin{center}
%\begin{tabular}{llrrcclrllrrr}
\tabcolsep=2.0pt
%\begin{tabular}{rrrrrrrrrrrrrrrrr}
\begin{tabular}{|p{1.5cm}|p{1.5cm}|p{1.5cm}|p{1.5cm}|p{1.5cm}|p{1.5cm}|p{1.5cm}|p{1.5cm}|p{1.5cm}|p{1.5cm}|p{1.5cm}|p{1.5cm}|p{1.5cm}|p{1.5cm}|p{1.5cm}|p{1.5cm}|p{1.5cm}|p{1.5cm}|}
\hline
\bf  & \bf Locali-zation & \bf Poker-hand & \bf Census-income & \bf Covtype & \bf Kddcup & \bf MITFa-ceSetA & \bf MITFa-ceSetB & \bf MITFa-ceSetC \\
\hline
\bf A1JE	& 0.4938 	& 0.4988 	& 0.2354 	& 0.3143 	& 0.0091 	& 0.0116 	& 0.0268 	& 0.0729       \\
\bf A2JE 	& 0.3653 	& 0.0763 	& 0.2031 	& 0.2546 	& 0.0061 	& 0.0106 	& 0.0239 	& 0.0630       \\
\bf A3JE 	& \bf 0.2813 	& 0.0763 	& 0.1674 	& 0.1665 	& 0.0053 	& 0.0096 	& 0.0215 	& 0.0550        \\
\hline
\bf A1DE 	& 0.3584 	& 0.4640 	& 0.0986 	& 0.2387 	& 0.0025 	& 0.0124 	& 0.0322	& 0.0417       \\
\bf A2DE 	& 0.2844 	& 0.1348 	& 0.0682 	& 0.1552 	& 0.0023 	& 0.0105 	& 0.0325 	& 0.0339        \\
\hline
\bf \DBL^1	& 0.4586 	& 0.4988 	& \bf 0.0433 	& 0.2576 	& 0.0017 	& 0.0012 	& 0.0047 	& 0.0244       \\
\bf \DBL^2 	& 0.3236 	& \bf 0.0021 	& 0.0686 	& 0.1381 	& 0.0014 	& 0.0002 	& 0.0007 	& 0.0007       \\
\bf \DBL^3 	& 0.2974 	& 0.0056 	& 0.0557 	& 0.0797 	& \bf 0.0013 	& \bf 0.0001 	& \bf 0.0005 	& \bf 0.0005       \\
\hline
\bf RF		& 0.2976 	& 0.0687 	& 0.0494 	& \bf 0.0669 	& 0.0015 	& 0.0012 	& 0.0022 	& 0.0013       \\
%\bf RF\tiny(num)& 0.2171 & 0.0687 & 0.0483 & 0.0402 & 0.0014 & 0.0013 & 0.0021 & -       \\
\hline
\end{tabular}
\end{center}
\label{tab_01LossResults}
\end{table}
% -----------------------------------------------------------------------
% -----------------------------------------------------------------------
\begin{table}[h] \center \scriptsize
\setlength{\topmargin}{2pt} \setlength{\tabcolsep}{2.5pt}\renewcommand{\arraystretch}{1}
\caption{\small RMSE values of A1JE, A2JE, A3JE, A1DE, A2DE, \DBL^1, \DBL^2, \DBL^3 and RF on \emph{Big} datasets.} 
%\textrm{OOT} indicates that running time exceeded $400$ hours.} 
\begin{center}
%\begin{tabular}{llrrcclrllrrr}
\tabcolsep=2.0pt
%\begin{tabular}{rrrrrrrrrrrrrrrrr}
\begin{tabular}{|p{1.5cm}|p{1.5cm}|p{1.5cm}|p{1.5cm}|p{1.5cm}|p{1.5cm}|p{1.5cm}|p{1.5cm}|p{1.5cm}|p{1.5cm}|p{1.5cm}|p{1.5cm}|p{1.5cm}|p{1.5cm}|p{1.5cm}|p{1.5cm}|p{1.5cm}|p{1.5cm}|}
\hline
\bf  & \bf Locali-zation & \bf Poker-hand & \bf Census-income & \bf Covtype & \bf Kddcup & \bf MITFa-ceSetA & \bf MITFa-ceSetB & \bf MITFa-ceSetC \\
\hline
\bf A1JE	& 0.2386 	& 0.2382 	& 0.4599 	& 0.2511 	& 0.0204 	& 0.1053 	& 0.1607 	& 0.2643 \\
\bf A2JE 	& 0.2115 	& 0.1924 	& 0.4231 	& 0.2256 	& 0.0170 	& 0.1006 	& 0.1516 	& 0.2455 \\
\bf A3JE 	& 0.1972 	& 0.1721 	& 0.3812 	& 0.1857 	& 0.0160 	& 0.0954 	& 0.1436 	& 0.2293 \\
\hline
\bf A1DE 	& 0.2090 	& 0.2217 	& 0.2780 	& 0.2174 	& 0.0103 	& 0.1079 	& 0.1746 	& 0.1989 \\
\bf A2DE 	& \bf 0.1890 	& 0.2044 	& 0.2269 	& 0.1779 	& 0.0098 	& 0.0983 	& 0.1745 	& 0.1530  \\
\hline
\bf \DBL^1	& 0.2330 	& 0.2382 	& \bf 0.1807 	& 0.2254 	& 0.0072 	& 0.0347 	& 0.0602 	& 0.1360 \\
\bf \DBL^2	& 0.2179 	& 0.1970 	& 0.250 	& 0.1802 	& 0.0068 	& 0.0123 	& 0.0248 	& 0.0257 \\
\bf \DBL^3	& 0.2273 	& \bf 0.0323 	& 0.2332 	& 0.1494 	& \bf 0.0065 	& \bf 0.0105 	& \bf 0.0198 	& \bf 0.0241       \\
\hline
\bf RF		& 0.1939 	& 0.1479 	& 0.1928 	& \bf 0.1336 	& 0.0072 	& 0.0296 	& 0.0484 	& 0.0651 \\
%\bf RF\tiny(num)	& 0.1530 & 0.0524 & 0.0412 & 0.0391 & 0.0013 & 0.0011 & 0.0016 & 0.0005\\
\hline
\end{tabular}
\end{center}
\label{tab_RMSEResults}
\end{table}
% -----------------------------------------------------------------------

%%%%%%%%%%%%%%%%%%%%%%%%%%%%%%%%%%%%%%%%%%%%%%%%%%%%%%%%%%%%%%%%%%%%%%%%%%%%%%%%%%%%%%%%%%%%%%%%%%%%%%%%
\bibliography{DBL}
%%%%%%%%%%%%%%%%%%%%%%%%%%%%%%%%%%%%%%%%%%%%%%%%%%%%%%%%%%%%%%%%%%%%%%%%%%%%%%%%%%%%%%%%%%%%%%%%%%%%%%%%

\end{document}